\pdfoutput=1

\documentclass{article}

\usepackage[main,final]{neurips_2026}

\usepackage[utf8]{inputenc} %
\usepackage[T1]{fontenc}    %
\usepackage{hyperref}       %
\hypersetup{hidelinks}
\usepackage{url}            %
\usepackage{booktabs}       %
\usepackage{amsfonts}       %
\usepackage{nicefrac}       %
\usepackage{microtype}      %
\usepackage{xcolor}         %
\usepackage{placeins}

\usepackage{graphicx}
\usepackage{subcaption}
\usepackage{comment}
\usepackage{verbatim}
\usepackage{threeparttable}
\usepackage[separate-uncertainty=true, multi-part-units=repeat]{siunitx}
\usepackage{multirow}
\usepackage{amsmath}
\usepackage{amssymb}
\usepackage{tabularx}
\usepackage{mathtools}
\usepackage{amsthm}
\usepackage[capitalize,noabbrev]{cleveref}
\usepackage[disable,textsize=tiny]{todonotes}

\usepackage{wrapfig}

\theoremstyle{plain}

\theoremstyle{definition}

\theoremstyle{remark}

\DeclareRobustCommand{\kg}[1]{\ifmmode{\color{black}#1}\else\leavevmode{\color{black} #1}\fi}
\newcommand{\name}{\texttt{ImmuVis} }
\newcommand{\nameNoSpace}{\texttt{ImmuVis}}
\newcommand{\nameConvNoSpace}{\texttt{ImmuVis}_{\texttt{Conv}}}
\newcommand{\nameConv}{\texttt{ImmuVis}_{\texttt{Conv}} }

\newcommand{\nameViT}{\texttt{ImmuVis}_{\texttt{ViT}} }
\newcommand{\nameSwin}{\texttt{ImmuVis}_{\texttt{Swin}} }
\newcommand{\stemOp}{\mathfrak{S}}
\newcommand{\hyperConvOp}{\mathfrak{H}_{e}}
\newcommand{\hyperConvOpDec}{\mathfrak{H}_{d}}
\newcommand{\headOp}{\mathfrak{R}}

\newcommand{\numImages}{24,405}
\newcommand{\numMarkers}{265}
\newcommand{\numDatasets}{28}
\newcommand{\numPatches}{over 17M}
\newcommand{\datasetName}{IMC17M }
\newcommand{\datasetNameNoSpace}{IMC17M}
\newcommand{\numPanels}{25}
\newcommand{\numHistopathologies}{14}
\newcommand{\minMarkers}{28}
\newcommand{\maxMarkers}{47}
\newcommand{\typicalMarkerRange}{35-45}

\newcommand{\patchSize}{128}

\newcommand{\patchMaskSize}{8}
\newcommand{\learningRateMax}{5 \times 10^{-4}}
\newcommand{\learningRateMin}{10^{-6}}

\title{\textbf{\texttt{\nameNoSpace}}: Hyperconvolutional Foundation Models for Imaging Mass Cytometry}

\author{%
Dawid Uchal$^{1*}$ \hspace{0.5em} Marcin Mo\.zejko$^{1*}$ \hspace{0.5em} Krzysztof Gogolewski$^{1,2*}$ \hspace{0.5em} Piotr Kupidura$^{1}$ \\
\textbf{Fabian Ozga}$^{1}$ \hspace{0.5em} 
\textbf{Szymon {\L}ukasik}$^1$ \hspace{0.5em} \textbf{Jakub Giezga{\l}a}$^1$ \hspace{0.5em} \textbf{Tomasz Noco\'n}$^1$ \hspace{0.5em} \textbf{Kacper Pietrzyk}$^1$\\
\textbf{Robert Pieniuta}$^{1}$ \hspace{0.5em} \textbf{Mateusz Sulimowicz}$^{1}$ \hspace{0.5em} 
\textbf{Michał Zmys{\l}owski}$^{1}$ \hspace{0.5em}
\textbf{Micha{\l} Orzy{\l}owski}$^{1}$ \\ \textbf{Tomasz Si{\l}kowski}$^{1}$ \hspace{0.5em}
\textbf{Karol Zagródka}$^{1}$  \hspace{0.5em} \textbf{Eike Staub}$^{3}$ \hspace{0.5em} \textbf{Ewa Szczurek}$^{1,4,5,6}$\\
$^1$Faculty of Mathematics, Informatics and Mechanics, University of Warsaw, Warsaw, Poland, \\
$^2$Center of Trustworthy AI for Life Sciences - International Research Agendas Programme,\\ University of Warsaw, Warsaw, Poland $^3$Merck Healthcare KGaA, Darmstadt, Germany,\\
$^4$Institute of AI for Health, Helmholtz Munich, Neuherberg, Germany,
$^5$Faculty of Mathematics,\\ Informatics and Statistics,  Ludwig-Maximilians-Universität München, Munich, Germany,\\
$^6$Munich Center for Machine Learning, $^{*}$shared first author order determined by coin flips \\
\texttt{em.szczurek@uw.edu.pl}
}

\begin{document}

\maketitle

\begin{abstract}
We present \nameNoSpace, a family of {\textbf{efficient foundation models}} for imaging mass cytometry (IMC), a high-throughput multiplex imaging technology that handles molecular marker measurements as image channels and enables large-scale spatial tissue profiling.
Unlike natural images, multiplex imaging lacks a fixed channel space, as real-world marker sets vary across studies, violating a core assumption of standard vision backbones.
To address this, \name introduces {\textbf{marker-adaptive hyperconvolutions}} that generate convolutional kernels from learned marker embeddings, enabling a single model to operate on 
\kg{any subset of the markers seen during pretraining.}
We pretrain \name on {\textbf{the largest dataset to date}}, \datasetName  (\numDatasets{} cohorts, \numImages{} images, \numMarkers{} markers, \numPatches{} patches),
using self-supervised masked reconstruction.
\kg{\name achieves the best average cell-typing and strong virtual staining performance, while remaining competitive with state-of-the-art approaches on clinical prediction} at substantially lower compute cost, and is the sole model that provides \textbf{calibrated uncertainty}.
These results position \name as a practical framework for real-world IMC modeling.
\end{abstract}

\section{Introduction}

Imaging Mass Cytometry (IMC)
profiles protein abundances in tissues at subcellular spatial resolution by measuring metal-tagged antibodies via mass spectrometry~\cite{Giesen2014,Chang2017}.
IMC is increasingly central to spatially resolved single-cell analysis of healthy and disease tissues, and both datasets and preprocessing pipelines are rapidly maturing~\cite{Milosevic2023, natureEditorial2024}.
Panel richness matters: larger marker panels better resolve cell types and functional states, enabling mapping of tissue
spatial neighborhoods~\cite{Bollhagen2024,deSouza2024}.

\begin{wrapfigure}{r}{0.51\textwidth}
    \centering
\includegraphics[width=0.5\textwidth]{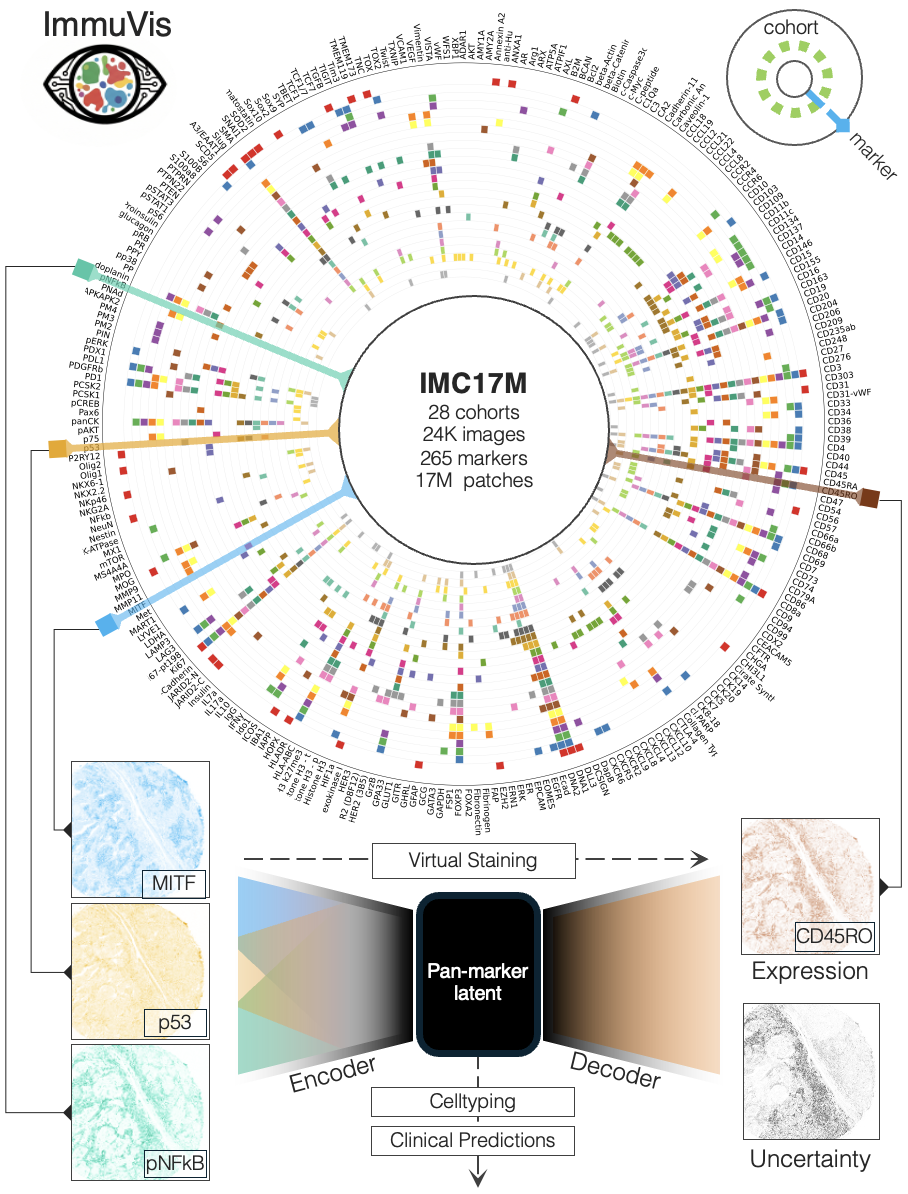}  \caption{ \textbf{Motivation for \nameNoSpace: real-world panel heterogeneity.} \datasetName exhibits strong cohort-marker diversity (rings: cohorts; radii: markers; colored ticks: measured markers), motivating a single model that operates on 
\kg{different marker subsets.}
\name encodes 
\kg{an observed panel} (e.g., MITF, p53, pNFkB) into a shared pan-marker latent space for downstream phenotyping (cell typing, clinical prediction) and instantiates a task-specific decoder to virtually stain requested targets (e.g., CD45RO), outputting both expression and predictive uncertainty.} \label{fig:motivation} 
\vspace{-1cm}
\end{wrapfigure}

Still, cohort-specific IMC panels are typically restricted to only tens of markers because of physical channel interference that reduces the number of concurrently usable metal isotopes for antibody tagging~\cite{Bendall2012DeepProfilersGuide,Chevrier2018SpilloverCompensation}.
In practice, as presented in Figure~\ref{fig:motivation},
cross-cohort
marker overlap is limited, making {\textbf{modelling and analysis across heterogeneous panels}} a practical necessity~\cite{Bussi2024}.
This also motivates \textbf{virtual staining}, i.e. predicting cohort-dependent \emph{missing-marker sets} from the \emph{measured} panel, to enable denser tissue phenotyping without additional wet-lab work~\cite{Latonen2024}.
Basic approaches to this task suffer from a fixed input problem, i.e. are limited to a fixed, narrow marker panel ~\cite{Lo2022,Shaban2023}, which prevents panel flexibility and straightforward extension to diverse marker sets.
VirTues~\cite{virtues} and Eva~\cite{Liu2025-EVA}, recent Transformer-based models, support variable marker inputs by concatenating marker expression-specific spatial tokens augmented with marker-identity encodings derived from protein language models.
However, since the spatial tokens provide only a global, flattened view per marker image patch, these models effectively ignore crucial local marker dependencies.
In contrast, variable input learning in other vision domains is often implemented via conditional operators, e.g. hypernetworks~\cite{ha2017hypernetworks} and dynamic convolutions~\cite{chen2020dynamicConvolution}, that modulate the feature extractor conditioned on the input in a computationally efficient manner; yet this approach remains underused in multiplex imaging.

Virtual staining is also a \textbf{reliability problem}: when the measured marker set provides weak evidence for a target marker (e.g., rare phenotypes or atypical tissue patterns), accurate point prediction is ill-posed and visually plausible reconstructions may still be biologically wrong~\cite{Dolezal2022}. In this regime, {\textbf{uncertainty}} must be a part of the model output, highlighting unconfident predictions.
Finally, cohort-scale use places a strong emphasis on {\textbf{inference time}}. Transformer-heavy virtual-staining models can become increasingly slow as marker count and spatial resolution grow~\cite{Papa2024}.
To summarize, to the best of our knowledge, no prior virtual-staining model combines \textbf{panel flexibility} and \textbf{calibrated uncertainty} with  \textbf{inference speed} for IMC.

To address these challenges, we introduce \nameNoSpace, a family of foundation models for IMC data that unifies variable-input learning, efficiency, and uncertainty in a single architecture. Our contributions are as follows:

\noindent \textbf{Hyperconvolutional architecture}: We introduce a channel-adaptive hyperconvolution module that provides operator-level channel adaptivity and generates convolutional kernels conditioned on learned marker embeddings, enabling a single model that uses standard computer vision backbones to process 
\kg{any combination of the pretrained vocabulary}
without architectural modification or retraining. 

\noindent \textbf{Family of efficient foundation models:}
We introduce a family of foundation models based on our hyperconvolutional architecture, that uses different established computer vision backbones spanning from fully convolutional $\nameConv$ that incorporates ConvNeXt v2 \cite{convnext2}, through window-attention-based $\nameSwin$ \cite{SwinTransformer}, up to token-based $\nameViT$  \cite{dosovitskiy2021an}. The entire family offers substantially lower inference cost than token-concatenation-based alternatives \cite{virtues, Liu2025-EVA}.

\noindent \textbf{Uncertainty-aware learning objective:} We model predictive uncertainty via modified Gaussian heteroscedastic regression, producing uncertainty estimates that highly correlate with reconstruction error, making virtual staining reliability-aware rather than purely point-estimated.

We pretrain \name on the largest IMC dataset compiled to our knowledge (\datasetNameNoSpace; \numImages{} images, \numMarkers{} markers, \numDatasets{} datasets) using masked-channel reconstruction, and show that the learned representations transfer across datasets 
\kg{to state-of-the-art performance in virtual staining and patch-level cell typing, with clinical prediction competitive with state-of-the-art approaches.}
Taken together, \name provides a practical, panel-flexible IMC foundation models family that is deployable at cohort scale and enables reliability-aware virtual staining.
More broadly, it offers a general recipe for variable-input foundation models in multiplex imaging and beyond. 
\kg{The code and pretrained models are available at \url{https://github.com/szczurek-lab/ImmuVis} and \url{https://huggingface.co/collections/SzczurekLab/immuvis}, respectively.}

\section{Related Work}

\paragraph{Handling heterogeneous panels and virtual staining for multiplex data.}
Virtual staining aims at prediction of unmeasured markers, increasing phenotyping depth and enabling retrospective upgrading of legacy cohorts without additional wet-lab assays~\cite{virtues,Zidane2023}.
Early approaches to this problem typically assume a fixed input panel and train U-Net-like architectures to predict new markers~\cite{Shaban2023, Ternes2022}. However, these methods lack panel flexibility, as they are restricted to datasets whose panels contain markers used during training and cannot exploit additional available markers.
To achieve panel flexibility, \cite{Liu2025-EVA, Kraus2024} adopt a channel-extended ViT design from ~\cite{bao2024channel}, which concatenates marker-specific spatial tokens with marker-identity encodings and processes them using a transformer model. However, this design incurs a heavy combinatorial marker$\times$space computational cost. VirTues~\cite{virtues} mitigates this limitation via marker-space attention factorization. Additionally, VirTues and EVA \cite{Liu2025-EVA} use pretrained language models embeddings as marker encodings enabling predictions for previously unseen markers. Still, in these transformer tokenization-based designs, marker identity is injected \emph{after} tokenization and therefore may underutilize fine-grained, local cross-marker dependencies that could be captured by conditioning the feature extractor itself.

\paragraph{Foundation-model trends for multiplex imaging.}
Large-scale self-supervised pretraining~\cite{Bommasani2021FoundationModels} has enabled transferable visual representations via masked prediction \cite{virtues, Liu2025-EVA, Kraus2024} or self-supervised distillation objectives ~\cite{oquab2024dinov}.
For multiplex imaging, self-supervision is attractive because heterogeneous collections can be exploited without dense per-marker labels. However, as IMC introduces modality-specific artifacts (e.g., hot pixels and ion-counting noise), a recent self-supervised foundation model for spatial proteomics, KRONOS, explicitly excludes ion-based modalities such as IMC from training and notes that they may require architectural modifications or modality-specific adaptation strategies~\cite{KRONOS2025}.

\section{Methodology}

\subsection{\name architecture}

Let $\mathcal{M}\!=\!\{m_1,\dots,m_N\}$ denote the global pan-cohort marker vocabulary \kg{observed during the training}.
Given two \emph{ordered} marker sets $\mathcal{I}_{e}, \mathcal{J}_{d}\!\subset\!\mathcal{M}$, indexed as
$\mathcal{I}_{e}\!=\!(i^{1},\dots,i^{C_e})$ and $\mathcal{J}_{d}\!=\!(j^{1},\dots,j^{C_d})$,
with $C_e\!=\!|\mathcal{I}_{e}|$ and $C_d\!=\!|\mathcal{J}_{d}|$,
\name instantiates a marker-specific encoder-decoder pair:

\begin{equation*}
    \name \left(\mathcal{I}_{e}, \mathcal{J}_{d}\right) =\left( \operatorname{Enc}^{ \mathcal{I}_{e}}, \operatorname{Dec}^{ \mathcal{J}_{d} }
    \right).
\end{equation*}

\begin{figure}[t]
  \begin{center}
    \centerline{\includegraphics[width=\textwidth]{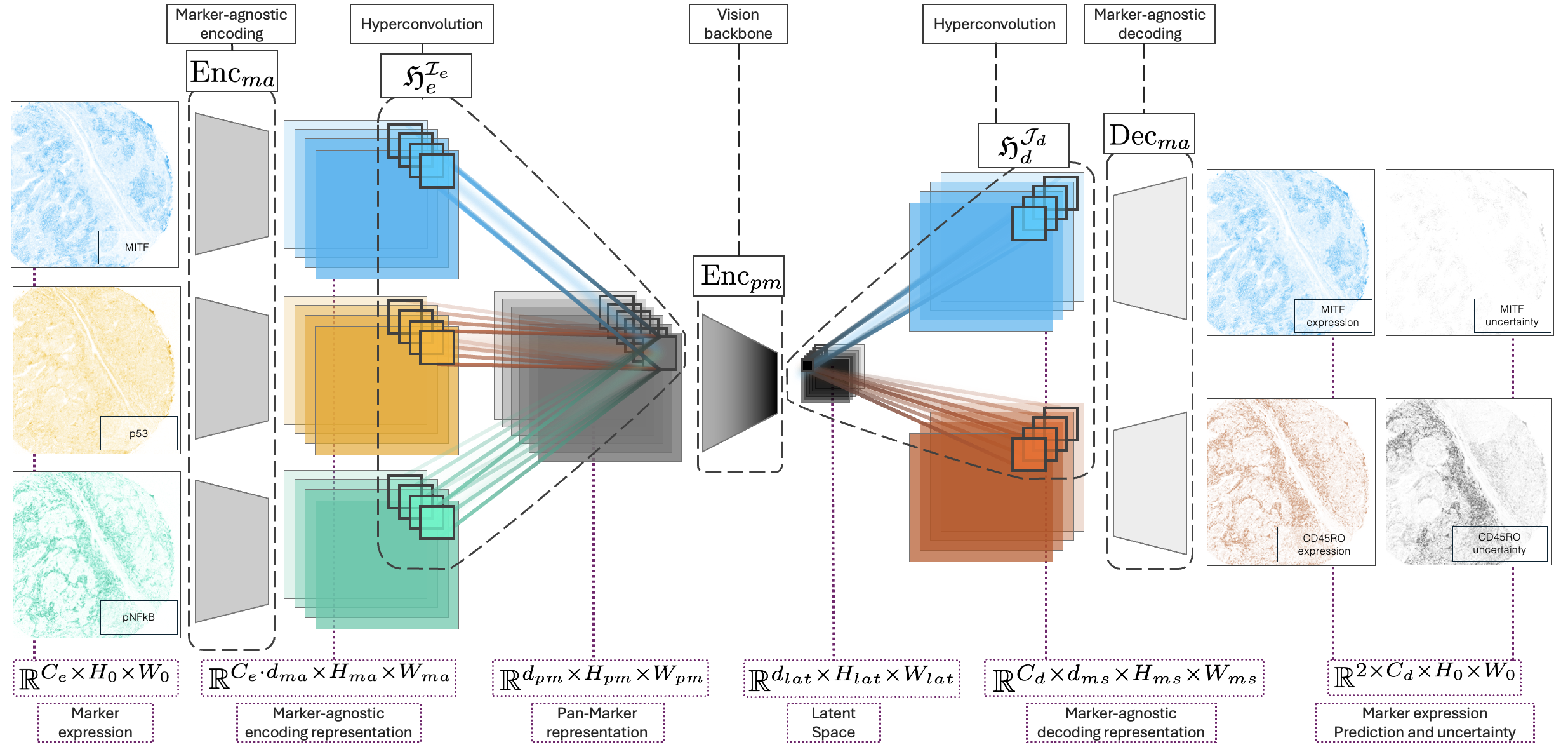}}
    \caption{
      \textbf{\name architecture overview.} Marker-agnostic encoder stems embed each input marker channel and a hyperconvolution module, conditioned on learned marker embeddings, fuses them into a shared pan-marker representation processed by a standard vision backbone. A symmetric hyperconvolution and marker-agnostic decoder instantiates the requested output marker set, predicting per-marker reconstructions together with pixel-wise uncertainty (modified heteroscedastic log-variance).
    }
    \label{fig:architecture}
  \end{center}
\end{figure}

The encoder
$\operatorname{Enc}^{\mathcal{I}_{e}}$
maps the multiplex image
$\mathbf{X}\!=\!\left(\mathbf{X}_{i^1}, \dots, \mathbf{X}_{{i^{C_e}}}\right)\!\in\! \mathbb{R}^{C_e \times H_{0} \times W_{0}}$,
to a pan-marker latent representation 
$\mathbf{Z}\!=\!\operatorname{Enc}^{\mathcal{I}_{e}} \left(\mathbf{X}^{}\right)\!\in\!\mathbb{R}^{d_{lat} \times H_{lat} \times W_{lat}}$ \kg{shared across the input markers}.
The decoder $\operatorname{Dec}^{\mathcal{J}_{d}}$ maps  $\mathbf{Z}$ to image-level point-wise  predictions for the markers in $\mathcal{J}_{d}$, producing the mean $\mathbf{{X}}^{\mu}$ and the corresponding log-variance $\mathbf{{X}}^{\log\sigma^{2}}$:
\[
(\mathbf{X}^{\mu},\,\mathbf{X}^{\log\sigma^{2}})
=
\operatorname{Dec}^{\mathcal{J}_{d}}(\mathbf{Z}) \in \mathbb{R}^{2\times C_d\times H_0\times W_0}.
\]
\name architecture overview is presented in Figure~\ref{fig:architecture}.
\subsubsection{$\operatorname{Enc}^{\mathcal{I}_{e}}$ hypernetwork}

Given
$\mathcal{I}_{e}$, the instantiated encoder
factors to a composition:
\begin{equation*}
    \mathbf{Z}=\operatorname{Enc}^{\mathcal{I}_{e}}(\mathbf{X})= \operatorname{Enc}_{pm} \circ \operatorname{\hyperConvOp^{\mathcal{I}_{e}}\circ \operatorname{Enc_{ma}}}(\mathbf{X}),
\end{equation*}
where $\operatorname{Enc}_{ma}$ is a \emph{marker-agnostic} encoder applied independently per input marker channel,
$\hyperConvOp^{\mathcal{I}_{e}}$ is a \emph{marker-conditional hyperconvolution operator} instantiated from marker embeddings for the input set $\mathcal{I}_{e}$,
and $\operatorname{Enc}_{pm}$ maps the resulting features into a shared \emph{pan-marker} latent space. Formal definitions are provided below.

\paragraph{Marker-agnostic encoding $\operatorname{Enc}_{ma}$.}

Let $\stemOp: \mathbb{R}^{1\times H_{0}\times W_{0}} \rightarrow \mathbb{R}^{d_{ma} \times H_{ma} \times W_{ma}}$ be a convolutional \textit{stem}
shared across markers, where $d_{ma}$ is the marker-agnostic feature width.
$\stemOp$ preprocesses each marker-channel independently, downsampling the image while capturing local spatial expression patterns.
Applying $\stemOp$ to each marker channel and concatenating along the channel axis yields
$\operatorname{Enc}_{ma}:\mathbb{R}^{C_e\times H_0\times W_0}\to\mathbb{R}^{C_e \cdot d_{ma}\times H_{ma}\times W_{ma}}$ given as:
\[
\mathbf{W}=\operatorname{Enc}_{ma}(\mathbf{X})
=\big[\,\stemOp(\mathbf{X}_{i^1});\dots;\stemOp(\mathbf{X}_{i^{C_e}})\,\big]_{\mathrm{0}}
\]
providing compressed, marker-agnostic representation $\mathbf{W}$ of $\mathbf{X}$.
Here $[\,\cdot;\cdot\,]_{\mathrm{k}}$ denotes the concatenation along the $\mathrm{k}$-th dimension.

\paragraph{Encoder Hyperconvolution operator $\hyperConvOp$.}
Let
$
\phi_{e}:\mathcal{M}\to\mathbb{R}^{d_{\mathrm{pm}}\times d_{\mathrm{ma}}\times h_{e}\times w_{e}},
$
be  a learnable marker-conditional convolutional kernel generator, where $d_{\mathrm{pm}}$ is the channel width of the pan-marker space, and $(h_e,w_e)$ is the spatial kernel size.
For $\mathcal{I}_e$, we formulate a hyperkernel $\mathbf{H}^{\mathcal{I}_{e}}_{e} \!\in\!\mathbb{R}^{d_{pm} \times C_e \cdot d_{ma} \times h_{e} \times w_{e}}$ as:
\begin{equation*}
\label{eq:hyperkernel}
\mathbf{H}^{\mathcal{I}_{e}}_{e}
\!=\!
\big[\phi_{e}(i_{}^{1});\dots;\phi_{e}(i_{}^{C_e})\big]_{\mathrm{1}}.
\end{equation*}
Then $\hyperConvOp^{\mathcal{I}_{e}} :  \mathbb{R}^{C_e \cdot d_{ma} \times H_{ma}\times W_{ma}} \rightarrow  \mathbb{R}^{d_{pm} \times H_{pm} \times W_{pm}}$ is a single dynamic convolution operator~\cite{Li2024KernelWarehouse,Chen_2020_CVPR_DynamicConv}:
\begin{equation*}
    \mathbf{V} = \hyperConvOp^{\mathcal{I}_{e}}\left(\mathbf{W}\right) = \mathbf{W} \otimes_{s,p} \mathbf{H}^{\mathcal{I}_{e}}_{e},
\end{equation*}
where $\otimes_{s,p}$ is a standard 2D convolution operation with stride $s$ and padding $p$, and $H_{pm}$ and $W_{pm}$ are the output spatial dimensions.
Now $\mathbf{V}$
is an embedding of $\mathbf{X}$ into a unified pan-marker representation. Since $\hyperConvOp^{\mathcal{I}_{e}}$ depends on $\mathcal{I}_{e}$, the induced hyperkernel adapts to the observed markers and enables capturing fine-grained local cross-marker dependencies.
\kg{Equivalently, fusion sums marker-specific linear convolutions,}
\[
\kg{\mathbf{V}=\sum_{m\in\mathcal{I}_e}\stemOp(\mathbf{X}_m)\otimes_{s,p}\phi_e(m).}
\]
\kg{Jointly permuting marker identities and their channels leaves this sum unchanged. Fusion cost grows linearly with the number of input markers, without a marker-by-marker interaction term.}

\paragraph{Pan-marker backbone $\operatorname{Enc}_{pm}$.}
Since $\mathbf{V}$ is a universal, fixed-sized embedding, it enables the application of standard computer vision backbones. Specifically,
$\operatorname{Enc}_{pm}:\mathbb{R}^{d_{pm} \times H_{pm} \times W_{pm}} \rightarrow \mathbb{R}^{d_{lat} \times H_{lat} \times W_{lat}}$
is any fixed-channel vision backbone. This choice lets us leverage advances in general-purpose vision models while remaining flexible to 
\kg{different input marker sets,}
producing the pan-marker latent representation 
$
    \mathbf{Z} =\operatorname{Enc}_{pm}(\mathbf{V}).
$

\subsubsection{$\operatorname{Dec}^{\mathcal{J}_{d}}$ hypernetwork}

Analogously to the encoding procedure, given target marker-set specific $\mathcal{J}_{d}$,
the instantiated decoder $\operatorname{Dec}^{\mathcal{J}_{d}}$ is a composition:
$
    \mathbf{X^{\mu}}, \mathbf{X}^{\log\sigma^{2}}=\operatorname{Dec}^{\mathcal{J}_{d}}(\mathbf{Z})= \operatorname{Dec}_{\mathrm{ma}} \circ \operatorname{\hyperConvOpDec^{\mathcal{J}_{d}}}(\mathbf{Z}),
$
where $\hyperConvOpDec^{\mathcal{J}_{d}}$ is a \emph{marker-conditioned} hyperconvolution that maps the latent representation to \textit{marker-specific} decoding features, and $\operatorname{Dec}_{\mathrm{ma}}$ produces per-marker mean predictions and log-variance for markers in $\mathcal{J}_{d}$. Formal definitions are provided below.

\paragraph{Decoder Hyperconvolution operator $\hyperConvOpDec$.}
Let
$\phi_{d}:\mathcal{M}\to\mathbb{R}^{d_{\mathrm{ms}}\times d_{\mathrm{lat}}\times h_d\times w_d},$
be  a learnable marker-conditional convolutional kernel generator, where $d_{\mathrm{ms}}$ is the width of the marker set specific decoder space, and $(h_d,w_d)$ is the spatial kernel size.
We define $\hyperConvOpDec^{\mathcal{J}_{d}}: \mathbb{R}^{d_{lat} \times H_{lat} \times W_{lat}} \rightarrow \mathbb{R}^{C_d \times d_{ms} \times H_{ms} \times W_{ms}}$ as:
\begin{equation*}
\begin{aligned}
     \mathbf{U} = \hyperConvOpDec^{\mathcal{J}_{d}}\left(\mathbf{Z}\right) =
     \left[\mathbf{Z} \otimes_{s,p} \phi_{d}\left(j_{}^{1}\right); \dots; \mathbf{Z} \otimes_{s,p}\phi_{d}\left(j_{}^{C_{d}}\right)\right]^{0}
\end{aligned}
\end{equation*}
where $\mathbf{U} = (\mathbf{U}_{j^{1}},\dots,\mathbf{U}_{j^{C_d}})$ constitutes an embedding of a latent space $\mathbf{Z}$ into a marker set specific space. Here $[\,\cdot;\cdot\,]^{\mathrm{k}}$ denotes stacking along the $\mathrm{k}$-th dimension.

\paragraph{Marker-agnostic decoding $\operatorname{Dec}_{ma}$.}

Let $\headOp: \mathbb{R}^{d_{ms}\times H_{ms}\times W_{ms}} \rightarrow \mathbb{R}^{2 \times H_{0} \times W_{0}}$,
be a convolutional \textit{head} operator, that provides marker predictions and uncertainties from a marker-specific representations. $\operatorname{Dec}_{ma}: \mathbb{R}^{C_{d}\times d_{ms} \times H_{ms} \times W_{ms}} \rightarrow \mathbb{R}^{2 \times C_{d} \times H_{0} \times W_{0}}$ applies $\headOp$ to each marker channel independently and stacks the results:
\begin{equation*}
\begin{aligned}
    \widehat{\mathbf{X}}=&\operatorname{Dec}_{ma}\left(\mathbf{U}\right)
    =
    \left[\headOp\left(\mathbf{U}_{j^{1}_{}}\right); \dots ;\headOp\left(\mathbf{U}_{j^{C_{d}}_{}}\right)\right]^{\mathrm{1}}.
\end{aligned}
\end{equation*}
Finally, we set $\mathbf{X^{\mu}}=\widehat{\mathbf{X}^0}$ and $\mathbf{X^{\log\sigma^{2}}}=\widehat{\mathbf{X}^{1}}$ to obtain the final point-wise prediction and its uncertainty approximation for markers from $\mathcal{J}_{d}$, respectively.

\subsection{Masked Modelling Task}\label{sec:training}
We train \name with a masked modelling objective~\cite{He2022_Masked_Autoenc}.
Let $\mathbf{X}_{\mathcal{A}}$ denote an multiplexed image $\mathbf{X}$ restricted only to markers from marker set $\mathcal{A}$. During the training, for each image $\mathbf{X}_{\mathcal{I}_\mathrm{img}}\in\mathbb{R}^{C_{\mathrm{img}}\times H\times W}$ with a marker set $\mathcal{I}_\mathrm{img}$  we:
(i) sample a target marker set $\mathcal{I}_{\mathrm{tgt}}\subseteq\mathcal{I}_{\mathrm{img}}$,
(ii) sample an input marker set $\mathcal{I}_{\mathrm{in}}\subset\mathcal{I}_{\mathrm{tgt}}$, and
(iii) apply patch-wise spatial masking on $\mathbf{X}_{\mathcal{I}_{\mathrm{in}}}$ to obtain a masked input $\tilde{\mathbf{X}}_{\mathcal{I}_{\mathrm{in}}}$. $\tilde{\mathbf{X}}_{\mathcal{I}_{\mathrm{in}}}$ is then fed to a model to obtain unmasked point-wise predictions and uncertainty estimations for all markers from set $\mathcal{I}_{\mathrm{tgt}}$:
\begin{equation*}
(\mathbf{X}^{{\mu}}_{\mathcal{I}_{\mathrm{tgt}}},
\mathbf{X}^{\log\sigma^2}_{\mathcal{I}_{\mathrm{tgt}}})
=
\nameNoSpace\big(\mathcal{I}_{\mathrm{in}},\,\mathcal{I}_{\mathrm{tgt}}\big)\left(\tilde{\mathbf{X}}_{\mathcal{I}_{\mathrm{in}}}\right).
\label{eq:masked_forward}
\end{equation*}

At test time, we set $\mathcal{I}_{\mathrm{tgt}}=\mathcal{I}_{\mathrm{img}}$.
For detailed procedures of sampling and masking see Appendix~\ref{app:masking_details}.

\paragraph{Loss function}\label{sec:training_uncertainty}
We train \name using a Gaussian heteroscedastic regression scheme~\cite{Stirn2022FaithfulHR}. It assumes that each element $\mathbf{X}_{c,h,w}$ of a target image $\mathbf{X}_{\mathcal{I}_{\mathrm{tgt}}}$ follows an independent Gaussian distribution:
\[
\textbf{X}_{c,h,w} \sim \mathcal{N}\!\left(\mathbf{X}^{\mu}_{c,h,w}, \exp\!\left(\mathbf{X}^{\log\sigma^2}_{c,h,w}\right)\right).
\]

Then the training objective is, up to an additive constant, the modified Gaussian negative log-likelihood, $\beta$-NLL, following ~\cite{seitzer2022on}:
\begin{equation*}
\begin{aligned}
\hat{\sigma}^2 &= \exp\left(\mathrm{clamp}(\mathbf{X}^{\log\sigma^2}_{\mathcal{I}_{\mathrm{tgt}}})\right), \quad \epsilon = 10^{-8},\\
\mathcal{L}_{\mathrm{\beta\text{-}NLL}}
&= \mathrm{mean}\!\left[ \lfloor \hat{\sigma}^{2\beta} \rfloor \left(
\frac{(\mathbf{X}_{\mathcal{I}_{\mathrm{tgt}}}-\mathbf{X}^{\mu}_{\mathcal{I}_{\mathrm{tgt}}})^2}{\hat{\sigma}^2+\epsilon} + \log\hat{\sigma}^2
\right)\right],
\end{aligned}
\label{eq:nll}
\end{equation*}
where the mean is taken over the batch, channels, and spatial dimensions, and $\lfloor \cdot \rfloor$ denotes the stop gradient operation. The variance $\hat{\sigma}^2$ is truncated using a gradient-preserving clamp to improve training stability.
For implementation details see Appendix~\ref{app:uncertainty_details}.

\section{Results}

\subsection{\datasetName dataset}
To train the \name  family, we curated \datasetNameNoSpace, the largest IMC corpus reported to date, composed of 17M patches from \numDatasets{} datasets spanning \numPanels{} unique marker panels (\numImages{} images and \numMarkers{} markers) across \numHistopathologies{} histologies (see Appendix~\ref{sec:dataset_details} and Table~\ref{tab:datasets}). For training we split images into train/test in a 4:1 ratio, stratifying by dataset panel to preserve panel- and histopathology-level representation. All patches derived from the same parent image were assigned to the same split to prevent patch-level leakage.

\subsection{\name model family}

\name is a model family in which the choice of modules' backbone instantiates a specific architecture. We consider three variations: (i) a fully convolutional $\nameConv$ based on ConvNeXt v2~\citep{convnext2}, (ii) a Swin Transformer-based $\nameSwin$~\citep{SwinTransformer}, and (iii) a ViT-based $\nameViT$~\citep{dosovitskiy2021an}. All variants use a pan-marker latent dimension $d_{\mathrm{lat}}=768$; architectural details are provided in Appendix~\ref{sec:model_conf}.

\paragraph{Implementation of $\phi_{e}$ and $\phi_{d}$}
We implement the convolutional kernel generators $\phi_{e}$ and $\phi_{d}$ as learnable marker-kernel lookup tables trained from scratch. 
This contrasts with VirTues~\citep{virtues} and EVA~\citep{Liu2025-EVA}, which rely on marker encodings derived from protein language models~\citep{Lin2023-ESM,Chen2023-GenePT}. 
While this choice restricts inference to markers observed during training, as per marker sequence- or gene-level similarity do not imply similar tissue-level expression patterns it lets the model learn marker-specific convolutional operators directly from IMC measurements (for detailed explanation see Appendix~\ref{app:marker-encodings-rationale}).

\paragraph{Training procedure} 
\kg{All backbone weights and marker embeddings are randomly initialized and trained end-to-end with a single masked-reconstruction objective. Stem and objective ablations are described in Appendix~\ref{app:ablation-results}.}
We trained all models for 200 epochs with AdamW~\cite{loshchilov2019decoupled} optimizer using weight decay $10^{-4}$ and a cosine annealing learning-rate schedule with linear warm-up (for full details see Appendix~\ref{sec:hyperparams}). \kg{For new panels with unseen markers, one can either skip markers not observed during the pretraining or train only their encoder and decoder embeddings with the remaining network frozen (Appendix~\ref{app:external-panels}).}

\paragraph{Preprocessing Overview}
For training \name model family,  we preprocess raw IMC images with variance-stabilizing $\text{arcsinh}(x/5)$ transformation, frequency-based Butterworth \cite{Butterworth1930} denoising, and intensity normalization, and then train/evaluate on fixed-size crops with standard spatial augmentations; full procedural details and all preprocessing parameters are given in Appendix~\ref{app:preprocessing}.

Because this pipeline differs from previously used IMC preprocessing workflows~\cite{virtues,Liu2025-EVA}, we include a dedicated signal-stability analysis in Appendix~\ref{app:signal_stability_analysis}, showing that our preprocessing excels at preserving meaningful marker variation while limiting technology-specific noise.

\subsection{Experiments}

We evaluate $\name$ variants across: (i) virtual staining and uncertainty estimation, (ii) representation learning for cell typing and clinical prediction, and (iii) computational efficiency. We benchmark primarily against VirTues~\citep{virtues}, which reports state-of-the-art performance for IMC data, and EVA~\citep{Liu2025-EVA}, which further reports improvements over VirTues on selected tasks. We use the publicly released VirTues checkpoint from Hugging Face\footnote{\url{https://huggingface.co/bunnelab/virtues}, snapshot as of December 9, 2025.} and the EVA checkpoint released in the authors' GitHub repository. \kg{Our attempts to train KRONOS v1~\citep{KRONOS2025} did not converge (Appendix~\ref{app:kronos-reproduction}).}

To disentangle the effects of training data and preprocessing, we train variants across different datasets and preprocessing pipelines. Model names containing \texttt{IMC17M} denote training on \datasetNameNoSpace. The suffix \texttt{OP} denotes our preprocessing, whereas \texttt{VP} denotes the VirTues preprocessing pipeline. For models without these suffixes, the original dataset and preprocessing of the corresponding method are used. 
\kg{ We evaluate EVA trained on \datasetName using \texttt{OP} and the released checkpoint evaluated with our preprocessing (the original IMC preprocessing was not fully specified). We did not train $\name$ on the VirTues dataset because some of their datasets were not publicly available (see Appendix~\ref{app:virtues-datasets}).}

\subsubsection{Virtual Staining}

\paragraph{Setup.}

To evaluate virtual staining in an out-of-cohort setting, \kg{we use ten held-out test panels}. %
Because $\name$ and VirTues use different preprocessing pipelines, we train \kg{$\name$ variants for out-of-cohort evaluation} and VirTues on \datasetName \kg{with all test data excluded}, using our preprocessing pipeline (\texttt{OP}). %
We use \texttt{OP} for the main comparison because \texttt{VP} requires statistics computed from the unseen test cohort, which violates the out-of-cohort setting (Appendix~\ref{app:panel-wise-scaling}). For completeness, we also train $\name$ variants with \texttt{VP} preprocessing and report HN results in Appendix~\ref{app:virtual-staining-vp}. For fair comparison with VirTues, 
we exclude DNA1/DNA2, which are absent from the VirTues training panel as they lack appropriate ESM embeddings
.

For each test image, we predict each of the $M$ target markers in a leave-one-marker-out setting. We report pixel-level MSE averaged across all images and markers, and Pearson correlation computed per marker across all images and then averaged over markers (for details see Appendix~\ref{app:virtual-staining-pearson-rationale}).
\paragraph{Results.}
\begin{table}[ht]
\centering\small
\caption{\kg{Virtual staining on ten test panels: MSE/Pearson. VirTues and EVA denote their IMC17M+OP variants with overlap-only inputs (full-panel comparison: Table~\ref{tab:full-panel-inputs}). Bold marks the lowest MSE and highest Pearson per panel, including ties at the reported precision.}}
\label{tab:virtual_staining_accuracy}
\begin{tabular}{lccccc}
\toprule
\kg{Panel} & \kg{EVA} & \kg{$\nameViT$} & \kg{$\nameConv$} & \kg{$\nameSwin$} & \kg{VirTues}\\
\midrule
\kg{Head\&Neck, Immucan~\cite{immucan_2025}} & \kg{.180/.51} & \kg{.081/.72} & \kg{.068/.77} & \kg{\textbf{.061}/\textbf{.78}} & \kg{.101/.73}\\
\kg{Allam et al.~\cite{Allam2022-imc}, Panel 1} & \kg{.067/.50} & \kg{.045/.64} & \kg{.037/\textbf{.66}} & \kg{.044/\textbf{.66}} & \kg{\textbf{.035}/.65}\\
\kg{Allam et al.~\cite{Allam2022-imc}, Panel 2} & \kg{.462/.47} & \kg{.287/.56} & \kg{.217/.62} & \kg{.187/.59} & \kg{\textbf{.185}/\textbf{.65}}\\
\kg{Hu et al.~\cite{Hu2023-imc}, Panel 1} & \kg{.108/.18} & \kg{.113/.32} & \kg{.092/.41} & \kg{.081/\textbf{.45}} & \kg{\textbf{.056}/.34}\\
\kg{Hu et al.~\cite{Hu2023-imc}, Panel 2} & \kg{.487/.47} & \kg{.457/.48} & \kg{\textbf{.427}/\textbf{.61}} & \kg{.531/.45} & \kg{.455/.56}\\
\kg{Hu et al.~\cite{Hu2023-imc}, Panel 3} & \kg{.021/.54} & \kg{.010/.58} & \kg{.017/.53} & \kg{\textbf{.005}/.55} & \kg{.011/\textbf{.65}}\\
\kg{Moldoveanu~\cite{Moldoveanu2022-imc}, Panel 1} & \kg{.361/.39} & \kg{.143/.55} & \kg{\textbf{.137}/\textbf{.57}} & \kg{.173/.53} & \kg{.201/.54}\\
\kg{Moldoveanu~\cite{Moldoveanu2022-imc}, Panel 2} & \kg{.302/.21} & \kg{.192/.35} & \kg{\textbf{.120}/\textbf{.43}} & \kg{.180/.41} & \kg{.185/.29}\\
\kg{Schulz~\cite{Schulz2018-imc}, Panel 1} & \kg{.470/.48} & \kg{.344/.66} & \kg{.475/.67} & \kg{\textbf{.282}/\textbf{.68}} & \kg{.346/.54}\\
\kg{Schulz~\cite{Schulz2018-imc}, Panel 2} & \kg{.341/.58} & \kg{.248/.63} & \kg{.267/\textbf{.69}} & \kg{.215/.64} & \kg{\textbf{.167}/\textbf{.69}}\\
\bottomrule
\end{tabular}
\end{table}

\kg{
As shown in Table~\ref{tab:virtual_staining_accuracy}, $\name$ variants achieve the lowest MSE on six of the ten panels, with $\nameConv$ and $\nameSwin$ leading on three panels each, while VirTues leads on the remaining four. $\name$ variants also achieve the highest Pearson correlation on eight panels, including a tie with VirTues on Schulz Panel 2 at the reported precision, supporting the preservation of marker-specific expression patterns. These results suggest that convolutional and window-attention pan-marker encoders are well suited to the local spatial structure of IMC images. Paired Wilcoxon signed-rank tests on image-marker MSE with Benjamini-Hochberg FDR correction (significance at $q<0.05$) yield wins/losses/non-significant differences versus VirTues of 6/1/3 for $\nameSwin$ and 3/3/4 for both $\nameConv$ and $\nameViT$ (Appendix~\ref{app:external-panels}). Further per-marker HN results are provided in Appendix~\ref{app:virtual-staining-op}.
} 

\subsubsection{Uncertainty estimation}
\label{sec:uncerainty_results}

\paragraph{Setup.}
To evaluate uncertainty fidelity, we use the heteroscedastic Gaussian head of the \name architecture, which 
encodes the predictive uncertainty through a log-variance estimate $\log\sigma^{2}$ for each reconstructed channel.
We assess whether the predicted uncertainty is reliably \emph{calibrated} in the sense of tracking reconstruction error.

We perform this analysis 
\kg{on all 40 markers of the IMMUcan Head \& Neck (HN) cohort using the leave-one-marker-out setup}
and evaluate it using Pearson correlation between log variance and reconstruction error, expected calibration error (ECE) \citep{Kuleshov2018}, and area under sparsification error (AUSE) \citep{Ilg2018, Poggi2020}, which measures how closely the predicted $\sigma$ ranks pixels by error compared to an oracle ranking. For details see Appendix~\ref{app:uncertainty_calibration}.

\kg{We also assess whether \nameNoSpace-predicted uncertainty can identify unreliable cell-level readouts from virtual staining, using PD-1 positivity in CD8+ cells as an example. In a retrospective analysis of 21,058 segmented CD8+ cells from IMMUcan NSCLC2 panel 1, we predict PD-1 leave-one-out with $\nameConv$ and compare predicted positivity with reference labels derived from measured PD-1 expression. For the details see Appendix~\ref{app:pd1}.}

\begin{wraptable}{r}{0.47\textwidth}
\vspace{-1.0em}
\centering
\caption{
Uncertainty calibration \kg{evaluation on the HN cohort}, measured by Pearson correlation (\kg{predicted uncertainty}
vs reconstruction error), expected calibration error (ECE), and area under sparsification error (AUSE, RMSE-base).
\textbf{Bold} denotes the best score; ECE values are equal at the reported precision.
}
\label{tab:uncertainty_calibration}
\begin{small}
\setlength{\tabcolsep}{3pt}
\begin{tabular}{lccc}
\toprule
Model & Pearson ($\uparrow$) & ECE ($\downarrow$) & AUSE ($\downarrow$) \\
\midrule
$\nameViT$  & .922 & .029 & .0135 \\
$\nameConv$ & \textbf{.941} & .029 & .0132 \\
$\nameSwin$ & .934 & .029 & \textbf{.0112} \\
\bottomrule
\end{tabular}
\end{small}
\end{wraptable}
\paragraph{Results.}

Table~\ref{tab:uncertainty_calibration} shows that all $\name$ variants produce uncertainty estimates that closely track reconstruction error, with Pearson correlations above .92. $\nameConv$ leads on Pearson (.941), while $\nameSwin$ achieves the lowest AUSE (.0112), indicating that its predicted $\sigma$ ranks pixels closest to the oracle ranking by true error. Although all variants obtain the same low ECE (.029), their expected calibration curves differ (see Figure~\ref{fig:app_reliability}). These results indicate that the heteroscedastic head provides a meaningful per-pixel confidence signal rather than an auxiliary output disconnected from reconstruction quality. For further uncertainty analyses see Appendix Figures~\ref{fig:app_loo_scatter}-\ref{fig:app_sparsification_mae}. \kg{Cell-level uncertainty distinguishes incorrect from correct PD-1 positivity calls with an ROC-AUC of .78. Error rates are 8.4\% and 77\% in the lowest- and highest-uncertainty deciles, respectively, showing that classification errors are concentrated among cells assigned higher uncertainty.}

\subsection{Representation Learning: Cell Typing}

\begin{wraptable}{r}{0.50\textwidth}
\vspace{-1.0em}
\centering
\caption{
Cell-typing Macro-F1 from 10-fold cross-validation.
All standard deviations were below $10^{-2}$.
For VirTues, values report the better score from either the original or \name{} cell-typing procedure (see Appendix~\ref{app:phenotyping-methods}).
\underline{Underline} denotes the best score within a model family; \textbf{bold} denotes the best score per endpoint.
}
\label{tab:cell_phenotyping_agg_f1}
\small
\begin{threeparttable}
\begin{tabular}{lcccc}
\toprule
Model & ~\citep{Eling2025} & \cite{Danenberg2022} & \cite{Cords2024} & Avg. \\
\midrule
\kg{$\text{mean-marker}_{\texttt{raw}}$} & \kg{.715} & \kg{.535} & \kg{.684} & \kg{.645} \\
\kg{$\text{mean-marker}_{\texttt{OP}}$} & \kg{\underline{.778}} & \kg{.563} & \kg{.705} & \kg{\underline{.682}} \\
\kg{$\text{mean-marker}_{\texttt{VP}}$} & \kg{.762} & \kg{\underline{.565}} & \kg{\underline{.706}} & \kg{.678} \\
\midrule
$\text{EVA}_{\texttt{OP}}$ & .602 & .510 & .691 & .601 \\
\kg{$\text{EVA}_{\texttt{IMC17M+OP}}$} & \kg{\underline{.694}} & \kg{\underline{.551}} & \kg{\underline{.697}} & \kg{\underline{.647}} \\
\midrule
$\text{VirTues}_{\texttt{}}$      & \underline{.726} & .569* & .711 & \underline{.669} \\
$\text{VirTues}_{\texttt{IMC17M}}$    & .703 & .555* & .712 & .657 \\
$\text{VirTues}_{\texttt{IMC17M+OP}}$ & .706 & \underline{.579*} & \underline{.716} & .667 \\
\midrule
$\nameViT$  & .753 & .592 & \underline{\textbf{.751}} & .699 \\
$\nameConv$ & \underline{\textbf{.790}} & \underline{\textbf{.592}} & .745 & \underline{\textbf{.709}} \\
$\nameSwin$ & .762 & .582 & .738 & .694 \\
\bottomrule
\end{tabular}
\begin{tablenotes}
\item[] * \scriptsize Original VirTues cell-typing procedure used.
\end{tablenotes}
\end{threeparttable}
\vspace{-1.0em}
\end{wraptable}

\paragraph{Setup}

To evaluate the quality of learned single-cell representations, we use three datasets: IMMUcan Phenotyping~\citep{Eling2025}, with expert-curated manual cell-type annotations (kept as a hold-out validation set), and Danenberg et al.~\citep{Danenberg2022} and Cords et al.~\citep{Cords2024}, that provide annotations based on unsupervised clustering. 
We benchmark $\name$ variants against $\text{EVA}_{\texttt{OP}}$, the original VirTues model, and VirTues variants trained on \datasetName with both \texttt{VP} ($\text{VirTues}_{\texttt{IMC17M}}$) and \texttt{OP} ($\text{VirTues}_{\texttt{IMC17M+OP}}$) preprocessing pipelines. \kg{Additionally, we include a mean-marker baseline, representing each segmented cell by its mean expression for each measured marker under raw, \texttt{OP}, or \texttt{VP} preprocessing. This tests whether learned image representations improve over cell-averaged marker expression.}

Single-cell patch extraction and embedding construction are described in Appendix~\ref{app:celltyping}. As our extraction procedure differs from that of VirTues, we evaluate all VirTues variants under both protocols and report the better of the two scores. 
We train a multinomial logistic regression on single-cell embeddings and report per-cell-type F1 scores obtained via 10-fold cross-validation.

\kg{To assess out-of-cohort transfer, we additionally evaluate the frozen $\nameConv$ encoder on the two Moldoveanu et al.~\cite{Moldoveanu2022-imc} panels without new-marker adaptation, comparing it with released VirTues (Appendix~\ref{app:external-panels}).}

\paragraph{Results}

Table~\ref{tab:cell_phenotyping_agg_f1} shows that $\nameConv$ achieves the best average Macro-F1 (.709), outperforming \kg{the strongest mean-marker baseline (.682),} the strongest VirTues variant (.669), and retrained EVA (.647). 
For the mean-marker baseline, OP and VP increase average Macro-F1
from .645 on raw inputs to .682 and .678, respectively.
This shows that preprocessing improves the predictive value of
cell-averaged marker expression even without a learned image
encoder, providing a stronger reference for evaluating the
pretrained representations.
On Eling, mean-marker OP (.778) exceeds $\nameViT$ (.753) and $\nameSwin$ (.762), while $\nameConv$ reaches .790. All three $\name$ variants exceed the mean-marker baselines on Danenberg, Cords, and on average.
\kg{On Moldoveanu, $\nameConv$ without new-marker adaptation achieves Macro-F1 scores of .765 on panel P1 and .643 on panel P2, compared with .734 and .617, respectively, for released VirTues, whose pretraining included this cohort (Table~\ref{tab:moldoveanu-adaptation}). \kg{These results support transfer of $\nameConv$ representations to a cohort absent from pretraining, without adapting the encoder.}}

\WFclear
\subsection{Representation Learning: Clinical predictions}

\paragraph{Setup}
We further evaluate whether $\name$ variants learn representations that transfer to patient-level clinical prediction. 
We consider two published IMC cohorts with clinical labels~\citep{Danenberg2022,Cords2024}. 
For each cohort, we extract patch embeddings from the encoder, aggregate them to the patient level by mean pooling, and train logistic regression and ABMIL~\citep{ABMIL2018} for each clinical endpoint. 
We report Macro-F1 across 10-fold cross-validation splits as mean$\pm$standard deviation, and compare against VirTues variants, $\text{EVA}_{\texttt{OP}}$, \kg{and $\text{EVA}_{\texttt{IMC17M+OP}}$} under the same protocol. 
\kg{We also include mean-marker baselines using raw, OP, and VP preprocessing. Appendix~\ref{app:clinical_results_under_vp} details their construction and statistical testing, and reports ABMIL results.}

\paragraph{Results}
Table~\ref{tab:clinical_representation} shows that $\name$ variants provide \kg{competitive} patient-level representations \kg{for clinical prediction}. 
$\nameSwin$ achieves the best average Macro-F1 (.641), followed by $\nameConv$ (.638),  \kg{compared with .627 for the strongest VirTues variant and .608 for the strongest mean-marker baseline.
Released and retrained EVA achieve .585 and .543, respectively.}
\kg{
The VP mean-marker baseline achieves the highest displayed PAM50 mean (.43).}
\kg{The differences between $\name$ and VirTues are not statistically significant on any of the seven endpoints (all $q\geq .34$).}
These results indicate that marker-adaptive pretraining yields representations that remain informative for clinical predictions under minimal downstream supervision. 
Appendix~\ref{app:clinical_results_under_vp} provides extended results and discusses the information limits of these endpoints.

\begin{table}[ht]
\caption{Representation learning for clinical endpoints (Macro-F1; higher is better).
Values report mean$\pm$std Macro-F1 across 10-fold cross-validation; \kg{bold marks the highest displayed mean, including rounded ties.}}

  \label{tab:clinical_representation}
  \centering
  \begin{small}
  \begin{threeparttable}
    \begin{tabular}{lcccccccc}
      \toprule
      {Model} &
        \multicolumn{4}{c}{Danenberg et al.~\cite{Danenberg2022}}
        & \multicolumn{3}{c}{Cords et al.~\cite{Cords2024}} \\
      \cmidrule(lr){2-5}\cmidrule(lr){6-8}
     & PAM50 & Grade & ER & ERBB2 & Subtype & Relapse & Grade & Avg.\\
      \midrule

\kg{$\text{mean-marker}_{\texttt{raw}}$}
& \kg{.41±.06} & \kg{.47±.05} & \kg{.77±.08}
& \kg{.75±.11} & \kg{.77±.03} & \kg{.56±.02}
& \kg{.49±.05} & \kg{.602}\\

\kg{$\text{mean-marker}_{\texttt{OP}}$}
& \kg{.42±.06} & \kg{.49±.06} & \kg{.76±.09}
& \kg{.75±.11} & \kg{.78±.03} & \kg{.56±.02}
& \kg{.49±.04} & \kg{.606}\\

\kg{$\text{mean-marker}_{\texttt{VP}}$}
& \kg{\textbf{.43±.05}} & \kg{.47±.05} & \kg{.75±.09}
& \kg{.77±.10} & \kg{.79±.03} & \kg{.55±.03}
& \kg{.49±.05} & \kg{.608}\\
\midrule
      {$\text{EVA}_{\texttt{OP}}$} &.38±.05 & .50±.08 & .75±.07 & .61±.08 & .79±.03 & .57±.05 & .50±.05 & .585 \\
      \kg{{$\text{EVA}_{\texttt{IMC17M+OP}}$}} & \kg{.34±.06} & \kg{.43±.05} & \kg{.68±.05} & \kg{.47±.002} & \kg{.79±.03} & \kg{.58±.04} & \kg{.51±.08} & \kg{.543} \\
      \midrule
      {$\text{VirTues}$} & .37±.05 & \textbf{.53±.05} & .78±.08 & .76±.11 & .82±.03 & .57±.03 & .51±.07 & .620 \\

      {$\text{VirTues}_{\texttt{IMC17M}}$}  & .36±.06 & .49±.05 & .77±.06 & .78±.13 & \textbf{.83±.02} & .58±.04 & .54±.06 & .621\\

      {$\text{VirTues}_{\texttt{IMC17M+OP}}$}  & .40±.08 & .49±.08 & .80±.06 & .77±.13 & \kg{\textbf{.83±.03}} & .57±.05 & .53±.09 & .627\\
      \midrule
      {$\nameViT$} & .40±.06 & .48±.05 & \kg{\textbf{.81±.04}} & .77±.12 & .82±.03 & .58±.04 & .54±.06 & .628\\

    {$\nameConv$} & .40±.09 & .52±.08 & \textbf{.81±.07} & \kg{\textbf{.80±.13}} & .81±.03 & .59±.03 & .54±.05 & .638\\

    {$\nameSwin$} & {.42±.09} & .49±.06 & .79±.05 & \textbf{.80±.13} & \kg{\textbf{.83±.03}} & \textbf{.61±.04} & \textbf{.55±.06} & \textbf{.641}\\

      \bottomrule
    \end{tabular}
  \end{threeparttable}
  \end{small}
\end{table}

\subsection{Computational efficiency}
\paragraph{Setup.}
We compare the inference-time computational cost of \name variants, VirTues, and $\text{EVA}_{\texttt{OP}}$ on a single forward pass for an IMC image of shape $[40,128,128]$. 
We report active parameter count, GFLOPs, mean$\pm$std inference time in milliseconds, and peak activation memory, all obtained on a single NVIDIA RTX 5000 Ada Generation GPU aggregated over 100 random input runs.

\paragraph{Results.}
 \begin{wraptable}{r}{0.60\textwidth}
\vspace{0em}
\centering
\caption{Computational efficiency.
Inference cost measured on 100 random inputs of shape $[40,128,128]$ on a single NVIDIA RTX 5000 Ada Generation GPU.
}
\label{tab:efficiency}
\begin{small}
\resizebox{0.60\textwidth}{!}{%
\begin{tabular}{lcccc}
\toprule
Model & Active param. & GFLOPs & Time [ms] & Mem. [GB] \\
\midrule
$\nameConv$ & 44.6M & \textbf{89.9} & 13.4$\pm$0.2 & 1.01 \\
$\nameSwin$ & 57.0M & 138.8 & 18.2$\pm$0.9 & 1.03 \\
$\nameViT$ & 105.2M & 141.5 & \textbf{11.1$\pm$0.3} & 1.39 \\
\midrule
VirTues & \textbf{42.5M} & 1049.3 & 28.5$\pm$0.5 & \textbf{0.45} \\
\midrule
$\text{EVA}$ & 122.4M & 3590.0 & 45.8$\pm$0.3 & 0.95 \\
\bottomrule
\end{tabular}%
}
\end{small}
\end{wraptable}
Table~\ref{tab:efficiency} shows that $\name$ variants operate in a substantially lower-compute time regime than prior foundation models for IMC. $\nameConv$ requires 89.9 GFLOPs and 13.4$\pm$0.2 ms per forward pass, compared with 1049.3 GFLOPs and 28.52$\pm$0.5 ms for VirTues, and 3590 GFLOPs and 45.82$\pm$0.3 ms for $\text{EVA}_{\texttt{OP}}$. Thus, $\nameConv$ reduces compute by more than an order of magnitude relative to VirTues and nearly 40-fold relative to EVA. These results transfer to being faster in wall-clock inference.
Among $\name$ variants, $\nameViT$ is the fastest in runtime (11.1$\pm$0.3 ms) but uses the largest parameter count and activation memory within the family (105.2M parameters, 1.39 GB). $\nameConv$ provides the most efficient overall trade-off, with the fewest parameters among $\name$ variants (44.6M), the lowest GFLOPs, and lower activation memory than $\nameViT$. $\nameSwin$ remains substantially more efficient than VirTues and EVA, while achieving strong virtual staining accuracy. \kg{Overall, these results combine strong cell typing and virtual staining with lower computational cost, while clinical prediction remains competitive with VirTues~\cite{virtues, Liu2025-EVA}.}

\section{Discussion}

$\name$ addresses a central practical constraint of IMC, namely the absence of a fixed channel marker space, by introducing marker-adaptive hyperconvolutions that generate convolutional kernels from learned marker embeddings. 
Such \kg{a} design enables a single foundation model to operate on any \kg{subset of markers seen during pretraining} without re-training and architectural changes. 
\kg{Virtual staining results across ten held-out test panels demonstrate out-of-cohort transfer. Across the cell-typing benchmarks, all $\name$ variants exceed the evaluated baselines on average, while clinical prediction remains competitive with VirTues.} 
\kg{The locality analysis suggests that} pan-marker representations enabled by \kg{the} hyperconvolutional architecture preserve local spatial contexts and capture short-range dependencies \kg{relevant to} tissue architecture and cellular interaction (see Appendix~\ref{app:locality}).
Beyond that, the heteroscedastic objective provides per-pixel uncertainty estimates that closely track reconstruction error, offering a practical reliability signal for downstream applications (e.g., prioritizing regions or markers requiring caution).

\noindent
\kg{\textbf{Model choice.}
$\nameConv$ offers the strongest average cell-typing performance and lowest computational cost within the family. $\nameSwin$ is preferable when reconstruction fidelity and uncertainty ranking are priorities, while $\nameViT$ offers the lowest inference latency at the cost of higher parameter count and activation memory.}

\noindent
\textbf{Limitations.} While the proposed design is computationally efficient and cohort-deployable, several limitations remain. 
\kg{First, markers absent from pretraining require learning new marker embeddings before they can be encoded or virtually stained.}
\kg{Second,} the training distribution is constrained by marker availability and dataset bias; rare biomarkers and under-represented tissue contexts may yield less reliable reconstructions, especially under \kg{out-of-cohort transfer.}
\kg{Third,} our likelihood model assumes conditional independence across pixels and channels, which is a pragmatic approximation but may under-represent structured noise or cross-channel dependencies in IMC.
\kg{Finally, the clinical cohorts are small, and the observed differences from Virtues are not statistically significant.}

\noindent
\textbf{Future overlook.} $\name$ could be extended to additional multiplex modalities (e.g., MIBI, CODEX), incorporate richer priors or structured uncertainty (e.g., spatially correlated noise), and develop more explainable representations characterizing cross-marker dependencies.
\kg{These extensions may require modality-specific preprocessing, as the current pipeline is designed and challenged with IMC.}
If IMC adoption reaches clinical practice, \name could support diagnostic pipelines through virtual staining of missing markers and calibrated uncertainty flagging low-confidence predictions.
Overall, our results suggest that operator-level channel adaptivity is an effective recipe for building practical foundation models in multiplex imaging, combining panel flexibility, efficiency, and reliability-aware predictions within a single architecture.

\clearpage
\begin{ack}

\paragraph{Funding.}
This project was supported by the IMMUcan project that has received funding from the Innovative Medicines Initiative 2 Joint Undertaking under grant agreement no. 821558. This Joint Undertaking receives support from the European Union’s Horizon 2020 Research and Innovation Programme and EFPIA (https://IMI.europa.eu).
This project has received funding from the European Union’s Horizon Europe research and innovation programme under grant agreement No. 101136552 (SPACETIME). 
This work was supported by the Polish National Science Centre SONATA BIS grant No. 2020/38/E/NZ2/00305 
and SONATA grant No. 2023/51/D/NZ2/02892.

\paragraph{Computational resources.}
We gratefully acknowledge Polish high-performance computing infrastructure PLGrid (HPC Center: ACK Cyfronet AGH) for providing computer facilities and support within computational grants no. PLG/2025/018195 and PLG/2026/019387.

\paragraph*{Conflicts of Interest.}
Nothing to declare.

\paragraph{Disclaimer.}
Funded by the European Union. Views and opinions expressed are however those of the authors only and do not necessarily reflect those of the European Union or the European Health and Digital Executive Agency (HaDEA). Neither the European Union nor the granting authority can be held responsible for them.

\end{ack}

\bibliographystyle{unsrt}
\bibliography{references} %

\newpage
\appendix

\section{Appendix}

\subsection{Sampling and masking procedures for Masked Modelling task}
\label{app:masking_details}

Each training sample consists of a full measured panel
$(\mathbf{X}_{\mathcal{I}_{\mathrm{img}}}, \mathcal{I}_{\mathrm{img}})$ with
$\mathbf{X}_{\mathcal{I}_{\mathrm{img}}}\in\mathbb{R}^{C_{\mathrm{im}}\times H\times W}$ where
$C_{\mathrm{img}} = |\mathcal{I}_{\mathrm{img}}|$.
We construct minibatches using a panel-grouped sampler, so all samples within a minibatch share $C_{\mathrm{img}}$ (the same panel size) using the following steps:
(i) target-set subsampling, (ii) full-channel dropout within the targets, and
(iii) patch-wise spatial masking of the remaining inputs.

\paragraph{(i) Target-set subsampling.}
For each minibatch we first sample a target size
\[
K \sim \mathrm{Unif}\Big(\{\lceil \alpha C_{\mathrm{img}}\rceil,\dots,C_{\mathrm{img}}\}\Big),
\]
where $\alpha$ by default is set to 0.75 and, for each sample, we choose $K$ markers uniformly without replacement from $\mathcal{I}_{\mathrm{img}}$
to form $\mathcal{I}_{\mathrm{tgt}}$ and the corresponding tensor
$\mathbf{X}_{\mathcal{I}_{\mathrm{tgt}}}\in\mathbb{R}^{K\times H\times W}$. Each sample in minibatch has its $K$ markers drawn independently.

\paragraph{(ii) Full-marker dropout within targets.}
Given $K$, we drop a non-empty subset of target markers by sampling
\[
M \sim \mathrm{Unif}\Big(\{1,\dots,\lceil \beta K\rceil\}\Big),
\]
where $\beta$ by default is set to 0.5, and selecting $K-M$ markers uniformly without replacement from $\mathcal{I}_{\mathrm{tgt}}$ to form the encoder input set
$\mathcal{I}_{\mathrm{in}}\subset \mathcal{I}_{\mathrm{tgt}}$.
The dropped channels $\mathcal{I}_{\mathrm{tgt}}\setminus \mathcal{I}_{\mathrm{in}}$ are completely unobserved by the encoder, but are included in the reconstruction targets. Each sample in minibatch has its $K-M$ channels drawn independently. This setting simulates a virtual staining scenario, where models aims at prediction of the markers missing from panel.

\paragraph{(iii) Patch-wise spatial masking.}
On the remaining input channels $\mathcal{I}_{\mathrm{in}}$, we follow a masking protocol from \cite{convnext2} and additionally mask spatial patches by setting them to zero.
Let $p=\patchMaskSize{}$ and assume $p$ divides $H$ and $W$.
Define $(h,w)=(H/p,W/p)$ and sample a Bernoulli mask on the patch grid
\[
\mathbf{m}\in\{0,1\}^{|\mathcal{I}_{\mathrm{in}}|\times h\times w},
\qquad \mathbb{P}[\mathbf{m}_{c,u,v}=1]=\rho,
\]
where $\rho$ by default is set to 0.6. We then expand each entry of $\mathbf{m}$ to a $p\times p$ block to obtain
$\mathbf{M}\in\{0,1\}^{|\mathcal{I}_{\mathrm{in}}|\times H\times W}$ and define the masked encoder input
\[
\tilde{\mathbf{X}}_{\mathcal{I}_{\mathrm{in}}}
= \mathbf{X}_{\mathcal{I}_{\mathrm{in}}}\odot (1-\mathbf{M}),
\]
where $\odot$ denotes element wise multiplication.

\subsection{Stabilizing log-variance training using a gradient clamping}
\label{app:uncertainty_details}

Let $\ell=\mathbf{X}^{\log\sigma^2}_{\mathcal{I}_{\mathrm{tgt}}}$ denote the predicted log-variance. For numerical stability, we use the clamped log-variance
\[
\hat{\ell} = \mathrm{clamp}(\ell;a,b),
\qquad (a,b)=(-15,15).
\]
In the \emph{forward pass}, $\mathrm{clamp}$ returns the hard clip
$\mathrm{clip}(\ell;a,b)=\min(\max(\ell,a),b)$.
In the \emph{backward pass}, the gradient is propagated unchanged inside the interval and is smoothly downweighted outside it:
\[
\frac{\partial\,\mathrm{clamp}(\ell;a,b)}{\partial \ell}
=
\begin{cases}
1, & a<\ell<b,\\
1-\tanh^2(\ell), & \ell\le a\ \text{or}\ \ell\ge b.
\end{cases}
\]

For the $\beta$-NLL loss, we set $\beta=0.5$, following its original authors suggestion~\cite{seitzer2022on}.

\subsection{Dataset details}
\label{sec:dataset_details}

\begin{table}[t]
\centering
\caption{\textbf{Cohort-level composition of \datasetName{}}. The dataset spans 28 cohorts, 24,405 images, 265 markers, and over 17M patches. IMMUcan BC2/NSCLC2 (P1/2)/SCCHN1 datasets are IMMUcan \cite{immucan_2025} proprietary data, awaiting publication. The \emph{VirTues} column denotes with \checkmark cohorts included in the VirTues pretraining corpus.
}
\label{tab:datasets}
\scriptsize
\setlength{\tabcolsep}{3pt}
\begin{tabular}{lrrrlc}
\toprule
Dataset & Subjects & Images & Markers & Histopathology & VirTues \\
\midrule
ajaib~\cite{Ajaib2025} & 5 & 34 & 34 & Glioblastoma & \\
bengsch \cite{Sali2024} & 54 & 108 & 42 & Hepatocellular carcinoma & \\
bowen-lynch \cite{Bowen2023} & 18 & 85 & 38 & Lynch Syndrome colorectal mucosa & \\
cho \cite{Cho2025} & 12 & 140 & 41 & Pancreatic adenocarcinoma & \\
cords \cite{Cords2024} & 1070 & 2070 & 42 & Non-small cell lung cancer & \checkmark \\
cords-fibro \cite{Cords2023} & 12 & 110 & 42 & Breast cancer & \checkmark \\
damond \cite{Damond2019} & 12 & 845 & 36 & Type I diabetes pancreas & \checkmark \\
danenberg \cite{Danenberg2022} & 693 & 794 & 39 & Breast cancer & \checkmark \\
ehret-p1 \cite{ehret} & 52 & 56 & 42 & Organoid & \\
ehret-p2 \cite{ehret} & 52 & 52 & 44 & Organoid & \\
einhaus \cite{Einhaus2025} & 24 & 94 & 41 & Oral squamous cell carcinoma & \\
haley-glio \cite{Haley2024} & 8 & 24 & 40 & Glioblastoma & \\
hoch-protein \cite{Hoch2022} & 69 & 167 & 46 & Melanoma & \checkmark \\
hoch-rna \cite{Hoch2022} & 69 & 166 & 41 & Melanoma & \checkmark \\
IMMUcan BC2~\cite{immucan_2025}\textsuperscript{\dag} & 223 & 813 & 40 & Breast cancer & \\
IMMUcan NSCLC2 (P1)~\cite{immucan_2025}\textsuperscript{\dag} & 192 & 745 & 40 & Non-small-cell lung cancer & \\
IMMUcan NSCLC2 (P2)~\cite{immucan_2025}\textsuperscript{\dag} & 190 & 566 & 44 & Non-small-cell lung cancer & \\
IMMUcan SCCHN1~\cite{immucan_2025}\textsuperscript{\ddag,\dag} & 218 & 743 & 40 & Head and neck cancer & \\
IMMUcan Phenotyping~\cite{Eling2025}\textsuperscript{\ddag} & & & 40 & pan cancer & \checkmark \\
jackson-basel~\cite{Jackson2020} & 285 & 381 & 39 & Breast cancer & \checkmark \\
jackson-zurich~\cite{Jackson2020} & 72 & 365 & 39 & Breast cancer & \checkmark \\
kucukkose \cite{Kkkse2023}& 9 & 27 & 34 & Colorectal cancer & \\
meyer \cite{meyer2025} & 215 & 495 & 39 & Breast cancer & \\
ohara \cite{Ohara2024} & 7 & 58 & 28 & Urothelial carcinoma & \\
rigamonti \cite{Rigamonti2024} & 108 & 158 & 29 & Non-small-cell lung cancer & \checkmark \\
steenbuck-immune \cite{Steenbuck2025} & 88 & 7557 & 47 & Type I diabetes pancreas & \\
steenbuck-islets \cite{Steenbuck2025} & 88 & 7558 & 47 & Type I diabetes pancreas & \\
sussman \cite{Sussman2024} & 9 & 35 & 35 & Pancreatic adenocarcinoma & \\
xu \cite{Xu2022} & 26 & 158 & 37 & Melanoma & \\
\midrule
\textbf{TOTAL} & \textbf{3880} & \textbf{24405} & & & \\
\bottomrule
\end{tabular}
\begin{minipage}{0.95\textwidth}
\footnotesize
\kg{\textsuperscript{\dag} IMMUcan BC2, NSCLC2 P1, NSCLC2 P2, and SCCHN1 are IMMUcan consortium proprietary datasets awaiting publication.}\\
\kg{\textsuperscript{\ddag} IMMUcan SCCHN1 and Phenotyping were held out from training for all \name variants and VirTues and used exclusively for out-of-cohort evaluation.}
\end{minipage}
\end{table}

\datasetName is a curated aggregation of \numDatasets{} IMC datasets comprising \numPanels{} distinct marker panels, totalling \numImages{} images and \numMarkers{} unique markers across \numHistopathologies{} histologies. A detailed composition of used datasets is presented in Table~\ref{tab:datasets}.
Throughout, we use \emph{dataset} to denote a source study/cohort, and \emph{panel} to denote its measured marker configuration; multiple datasets may share a panel, and a single study may provide multiple panels (e.g., protocol variants).

The corpus spans diverse biological contexts, including tumour microenvironments (e.g., breast, lung, glioblastoma, colorectal, melanoma, oral squamous cell carcinoma, urothelial carcinoma), immune system characterisation, and tissue-specific studies (e.g., pancreas in Type~I diabetes and organoids). Panel sizes range from \minMarkers{} to \maxMarkers{} markers, with most panels in the \typicalMarkerRange{} range, reflecting the strong marker-set heterogeneity typical of IMC collections.

For all experiments, we partition images into train/test with a 4:1 ratio, stratified by panel to ensure that each panel contributes proportionally to both splits and to reduce confounding by panel-specific acquisition characteristics. Table~\ref{tab:datasets} reports per-dataset counts and histopathologies; datasets labelled IMMUcan (except IMMUcan Phenotyping~\cite{Eling2025}) are consortium data and are currently proprietary (awaiting publication).

\subsubsection{Notes on VirTues overlap.}\label{app:virtues-datasets}
VirTues was pretrained on 14 datasets~\cite{virtues}. Eight of these overlap with \datasetName{}, corresponding to ten \datasetName{} cohorts after cohort-level splitting: cords, cords-fibro, damond, danenberg, hoch-protein, hoch-rna, IMMUcan Phenotyping, jackson-basel, jackson-zurich, and rigamonti. The remaining six VirTues pretraining datasets (Allam et al.~\cite{Allam2022-imc}, Hu et al.~\cite{Hu2023-imc}, Moldoveanu et al.~\cite{Moldoveanu2022-imc}, Schultz et al.~\cite{Schulz2018-imc}, Wang et al.~\cite{Wang2023-imc}, and Zhu et al.~\cite{Zhu2025-imc}) are not part of \datasetName{}. In particular, Zhu et al.~\cite{Zhu2025-imc} dataset is behind DTA, while Wang et al.~\cite{Wang2023-imc} under protected zenodo access. Meyer et al.~\cite{meyer2025} was used by VirTues only for downstream transfer evaluation, not for pretraining. 

\subsection{\name Models configurations}\label{sec:model_conf}

\kg{In this section we introduce the detailed architecture of $\nameConvNoSpace$, $\nameSwin$, and $\nameViT$ models. We considered the following configurations for the masked modelling task:}
\begin{itemize}
    \item Marker-Agnostic stages $L_{ma}\in\{\texttt{None}, 1, 2\}$,
    \item Hyperkernel size $(h_{e}\times w_{e})\in\{(1\times 1), (3 \times 3)\}$
    \item Stage dimensionalities for $\texttt{CONV}$ and $\texttt{SWIN}$ variants $\in\{(192, 384, 768), (128, 256, 512)\}$,
    \item Hidden size for \texttt{VIT} variant $\in\{512, 768\}$,
\end{itemize}

The details of all selected architectures configurations are shown in Table~\ref{tab:model_parameters}.

\begin{table}[h]
\centering
\caption{\name models architecture and parameter specifications}
\label{tab:model_parameters}
\footnotesize
\begin{tabularx}{\linewidth}{@{}p{0.225\linewidth}p{0.25\linewidth}X@{}}
\toprule
Component & Parameter & Value / Specification \\ \midrule
\multicolumn{3}{l}{$\nameConv$} \\ \midrule
Encoder (Marker-Agnostic)
 & Stages ($L_{ma}$) & 1 \\
 & Stage composition & 4-fold Downsampling + 4 ConvNeXt v2 blocks \\
 & Stage dimension ($d_{ma}$) & 16 \\
 \midrule
Encoder (Pan-Marker)\textbf{ }
 & Hyperkernel size ($h_e \times w_e$) & $1 \times 1$ \\
 & Pan-Marker dimension ($d_{pm}$) & 192 \\
 & Stages ($L_{pm}$) & 3 \\
 & Stage composition & 4-fold Downsampling (skipped in first stage) + 4 ConvNeXt v2 blocks \\
 & Stage dimensionalities & 192, 384, 768 \\ \midrule
Latent feature map & Dimension ($d_{lat}$) & 768 \\
 & Shape ($H_{lat} \times W_{lat}$) & $16 \times 16$ \\ \midrule
Decoder & Hyperkernel size ($h_d \times w_d$) & $1 \times 1$ \\
 & Marker-Specific dimension ($d_{ms}$) & 512 \\
 & Marker-Agnostic decoder & 1 ConvNeXt v2 Block $(B_{dec} = 1)$  \\ \midrule

 \multicolumn{3}{l}{$\nameSwin$} \\ \midrule
Encoder (Marker-Agnostic)
 & Stages ($L_{ma}$) & 1 \\
 & Stage composition & 4-fold Downsampling + 4 Swin Transformer blocks \\
 & Stage dimension ($d_{ma}$) & 16 \\
 \midrule
Encoder (Pan-Marker)\textbf{ }
 & Hyperkernel size ($h_e \times w_e$) & $1 \times 1$ \\
 & Pan-Marker dimension ($d_{pm}$) & 192 \\
 & Stages ($L_{pm}$) & 3 \\
 & Stage composition & 4-fold Downsampling (skipped in first stage) + 4 Swin Transformer blocks \\
 & Stage dimensionalities & 192, 384, 768 \\ \midrule
Latent feature map & Dimension ($d_{lat}$) & 768 \\
 & Shape ($H_{lat} \times W_{lat}$) & $16 \times 16$ \\ \midrule
Decoder & Hyperkernel size ($h_d \times w_d$) & $1 \times 1$ \\
 & Marker-Specific dimension ($d_{ms}$) & 512 \\
 & Marker-Agnostic decoder & 1 ViT Block $(B_{dec} = 1)$  \\ \midrule

 \multicolumn{3}{l}{$\nameViT$} \\ \midrule
 Encoder (Marker-Agnostic) & Stages ($L_{ma}$) & 0 (Identity) \\
 & Dimension ($d_{ma}$) & 1 \\ \midrule
 Encoder (Pan-Marker) & Hyperkernel / Patch size ($h_e \times w_e$) & $8 \times 8$ (Stride 8) \\
 & Pan-Marker dimension ($d_{pm}$) & 768 \\
 & Layers & 16 \\
 & Attention heads & 8 \\ \midrule
Latent feature map & Dimension ($d_{lat}$) & 768 \\
 & Shape ($H_{lat} \times W_{lat}$) & $16 \times 16$ \\ \midrule
Decoder & Hyperkernel size ($h_d \times w_d$) & $1 \times 1$ \\
 & Marker-Specific dimension ($d_{ms}$) & 512 \\
 & Marker-Agnostic decoder & 1 ViT Block $(B_{dec} = 1)$ \\

\bottomrule
\end{tabularx}

\end{table}

\subsubsection{\kg{Stem and objective ablations}}
\label{app:ablation-results}
\kg{Table~\ref{tab:stem-ablation} varies the marker-agnostic stem depth of $\nameConv$ at $d_{lat}=768$. With no stem ($L_{ma}=0$), the hyperkernel acts directly on pixels and is equivalent to a fixed-channel convolution over the vocabulary with missing channels zero-filled. The one-stage stem improves both downstream scores over this control; a second stage lowers validation NLL without improving those scores. These results do not establish validation NLL alone as a model-selection rule.}
\begin{table}[ht]
\centering\small
\caption{\kg{Conv stem ablation. Cell typing is Macro-F1; lower validation NLL and staining MSE are better.}}
\label{tab:stem-ablation}
\begin{tabular}{lccc}
\toprule
\kg{$L_{ma}$} & \kg{Validation NLL} & \kg{Cell typing} & \kg{Staining MSE}\\
\midrule
\kg{0 (fixed-channel control)} & \kg{$-6.601$} & \kg{.758} & \kg{.0672}\\
\kg{1 (used)} & \kg{$-7.410$} & \kg{.790} & \kg{.0573}\\
\kg{2} & \kg{$-7.546$} & \kg{.788} & \kg{.0833}\\
\bottomrule
\end{tabular}
\end{table}
\kg{The matched objective comparison in Table~\ref{tab:objective-ablation} changes only the training objective. Plain MSE slightly improves Conv staining, whereas $\beta$-NLL improves its cell typing and clinical scores and provides the uncertainty output. 
}
\begin{table}[ht]
\centering\small
\caption{\kg{Conv Objective ablation. Cell typing and clinical prediction report Macro-F1.}}
\label{tab:objective-ablation}
\begin{tabular}{lcccc}
\toprule
\kg{Objective} & \kg{Staining MSE} & \kg{Pearson} & \kg{Cell typing} & \kg{Clinical}\\
\midrule
\kg{$\beta$-NLL} & \kg{.0573} & \kg{.800} & \kg{.790} & \kg{.637}\\
\kg{MSE} & \kg{.0501} & \kg{.806} & \kg{.768} & \kg{.632}\\
\bottomrule
\end{tabular}
\end{table}

\subsubsection{Marker-agnostic decoder architecture}

Let $\mathbf{U}_{j^c}$ be the output of the decoder Hyperconvolution operator $\hyperConvOpDec$ for target marker ${j^c}$.  Then $\headOp$ of the marker-agnostic decoder $\operatorname{Dec}_{\text{ma}}$ performs the following steps:

\begin{equation*}
\mathbf{D}_{j^c} =
\text{Blocks}^{(B_{dec})}(\mathbf{U}_{j^c})
\end{equation*}
\begin{equation*}
\mathbf{P}_{j^c} = \text{Conv}_{1 \times 1}(\mathbf{D}_{j^c}, 2\lambda^2)
\end{equation*}
\begin{equation*}
\hat{\mathbf{X}}_{j^c} = \text{PixelShuffle}(\mathbf{P}_{j^c})
\end{equation*}

where $\lambda = 8$ is the total upsampling factor. Since in the preprocessing stage of \datasetName we normalize marker intensities to a $[0,1]$ interval, we map

\[
\hat{\mathbf{X}}^0 =\mathrm{sigmoid}(\hat{\mathbf{X}}^0).
\]

\subsubsection{Decoder hyperkernel size}
All three \name variants use $1\times 1$ hyperkernels in the decoder ($h_d\times w_d = 1\times 1$). We evaluated larger spatial decoder kernels (e.g., $3\times 3$) but these configurations led to consistent training instability and divergence (larger decoder kernels substantially increased parameter count). No formal ablation is reported as no stable larger-kernel decoder was obtained.

\subsubsection{Comment on the $\nameViT$ architecture}

For $\nameViT$ we initially evaluated a configuration analogous to $\name$ that employed a one-stage Marker-agnostic encoder, however training of this configuration diverged.  Subsequently, we investigated a pixel-wise hyperkernel embedding strategy equivalent to the $L_{ma}=0$ configuration, followed by a standard patch tokenizer, but it also suffered from the convergence difficulties. Consequently, we adopted the minimal strategy with the hyperkernel acting directly as a multiplex tokenizer ($\operatorname{Enc}_{\text{ma}}$ was set as an identity), creating pan-marker patch embeddings from the original input.

\subsection{Marker-conditional kernel generators}
\label{app:marker-encodings-rationale}
For both encoder and decoder generators $\phi_{e}, \phi_{d}$ we first map markers to their vocabulary indices $i\in\mathcal{N}$. Then, each marker index is assigned to a standard learnable embedding of a flattened intended kernel shape (either $d_{\mathrm{pm}}\cdot d_{\mathrm{ma}}\cdot h_{e}\cdot w_{e}$ for encoder and $d_{\mathrm{ms}}\cdot d_{\mathrm{lat}}\cdot h_{d}\cdot w_{d}$ for decoder), subsequently reshaped to the final kernel dimensions.

\subsubsection{Caveats of protein language model-based embeddings}

Contrary to VirTues \cite{virtues} and EVA \cite{Liu2025-EVA}, \name uses randomly initialized, learnable marker embeddings trained from scratch instead of PLM-based embeddings \cite{Lin2023-ESM, Chen2023-GenePT}. This design choice is motivated by two main reasons:

\begin{itemize}
\item PLM embeddings are poorly defined for a substantial fraction of IMC markers; for example, DNA1/2, CD45RA+RO, and panCK lack unique UniProt entries or aggregate multiple proteins.
\item Sequence similarity does not imply spatial or functional similarity in IMC, with notable examples including CD45RA/CD45RO (which have nearly identical ESM / GenePT embeddings but an expression correlation of only 0.35) and CK subtypes.
\end{itemize}

To illustrate the latter, we tested VirTues zero-shot marker prediction on AMY1A, which is present in the Steenbuck dataset \cite{Steenbuck2025} included in IMC17M, but absent from VirTues' training data. We selected this particular marker as AMY2A, a closely related isoform with a similar ESM2 embedding ($L_{2}=0.154$) and high sequence similarity, was present in VirTues' training data. We further compared two conditions:

\begin{itemize}
\item \textbf{Zero-shot:} VirTues predicts AMY1A based solely on the ESM embedding (AMY1A was absent from the training data, but with similar AMY2A present).
\item \textbf{Trained:} $\text{VirTues}_{\texttt{IMC17M}}$, trained on IMC17M, with both AMY1A and AMY2A using VirTues preprocessing.
\end{itemize}

\begin{figure}[th]
  \begin{center}
    \centerline{\kg{\includegraphics[width=0.7\textwidth]{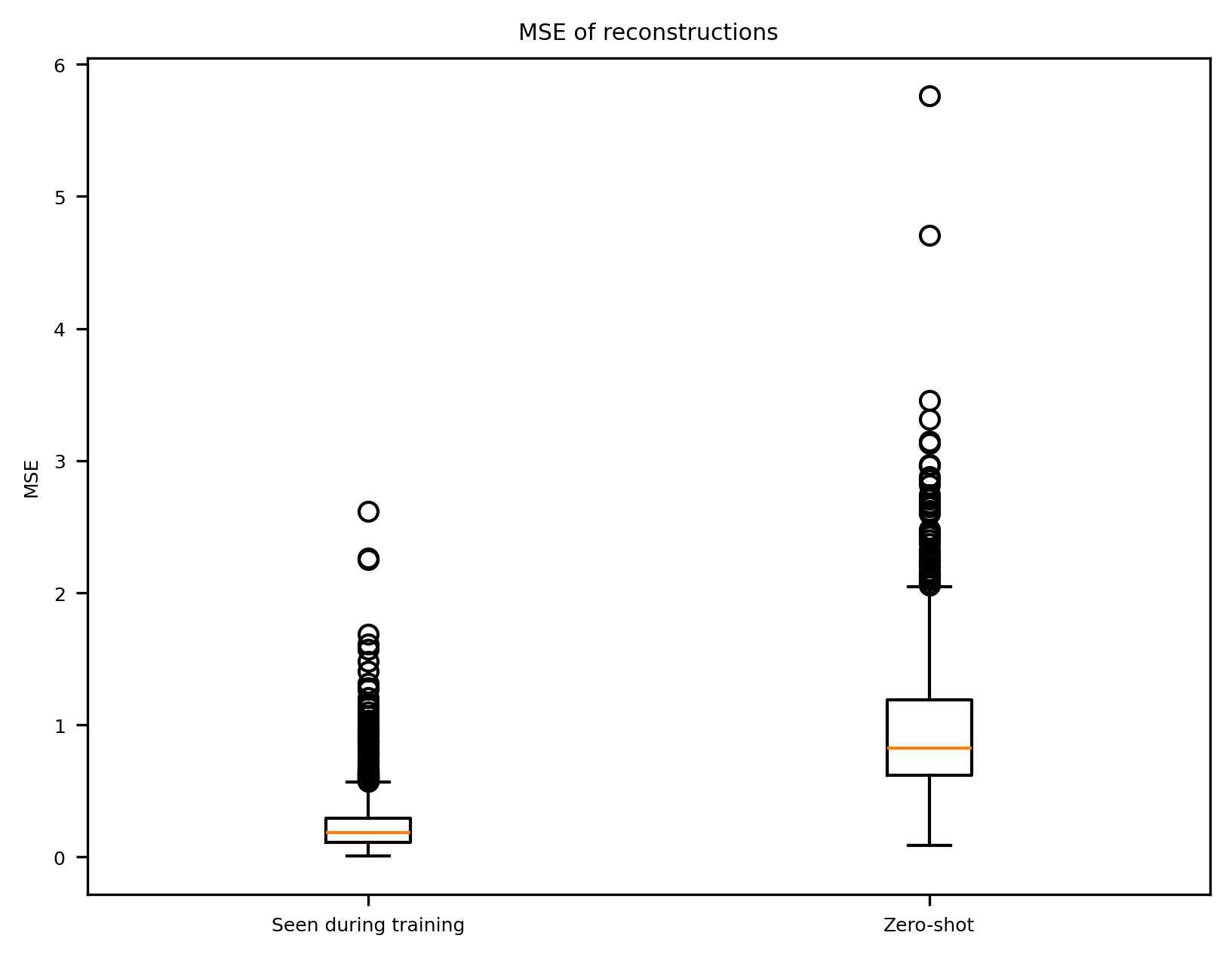}}
    }
    \caption{MSE of VirTues reconstructions in trained vs zero-shot settings.}
    \label{fig:mse-zero-shot}
  \end{center}
\end{figure}

\begin{figure}[th]
  \begin{center}
    \centerline{\kg{\includegraphics[width=0.7\textwidth]{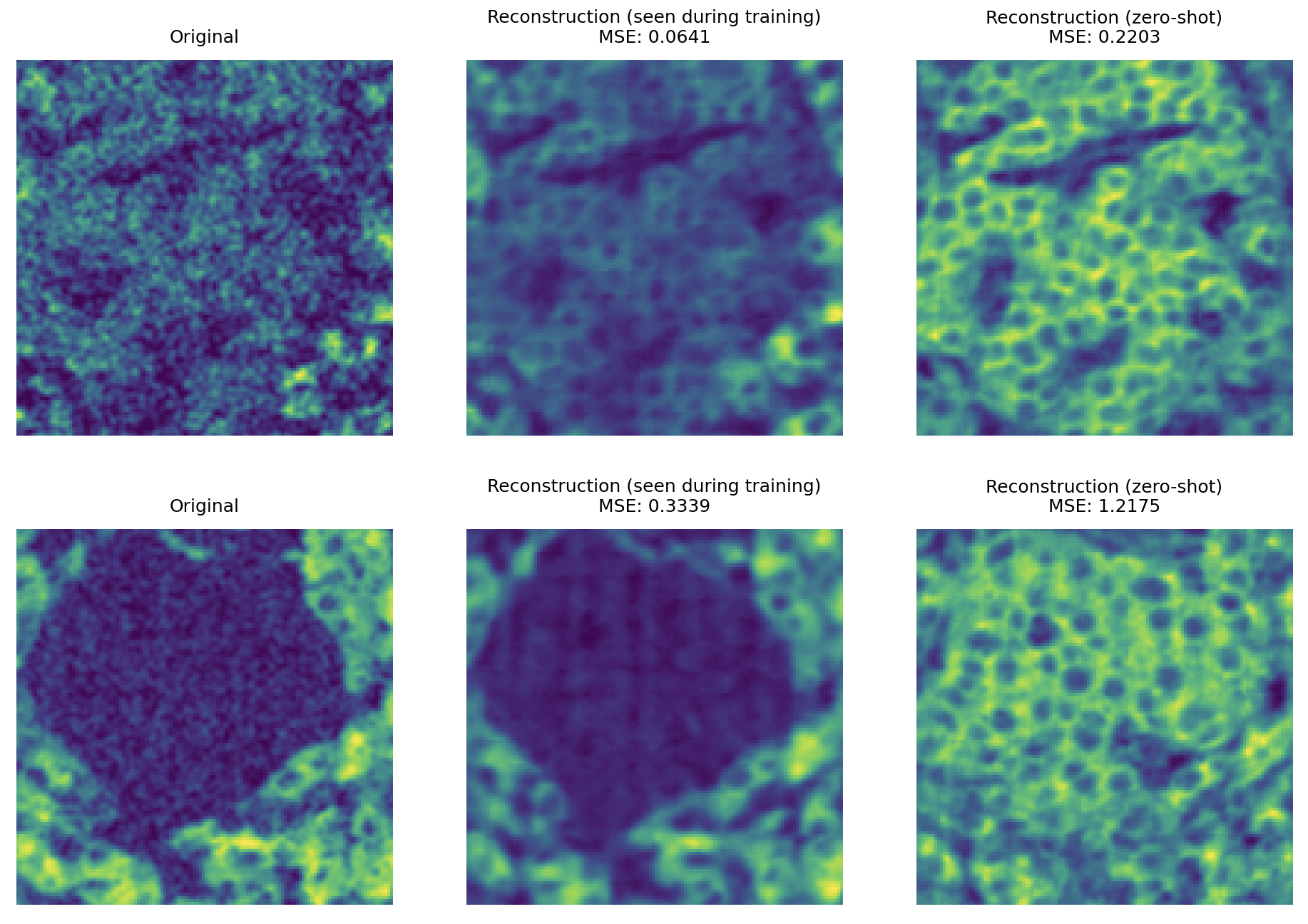}}
    }
    \caption{Examples of VirTues reconstructions for trained and zero-shot settings.}
    \label{fig:mse-zero-shot-recons}
  \end{center}
\end{figure}

Despite the high sequence similarity between AMY1A and AMY2A, reconstruction quality degrades dramatically in the zero-shot setting (mean MSE 0.96 vs. 0.24; see Figure~\ref{fig:mse-zero-shot}, for example reconstructions see Figure~\ref{fig:mse-zero-shot-recons}). This demonstrates that ESM sequence similarity does not reliably transfer to spatial co-expression similarity in IMC even when the accurate prediction is possible (as confirmed by a \textbf{Trained} setup).

\subsection{Optimization and training protocol.}\label{sec:hyperparams}
All models are trained with AdamW~\cite{loshchilov2019decoupled} using weight decay $10^{-4}$ and a cosine annealing learning-rate schedule with linear warmup. The learning rate increases linearly during the first 5 epochs to a peak of $\learningRateMax{}$, and is then annealed to a final value of $\learningRateMin{}$ by the end of training. We use mixed-precision training with \texttt{bfloat16} for efficiency. We used batch size of 8, as we observed lower batches tend to improve the convergence of models. We additionally apply $\ell_2$-norm gradient clipping with maximum norm $1.0$. We train all \name variants for 200 epochs.

\paragraph{Training compute resources.}
All main models were trained on NVIDIA A100 SXM 40\,GB GPUs, using one GPU per run. Pretraining each \name variant on 200 epochs required approximately 4 days. Downstream linear-probe experiments were run on CPU and required less than 24 hours per dataset. 

\subsection{IMC Preprocessing Pipeline}
\label{app:preprocessing}

IMC raw image data requires additional preprocessing prior to model training. We apply the following steps in order.

\paragraph{Arcsinh transformation.}
We apply a pixel-wise \texttt{arcsinh} transform with cofactor $5$ to compress the high dynamic range of IMC intensities while amplifying foreground signal and accommodating zero/negative values.

\paragraph{Denoising via low-pass filtering.}
We observed that IMC-specific background noise and hot pixels can destabilise
optimisation, particularly when training on smaller image crops.
To address this, we applied a Butterworth low-pass filter
\cite{Butterworth1930} implementation from \texttt{skimage.filters.butterworth} (v 0.25.2)
\cite{scikit-image}, with cutoff frequency ratio~$0.2$ relative to the Nyquist
frequency and filter order~$2$.
These settings provide a maximally flat passband response up to $0.2f_\mathrm{N}$,
attenuating high-frequency spatial noise while preserving low-frequency tissue
structure.

\paragraph{Panel-wise intensity standardization.}
We further normalize marker intensities to a $[0,1]$ interval using a modified min-max scaling applied independently per dataset panel. We fix the lower bound at $0$ and set the upper bound to the 99th percentile of the pooled intensity distribution across all markers within the panel, rounded up to one decimal place. This improves cross-dataset comparability while limiting the influence of extreme outliers.

\paragraph{Subimage extraction and augmentation.}
The model input is a $\patchSize{}\times\patchSize{}$ crops, where each pixel corresponds to approximately $1\,\mu\mathrm{m}^2$ tissue area. To mitigate I/O bound during training, we extracted non-overlapping subimages of shape $256\times256$ from the training split images. During training, we apply random rotation, reflection and cropping to generate crops from subimages of intended shape. For one training epoch, all subimages from all images are used once. During evaluation, we extract image crops via a deterministic central crop from each source image to ensure consistency and reproducibility.

\subsection{Comparison with VirTues preprocessing}
\label{app:signal_stability_analysis}

Because our IMC Preprocessing pipeline differs from preprocessing of previously published methods~\cite{virtues, Liu2025-EVA}, below we will provide a justification for our choices. As EVA~\cite{Liu2025-EVA} does not provide preprocessing details for IMC, we will concentrate on VirTues~\cite{virtues} preprocessing pipeline.

\subsubsection{\name arcsinh vs VirTues log1p transformation}

Following established IMC/CyTOF analyses practice~\cite{Bollhagen2025-IMC-preprocessing,Windhager2023-IMC-preprocessing,Hosogane2023-IMC-preprocessing,Bendall2011-CyTOF-preprocessing}, we apply
$\texttt{arcsinh}(x/5)$ to raw per-pixel ion counts rather than the $\texttt{log1p}$
transformation used by VirTues~\cite{virtues}. The distinction matters because IMC pixel
values are discrete, non-negative ion counts, and the two transformations treat
low-count measurements very differently.

$\texttt{log1p}$ is asymptotically logarithmic, so equal multiplicative intervals receive
equal compression regardless of the count magnitude.
Concretely, the mapped differences for three count ranges of identical multiplicative width
($\times 4$) are:
\begin{align*}
  \texttt{log1p}(3)  - \texttt{log1p}(0)   &= \ln 4 \approx 1.39, \\
  \texttt{log1p}(255)- \texttt{log1p}(63)  &= \ln 4 \approx 1.39, \\
  \texttt{log1p}(1023)-\texttt{log1p}(255) &= \ln 4 \approx 1.39.
\end{align*}
The first interval (0--3 counts) falls squarely within the range attributable to
spatial or channel spill-over noise; the latter two correspond to biologically
meaningful low-to-medium and medium-to-high expression transitions.
Under \texttt{log1p}, all three differences receive identical weight, inflating
the contribution of noise-level counts in any subsequent distance- or
covariance-based computation.

$\texttt{arcsinh}(x/5)$ alleviates this by behaving approximately linearly near
zero and logarithmically for large $x$, with the cofactor $c=5$ controlling the
transition point.
The same three intervals now map to:
\begin{align*}
  \texttt{arcsinh}(3/5)  - \texttt{arcsinh}(0/5)   &\approx 0.57, \\
  \texttt{arcsinh}(255/5)- \texttt{arcsinh}(63/5)  &\approx 1.39, \\
  \texttt{arcsinh}(1023/5)-\texttt{arcsinh}(255/5) &\approx 1.39.
\end{align*}
The noise-level interval is compressed by a factor of roughly 2.5 comparing to the weight assigned
to the biologically informative ranges, while the latter two remain
comparably scaled, preserving the dynamic range that matters for biological analysis.

\subsubsection{\name Denoising via low-pass filtering vs VirTues Gaussian blur}

\begin{figure}[th]
  \begin{center}
    \centerline{\kg{\includegraphics[width=\textwidth]{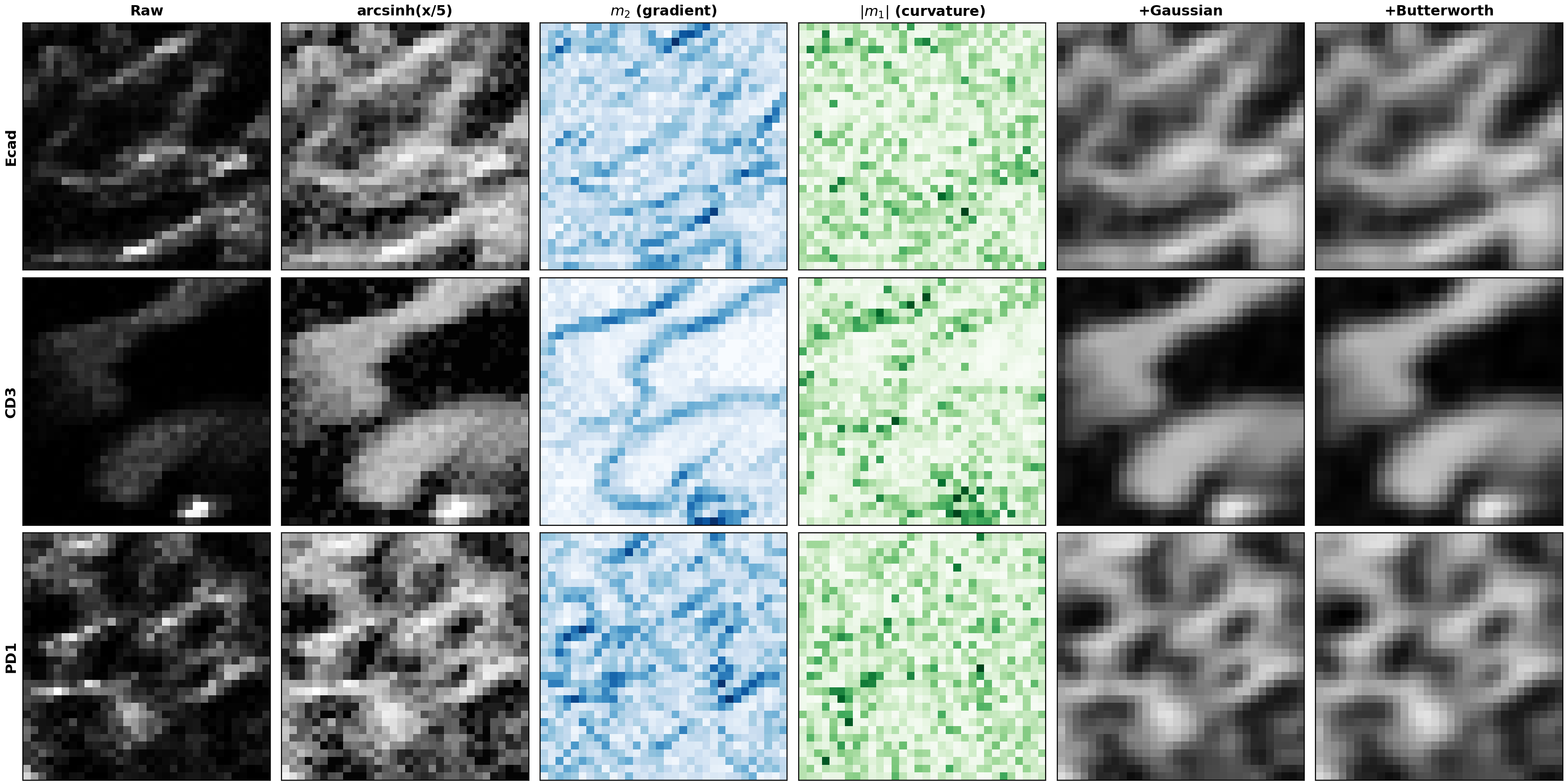}}
    }
    \caption{
    \textbf{Exemplary qualitative comparison of spatial preprocessing on $32\times32$ IMC patches from~\cite{Eling2025}.}
    Each row shows a single antibody channel: E-cadherin (Ecad, epithelial marker),
    CD3 (T-cell marker), and PD-1.
    The six columns are: raw counts; arcsinh$(x/5)$ normalisation; the corresponding
    omnidirectional gradient $m_2$ and absolute curvature $|m_1|$ maps (computed on
    the arcsinh-normalised full image and coloured with sequential blue and green
    scales, respectively); arcsinh$+$Gaussian blur ($\sigma{=}1$\,px, $3{\times}3$
    kernel, analogous to VirTues preprocessing); and arcsinh$+$Butterworth low-pass filter (order~2, cutoff~0.2 Nyquist, used by \name).
    Intensity columns share a common colour scale across arcsinh and filtered variants
    within each row; the raw column and the two metric columns are each scaled
    independently per row.
    }
    \label{fig:supp_IMC_transformations_patches}
  \end{center}
\end{figure}

Instead of 3x3-truncated Gaussian blur used by VirTues we used a Butterworth~\cite{Butterworth1930} low-pass filtering (as described in Appendix~\ref{app:preprocessing}). The main reason behind this choice is a high intrinsic noise present in the IMC data (see Figure~\ref{fig:supp_IMC_transformations_patches}). To quantitatively measure this noise we introduce two pixel-level metrics: \textit{gradient} and \textit{curvature}.
Let $x_0$ denote the intensity of a pixel with eight-connected neighbours
$\{x_k\}_{k=1}^{8}$ (N, NE, E, SE, S, SW, W, NW).
The \emph{curvature}
\begin{equation}
  m_1 = x_0 - \frac{1}{8}\sum_{k=1}^{8} x_k
\end{equation}
measures how much a pixel deviates from its local neighbourhood mean; it is
equivalent to the discrete isotropic Laplacian and is sensitive to
high-frequency structure such as shot noise and ringing artefacts.
The \emph{gradient}
\begin{equation}
  m_2 = \sqrt{\!\left(\frac{x_E - x_W}{2}\right)^{\!2}
              + \left(\frac{x_S - x_N}{2}\right)^{\!2}}
\end{equation}
is the Euclidean magnitude of the central-difference gradient along the two
principal axes, measuring the strength of local intensity transitions
irrespective of orientation.
Both metrics are computed after applying the full preprocessing pipeline;
$|m_1|$ is used to obtain a non-negative curvature magnitude.
Periodic boundary conditions are applied via circular array shifts.

\begin{figure}[ht]
  \begin{center}
    \centerline{\kg{\includegraphics[width=0.95\textwidth]{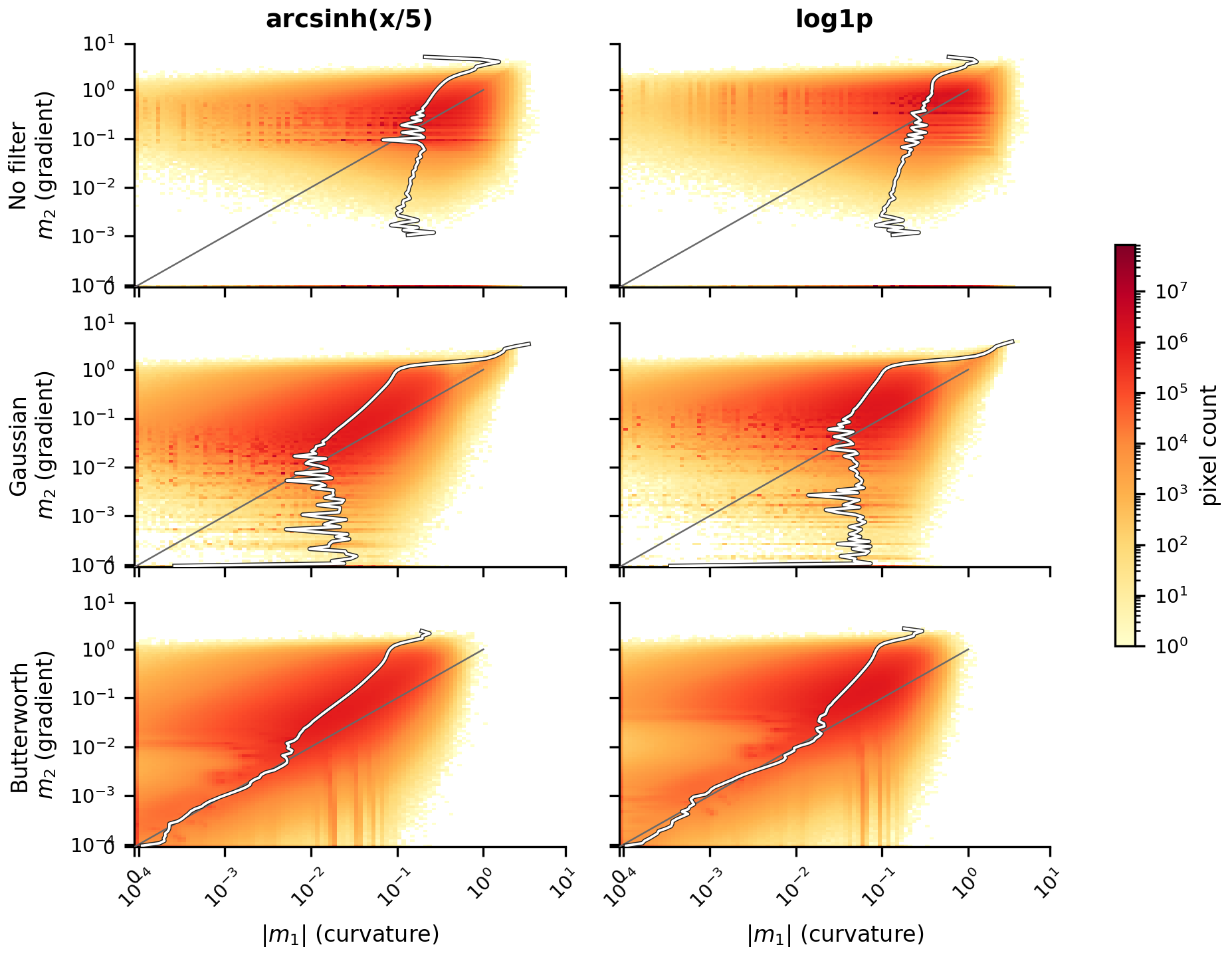}}
    }
    \caption{
    \textbf{Joint distribution of curvature $|m_1|$ and gradient $m_2$ under six preprocessing pipelines.}
    Rows: no spatial filter / Butterworth / Gaussian; columns: arcsinh$(x/5)$ /
    $\log(1+x)$ normalisation.
    Each panel accumulates a 2D histogram over all pixels, images, and antibody
    channels, using a mixed bin scale (one zero-anchor bin $[0,10^{-4}]$ followed by
    99 log-spaced bins up to $10$).
    All six panels share a logarithmic colour scale.
    The white curve shows $\mathbb{E}[|m_1|\mid m_2]$; the diagonal $|m_1|=m_2$
    is for a reference. 
    Both filters reduce pixel curvature relative to the unfiltered signal (top row);
    Gaussian smoothing introduces discrete curvature artefacts at fixed $|m_1|$ values, whereas Butterworth low-pass filtering yields a smoother, more uniform $(|m_1|, m_2)$ distribution.
    }
    \label{fig:supp_IMC_transformations}
  \end{center}
\end{figure}

The analysis of these pixel-level metrics computed on IMC images from~\cite{Eling2025} reveal interesting noise dependencies (see Figure~\ref{fig:supp_IMC_transformations}). For both raw data (without filtering) and Gaussian blur, there is a significant curvature divergence for a flat image areas ($|m2|<10^{-2}$). Low-pass filtering removes this divergence providing almost linear dependency between $|m1|$ and $|m2|$ within this range (as visualized by $\mathbb{E}[|m_1|\mid m_2]$ curve).

\subsubsection{\name min--99th-percentile panel-wise scaling vs.\ VirTues per-marker z-score intensity standardization}\label{app:panel-wise-scaling}

Compared with our min--99th-percentile scaling computed over the full dataset, VirTues applies per-marker z-score normalization based on the statistics computed on cell masks. \kg{MSE values obtained under these two preprocessing protocols are therefore not directly comparable.} Our choice is motivated by three main reasons:

\begin{itemize}
    \item Even after preprocessing with \texttt{log1p} or \texttt{arcsinh}, IMC data remain zero-inflated and hence highly non-Gaussian. Therefore, we use min--99th-percentile scaling, which is better suited to this type of data.
    
    \item The per-marker, per-dataset statistics required for VirTues z-score normalization are unavailable for markers absent from a panel, what makes results of virtual staining difficult to interpret in such cases, as they cannot be converted back to an absolute value scale. Additionally such transformation invalidates out-of-panel and out-of-cohort task assumptions. In contrast, \name quantile scaling is zero-preserving, and because the scaling quantile is computed over the full dataset, it can be easily approximated using only the markers present in the panel.

    \item As z-score statistics used by VirTues require cell segmentation, this approach is dependent on a segmentation model and quality of the segmentation masks.
\end{itemize}

\subsection{KRONOS reproduction}\label{app:kronos-reproduction}

The initial attempts to pre-train a self-supervised foundation model KRONOS~\cite{KRONOS2025} on our IMC dataset used the native KRONOS implementation. This architecture relied on a standard Vision Transformer stem and incorporated per-channel patch embeddings together with sinusoidal marker encodings. However, these initial experiments exhibited severe optimisation instabilities, resulting in consistent loss divergence. 

This empirical challenge is consistent with observations reported by the original KRONOS authors, who explicitly excluded ion-based modalities, such as IMC, from their training corpus, noting that such data might require modality-specific architectural adaptations. In addition, unlike \name and VirTues, KRONOS employs ambiguous marker encodings, where markers are arranged in an ordered list such that biologically or semantically similar markers are positioned close to one another. This introduces implicit assumptions about marker relationships and ordering that may not generalise well across heterogeneous IMC panels. Despite evaluating multiple alternative marker orderings, none of these configurations resulted in stable convergence during training.

To mitigate the exploding loss, we first experimented with expanding the marker embedding vocabulary using custom sinusoidal encodings and explicitly injecting spatial coordinate information via a meshgrid representation. However, these modifications proved insufficient to stabilise the training dynamics.

\subsection{\kg{Transfer to external panels}}
\label{app:external-panels}
\kg{The ten test panels comprise held-out IMMUcan Head \& Neck~\cite{immucan_2025} and nine external panels from four studies absent from IMC17M: Allam et al.~\cite{Allam2022-imc}, Hu et al.~\cite{Hu2023-imc}, Moldoveanu et al.~\cite{Moldoveanu2022-imc}, and Schulz et al.~\cite{Schulz2018-imc}. These four studies were included in the original VirTues pretraining corpus.
The retrained VirTues and EVA baselines below use IMC17M and OP. All models receive the same in-vocabulary inputs and are scored on the same overlapping targets; no new-marker adaptation is used in this comparison.}
\kg{$\nameConv$ and $\nameSwin$ each achieve the lowest panel-average MSE on three of the ten panels; VirTues leads on four. Paired Wilcoxon signed-rank tests on image--marker MSE with Benjamini--Hochberg FDR correction yield wins/losses/non-significant differences versus VirTues of 6/1/3 for $\nameSwin$, and 3/3/4 for both $\nameConv$ and $\nameViT$. Thus, the multi-panel staining advantage is carried by $\nameSwin$.}

\paragraph{\kg{Embedding-only adaptation.}}
\kg{The external Moldoveanu cohort contains 48 markers, of which 43 are in the pretrained vocabulary. For $\nameConv$, we add embeddings for CCR7, CD29, MEK1-2, pMEK1-2, and S100, training only these embeddings with the remaining network frozen. Table~\ref{tab:moldoveanu-adaptation} compares adaptation with no adaptation and with pretraining that includes the cohort. The no-adaptation staining score is omitted because it averages only 43 markers and is not directly comparable to the 48-marker evaluations. %
Embedding-only adaptation used the pretraining objective and hyperparameters and took approximately 2 hours on one A100 (about 2\% of pretraining compute).}
\begin{table}[ht]
\centering\small
\caption{\kg{Moldoveanu transfer with $\nameConv$: cell-typing Macro-F1 for panels P1/P2 and staining on all 48 markers where available. The cohort was included in original VirTues pretraining.}}
\label{tab:moldoveanu-adaptation}
\begin{tabular}{lcccc}
\toprule
\kg{Setting} & \kg{P1} & \kg{P2} & \kg{MSE} & \kg{Pearson}\\
\midrule
\kg{No adaptation} & \kg{.765} & \kg{.643} & \kg{--} & \kg{--}\\
\kg{New-marker embeddings only} & \kg{.765} & \kg{.641} & \kg{.081} & \kg{.531}\\
\kg{Cohort included in pretraining} & \kg{.776} & \kg{.649} & \kg{.040} & \kg{.687}\\
\kg{Original VirTues} & \kg{.734} & \kg{.617} & \kg{--} & \kg{--}\\
\bottomrule
\end{tabular}
\end{table}
\kg{Downstream tasks on the other three external studies were not run.}

\paragraph{\kg{Full-panel versus overlap-only inputs.}}
\label{app:full-panel-inputs}
\kg{For the five panels with markers absent from IMC17M, we compare full-panel and overlap-only inputs for the retrained IMC17M+OP baselines. Both conditions are scored on the same overlapping targets (Table~\ref{tab:full-panel-inputs}). Extra marker inputs worsen VirTues MSE on three of five panels and Pearson on four; EVA MSE improves on four and worsens on one. Thus, unseen-marker inputs are not uniformly beneficial, and the VirTues pattern does not generalize to EVA.}
\begin{table}[ht]
\centering\small
\caption{\kg{MSE/Pearson on shared targets with full-panel (F) and overlap-only (O) inputs.}}
\label{tab:full-panel-inputs}
\begin{tabular}{lcccc}
\toprule
\kg{Panel} & \kg{VirTues F} & \kg{VirTues O} & \kg{EVA F} & \kg{EVA O}\\
\midrule
\kg{A1} & \kg{.0348/.62} & \kg{.0350/.65} & \kg{.0659/.50} & \kg{.0672/.50}\\
\kg{A2} & \kg{.1853/.65} & \kg{.1853/.65} & \kg{.3498/.48} & \kg{.4618/.47}\\
\kg{H1} & \kg{.0661/.29} & \kg{.0559/.34} & \kg{.1060/.18} & \kg{.1080/.18}\\
\kg{M1} & \kg{.2634/.52} & \kg{.2011/.54} & \kg{.3577/.37} & \kg{.3612/.39}\\
\kg{M2} & \kg{.2299/.24} & \kg{.1849/.29} & \kg{.3112/.20} & \kg{.3019/.21}\\
\bottomrule
\end{tabular}
\end{table}

\subsection{Virtual staining results for VirTues preprocessing (\texttt{VP})}\label{app:virtual-staining-vp}

\begin{figure}[ht]
  \begin{center}
    \centerline{\kg{\includegraphics[width=\textwidth]{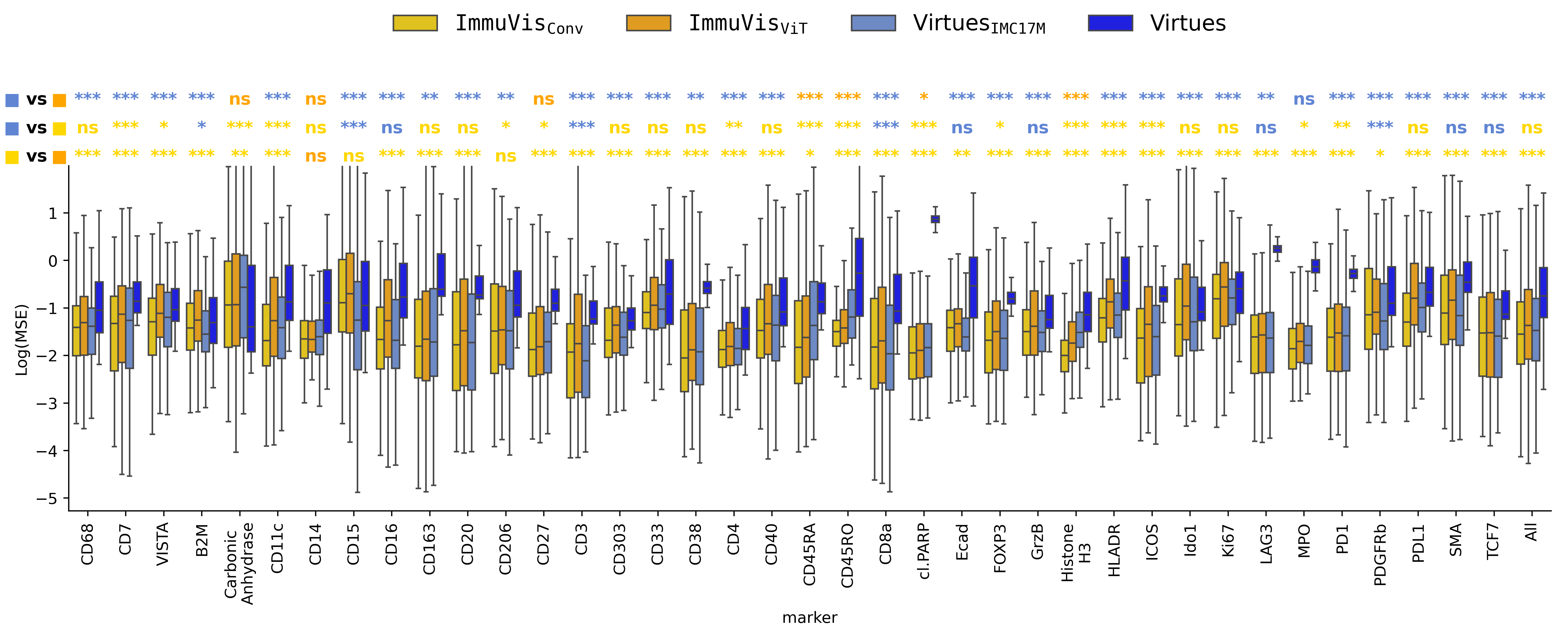}}}
    \caption{
      \textbf{Quantitative virtual staining evaluation for MSE on the Head \& Neck cohort for models train using VirTues preprocessing (\texttt{VP}).}
      Per-marker reconstruction accuracy measured by image-level $\log(\mathrm{MSE})$ for $\nameConv$, $\nameViT$, $\text{VirTues}_{\texttt{IMC17M}}$ and VirTues (original checkpoint from ~\cite{virtues}).
     Each boxplot summarizes the distribution of image-level scores, where each image score is obtained by averaging patch-level errors over all patches from that image.
        The three rows above the plot report paired Wilcoxon signed-rank tests results across images for the corresponding model pairs (as indicated in the left top corner), with significance after FDR correction (ns - not significant; $(^{*}) <\!0.05$; $(^{**}) <\!0.01$; $(^{***}) <\!0.001$).
    }
    \label{fig:virtual_staining_quantitatively_vp}
  \end{center}
\end{figure}

\begin{figure}[ht]
  \begin{center}
    \centerline{\kg{\includegraphics[scale=0.9]{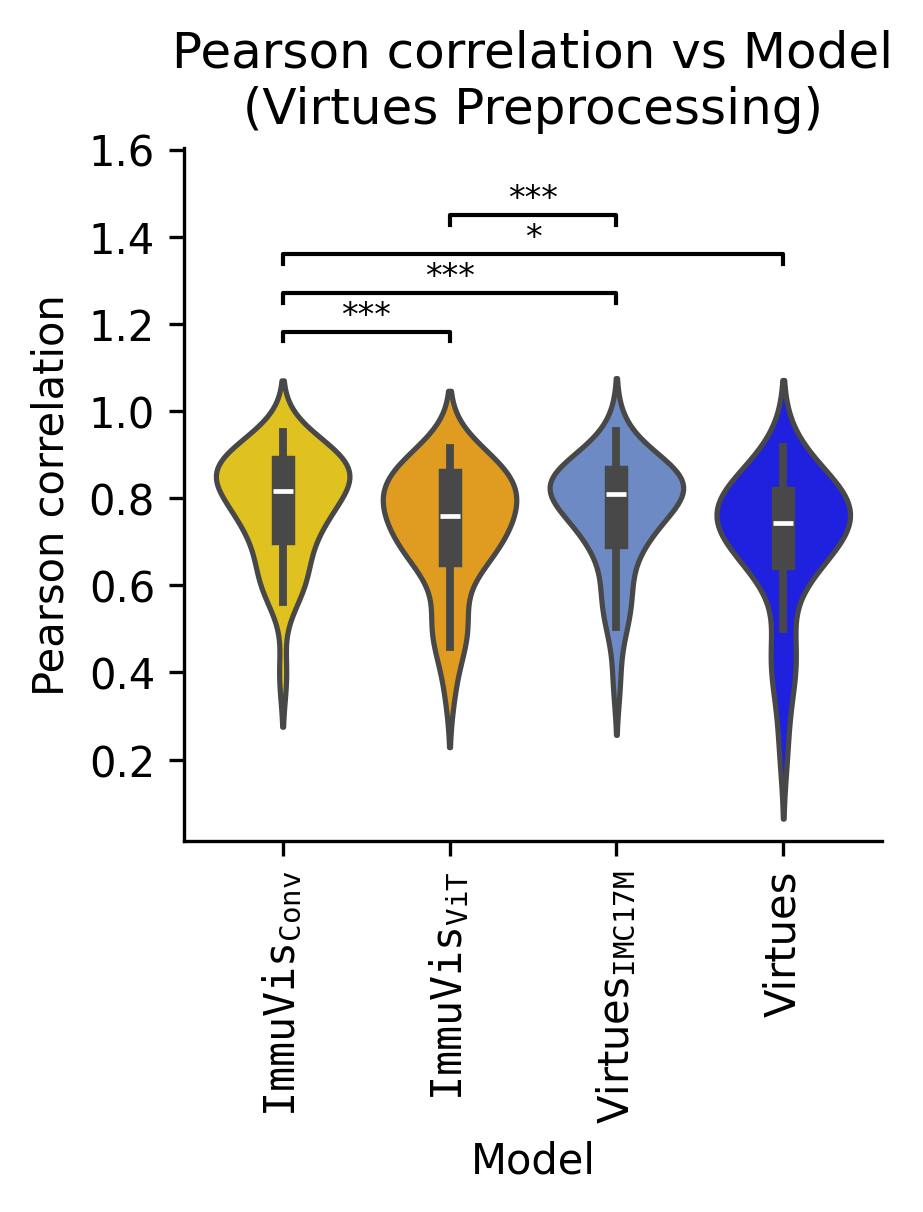}}}
    \caption{
      \textbf{Quantitative virtual staining evaluation for Pearson correlation  on the Head \& Neck cohort for models trained using VirTues preprocessing (\texttt{VP}).}
      Per-model Pearson correlation measured on a marker-level for $\nameConv$, $\nameViT$, $\text{VirTues}_{\texttt{IMC17M}}$ and VirTues (original checkpoint from ~\cite{virtues}).
     Each boxplot summarizes the distribution of marker-level Pearson correlation computed across all images. P-values are obtained via Wilcoxon signed-rank tests results across markers for the corresponding model pairs (ns - not significant; $(^{*}) <\!0.05$; $(^{**}) <\!0.01$; $(^{***}) <\!0.001$).
    }
    \label{fig:virtual_staining_pearson_vp}
  \end{center}
\end{figure}

Quantitative results for virtual staining on HN with VirTues preprocessing (\texttt{VP}) are presented in Table~\ref{tab:virtual_staining_accuracy_vp}. 
A marker-wise MSE breakdown, together with statistical significance evaluation, are presented in Figure~\ref{fig:virtual_staining_quantitatively_vp}. Figure~\ref{fig:virtual_staining_pearson_vp} presents Pearson correlation distributions together with statistical significance evaluation.

\begin{table}
\vspace{-1.0em}
\centering
\caption{
Virtual staining accuracy measured by MSE and Pearson correlation under the VirTues preprocessing protocol.
\textbf{Bold} denotes the best score.
}
\label{tab:virtual_staining_accuracy_vp}
\begin{small}
\begin{tabular}{lcc}
\toprule
Model & MSE ($\downarrow$) & Pearson ($\uparrow$) \\
\midrule
$\text{VirTues}$ & .797 & .700 \\
$\text{VirTues}_{\texttt{IMC17M}}$ & .410 & .764 \\
\midrule
$\nameViT$  & .502 & .731 \\
$\nameConv$ & \textbf{.400} & \textbf{.783} \\
\bottomrule
\end{tabular}
\end{small}
\vspace{-1.0em}
\end{table}

\subsection{On computing Pearson correlation across the whole dataset rather than on a per-image basis}
\label{app:virtual-staining-pearson-rationale}

Contrary to VirTues~\cite{virtues}, we compute the Pearson correlation between the ground truth and the model predictions on a per-marker basis across the entire dataset, rather than independently for each image. The main motivation for this choice is that Pearson correlation computed on a per-image basis is linearly equivalent to the mean squared error (MSE) computed on data that have been z-scored independently for each image. This method completely loses cross-image marker dependencies and may additionally suffer from variance explosion for frequent images and markers with near-constant zero expression values.

To demonstrate this linear equivalence, assume that two variables \(x\) and \(y\) are z-scored, i.e.,
\[
\mathbb{E}[x] = \mathbb{E}[y] = 0,
\qquad
\mathrm{Var}(x) = \mathrm{Var}(y) = 1.
\]

The mean squared error between \(x\) and \(y\) is
\[
\mathrm{MSE}(x,y)
=
\mathbb{E}\left[(x-y)^2\right].
\]

Expanding the square gives
\[
\mathrm{MSE}(x,y)
=
\mathbb{E}[x^2]
+
\mathbb{E}[y^2]
-
2\mathbb{E}[xy].
\]

Because \(x\) and \(y\) are z-scored,
\[
\mathbb{E}[x^2] = \mathbb{E}[y^2] = 1,
\]
and since their means are zero,
\[
\mathrm{Corr}(x,y)
=
\mathbb{E}[xy].
\]

Therefore,
\[
\mathrm{MSE}(x,y)
=
1 + 1 - 2\,\mathrm{Corr}(x,y)
=
2 - 2\,\mathrm{Corr}(x,y).
\]

Rearranging yields
\[
\mathrm{Corr}(x,y)
=
1 - \frac{1}{2}\mathrm{MSE}(x,y).
\]

Thus, for z-scored variables, Pearson correlation and mean squared error are linearly related.

\subsection{Extended virtual staining results for \name preprocessing (\texttt{OP})}\label{app:virtual-staining-op}

\begin{figure}[ht]
  \begin{center}
    \centerline{\kg{\includegraphics[width=\textwidth]{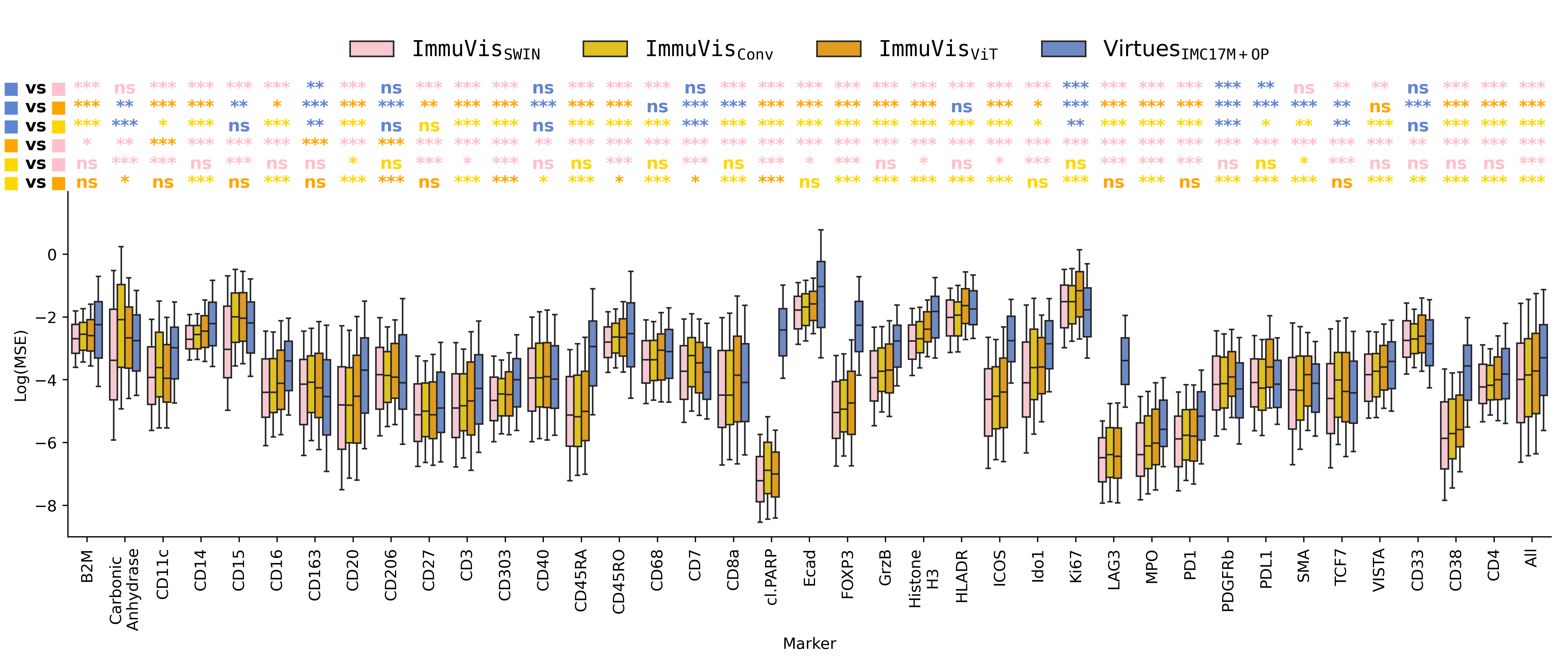}}}
    \caption{
    \textbf{Quantitative virtual staining evaluation for MSE on the Head \& Neck cohort for models train using \name preprocessing (\texttt{OP}).}
      Per-marker reconstruction accuracy measured by image-level $\log(\mathrm{MSE})$ for $\nameConv$, $\nameSwin$, $\nameViT$, and $\text{VirTues}_{\texttt{IMC17M+OP}}$.
     Each boxplot summarizes the distribution of image-level scores, where each image score is obtained by averaging patch-level errors over all patches from that image.
        The three rows above the plot report paired Wilcoxon signed-rank tests results across images for the corresponding model pairs (as indicated in the left top corner), with significance after FDR correction (ns - not significant; $(^{*}) <\!0.05$; $(^{**}) <\!0.01$; $(^{***}) <\!0.001$).
    }
    \label{fig:virtual_staining_quantitatively_op}
  \end{center}
\end{figure}

\begin{figure}[ht]
  \begin{center}
    \centerline{\kg{\includegraphics[scale=0.9]{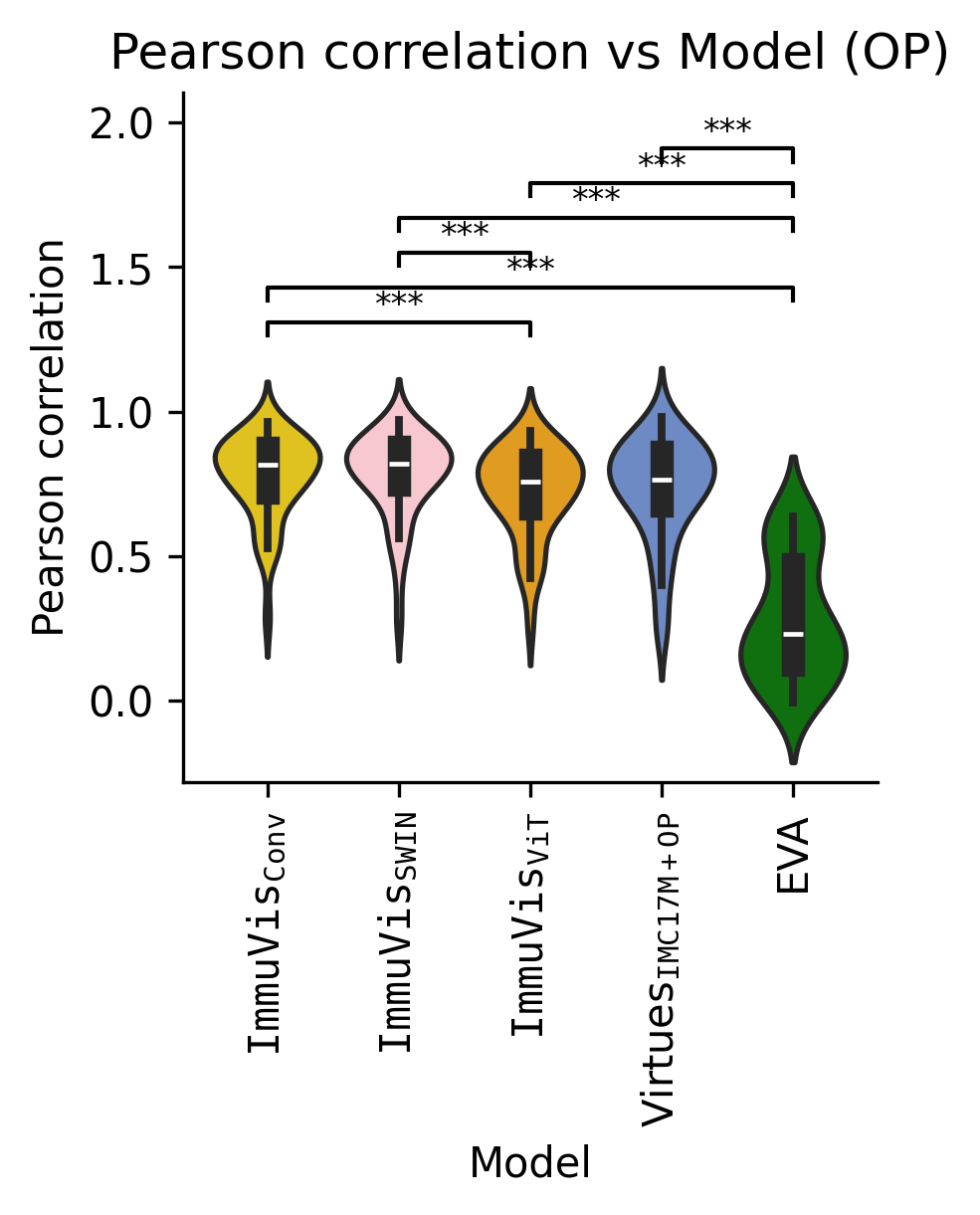}}}
    \caption{
      \textbf{Quantitative virtual staining evaluation for Pearson correlation  on the Head \& Neck cohort for models train using \name preprocessing (\texttt{OP}).}
      Per-model Pearson correlation measured on a marker-level for $\nameConv$, $\nameSwin$, $\nameViT$, $\text{VirTues}_{\texttt{IMC17M+OP}}$ and \kg{$\text{EVA}_{\texttt{OP}}$ (released checkpoint evaluated outside its native preprocessing).}
     Each boxplot summarizes the distribution of marker-level Pearson correlation computed across all images. P-values are obtained via Wilcoxon signed-rank tests results across markers for the corresponding model pairs (ns - not significant; $(^{*}) <\!0.05$; $(^{**}) <\!0.01$; $(^{***}) <\!0.001$).
    }
    \label{fig:virtual_staining_pearson_op}
  \end{center}
\end{figure}

A marker-wise MSE breakdown on HN, together with statistical comparisons, is presented in Figure~\ref{fig:virtual_staining_quantitatively_op}. Figure~\ref{fig:virtual_staining_pearson_op} presents Pearson correlation distributions together with statistical comparisons.

\subsection{Extended uncertainty calibration analysis}
\label{app:uncertainty_calibration}

\kg{This appendix expands the headline results in Section~\ref{sec:uncerainty_results} with the full set of calibration diagnostics. All figures are computed on out-of-cohort leave-one-out (LOO) virtual staining over the IMMUcan SCCHN1 test split~\cite{immucan_2025} (the cohort was excluded at training), across all 40 markers of the cohort. Per model this comprises 149 patches $\times$ 40 markers $=$ 5\,960 patch--channel reconstructions.}

\paragraph{Notation.}
For each pixel let $\mu$ denote the predicted mean, $\sigma^{2}$ the predicted variance from the heteroscedastic head, $y$ the arcsinh-normalised target, and $r=y-\mu$ the residual. Throughout, $|r|$ denotes absolute residual at the pixel level and $\mathrm{MAE}$ the per-(image, channel) mean of $|r|$.

\subsubsection{Per-(image, channel) error--uncertainty correlation}
\label{app:loo_scatter}

\textbf{Method.}
For every (image, channel) pair we summarise the predicted uncertainty by $\log\overline{\sigma^{2}} = \log\!\bigl(\tfrac{1}{N}\sum_{i}\sigma_{i}^{2}\bigr)$ and the reconstruction error by $\log\mathrm{MAE}$. Each LOO reconstruction contributes one such (uncertainty, error) point in log--log space. Pearson correlation is computed on the 5\,960 patch--channel rows per model; we overlay an ordinary-least-squares fit per model.

\begin{figure}[h]
  \centering
  \kg{\includegraphics[width=0.95\textwidth]{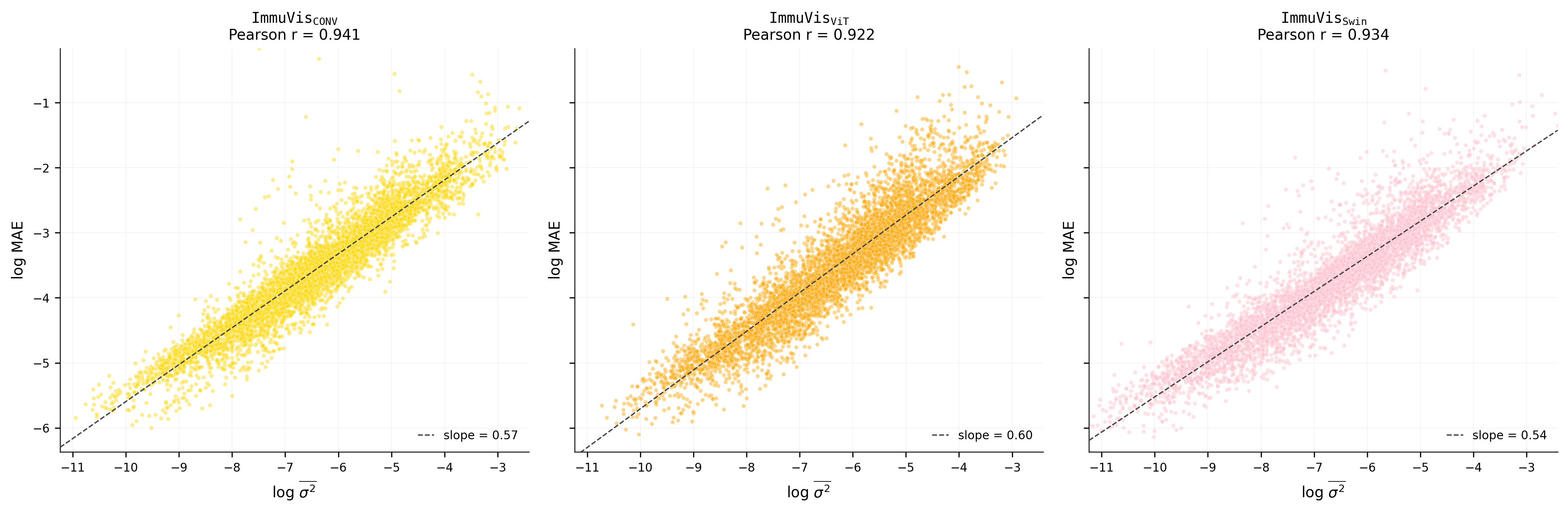}}
  \caption{\kg{\textbf{Predicted uncertainty tracks reconstruction error across (image, channel) pairs under out-of-cohort LOO virtual staining.} Each point is one LOO reconstruction (5\,960 per model). Strong log--log scaling between mean predicted variance and MAE holds for all three \name variants: Pearson $r_{\mathrm{CONV}}\!=\!.941$, $r_{\mathrm{ViT}}\!=\!.922$, $r_{\mathrm{Swin}}\!=\!.934$. The OLS fit slopes ($\nameConv$, $\nameViT$, $\nameSwin$) are $.57$, $.60$, $.54$ respectively. The relationship confirms that the heteroscedastic head captures local ambiguity in the reconstruction.}}
  \label{fig:app_loo_scatter}
\end{figure}

\subsubsection{Reliability of the predictive Gaussian}
\label{app:reliability}

\begin{figure}[h]
  \centering
  \kg{\includegraphics[width=0.7\textwidth]{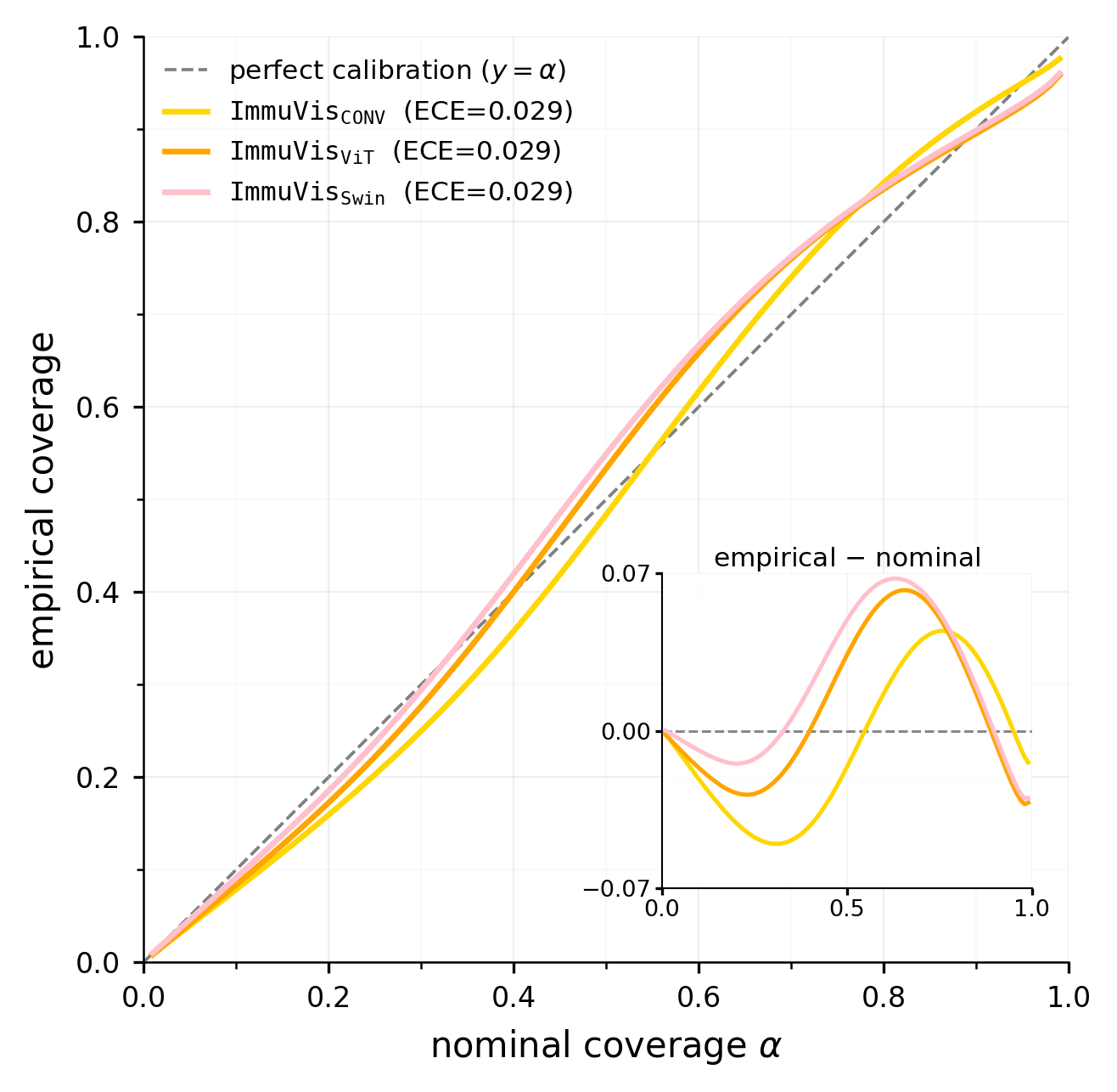}}
  \caption{\kg{\textbf{Reliability diagram - global, out-of-cohort LOO.} Empirical coverage of the predictive Gaussian's central $\alpha$-interval as a function of nominal $\alpha$. All three \name variants closely track the perfect-calibration line and their ECE values agree to within $.0006$. The inset, plotted on a magnified $y$-axis, reveals that the three architectures miscalibrate in qualitatively different directions: $\nameConv$ predominantly under-covers (peaking near $\alpha\!\approx\!0.3$), while $\nameViT$ and $\nameSwin$ predominantly over-cover (peaking near $\alpha\!\approx\!0.7$, with $\nameSwin$ deviating most). Integrating the absolute deviations gives nearly identical ECE values ($.029\pm.0003$), indicating that overall calibration quality is comparable across architectures despite distinct miscalibration shapes.}}
  \label{fig:app_reliability}
\end{figure}

\textbf{Method.}
A reliability diagram tests whether the predictive distribution $\mathcal{N}(\mu,\sigma^{2})$ assigns calibrated coverage to its central intervals. For each nominal level $\alpha\in\{.01,.02,\dots,.99\}$ (99-point grid) we compute the half-width $z_{\alpha}\!=\!\Phi^{-1}\bigl((1+\alpha)/2\bigr)$ and the empirical coverage
\[
\widehat{\mathrm{cov}}(\alpha) \;=\; \frac{1}{N_{\mathrm{px}}} \sum_{i=1}^{N_{\mathrm{px}}} \mathbf{1}\!\bigl[|r_{i}| \le z_{\alpha}\,\sigma_{i}\bigr] ,
\]
on all pixels. Implementation note: the per-$\alpha$ indicator is counted once per (patch, channel) and summed at the group level, so the coverage at every $\alpha$ is computed losslessly without weighted-average approximation. A perfectly calibrated heteroscedastic head produces $\widehat{\mathrm{cov}}(\alpha)\!=\!\alpha$ for every $\alpha$ - the diagonal $y=\alpha$ in the main panel of Figure~\ref{fig:app_reliability}. The single-number summary is the regression Expected Calibration Error
\[
\mathrm{ECE} \;=\; \mathbb{E}_{\alpha}\,\bigl|\widehat{\mathrm{cov}}(\alpha) - \alpha\bigr|
\;\;\approx\;\; \frac{1}{99}\sum_{\alpha}\bigl|\widehat{\mathrm{cov}}(\alpha) - \alpha\bigr|,
\]
following \citep{Kuleshov2018} which is the metric reported in Table~\ref{tab:uncertainty_calibration}.

\subsubsection{Sharpness vs.\ calibration}
\label{app:sharpness_vs_ece}

\begin{figure}[h]
  \centering
  \kg{\includegraphics[width=\textwidth]{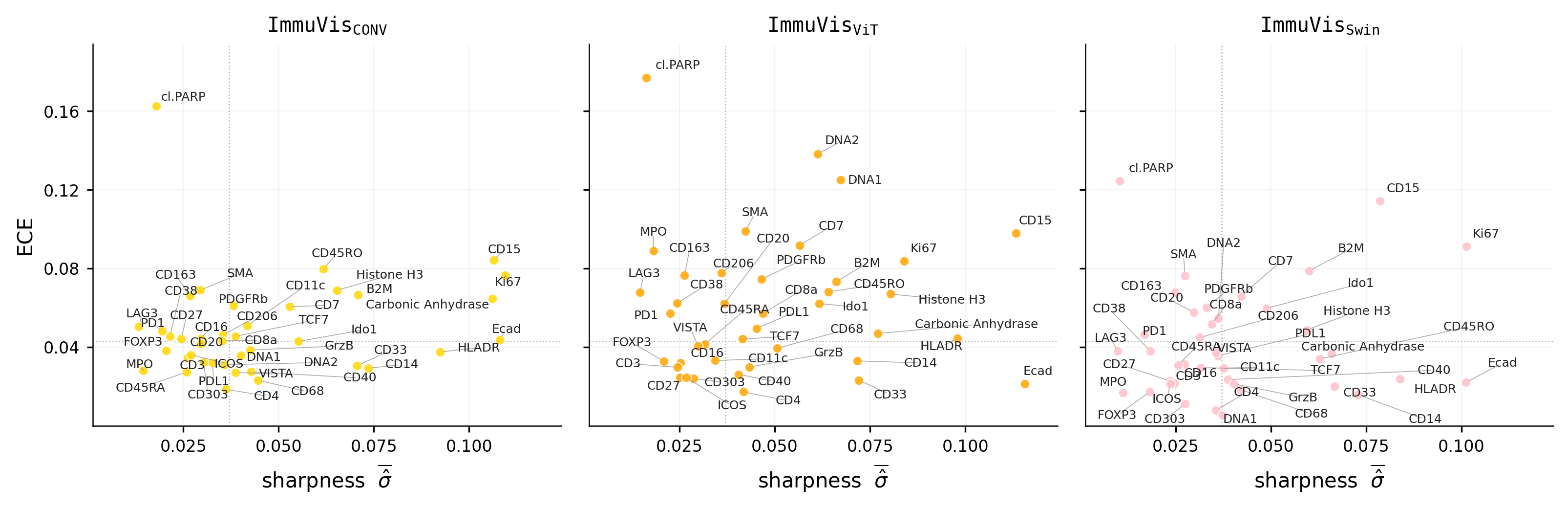}}
  \caption{\textbf{Sharpness vs.\ ECE per marker.} Each point is one of the 40 markers in the Head \& Neck cohort. The bulk of markers cluster in the lower-left quadrant (mean $\sigma \lesssim .05$, ECE $<\!.05$), confirming that the calibrated coverage in Figure~\ref{fig:app_reliability} is achieved at narrow predictive intervals.}
  \label{fig:app_sharp_vs_ece}
\end{figure}

\textbf{Method.}
Low ECE confirms that the predictive intervals contain the right fraction of pixels but not that those intervals are tight. Sharpness captures this complementary axis \citep{Gneiting2007}; for a marker $m$ we report the pixel-weighted mean of the predicted standard deviation,
\[
\mathrm{sharpness}(m) \;=\; \frac{\sum_{\mathrm{pc}\,:\,\mathrm{marker}=m} N_{\mathrm{pc}}\,\overline{\sigma}_{\mathrm{pc}}}{\sum_{\mathrm{pc}\,:\,\mathrm{marker}=m} N_{\mathrm{pc}}} .
\]
The global sharpness, averaged across the 40 markers, is $.046$, $.048$, and $.041$ for $\nameConv$, $\nameViT$, and $\nameSwin$ respectively. Plotted jointly with per-marker ECE in Figure~\ref{fig:app_sharp_vs_ece}, the desirable region is the lower-left (small $\sigma$ \emph{and} well-calibrated). The upper-right corner (large $\sigma$, large ECE) flags markers where the model is both unsure and miscalibrated; the upper-left flags markers where the model is precise but its uncertainty does not match the residuals.

\subsubsection{Sparsification curves and AUSE}
\label{app:sparsification}

\textbf{Method.}
The sparsification analysis evaluates how well $\sigma$ \emph{ranks} pixels by error, integrating over thresholds rather than picking one. We sort all pixels by their predicted $\sigma$ and progressively remove the most-uncertain fraction $f\in[0,1)$; the residual error of the kept set, with RMSE as the base metric (matching Table~\ref{tab:uncertainty_calibration}), is
\[
\mathrm{err}_{\sigma}(f) \;=\; \mathrm{RMSE}\!\Bigl(\bigl\{r_{i} \;:\; \sigma_{i} \le Q_{1-f}(\sigma)\bigr\}\Bigr) .
\]
The same construction with MAE in place of RMSE gives an alternative AUSE that is more sensitive to the bulk of the residual distribution and less to outliers; we report it in Figure~\ref{fig:app_sparsification_mae} as a complementary view.
The \emph{oracle} curve $\mathrm{err}_{\mathrm{oracle}}(f)$ replaces $\sigma$ with $|r|$ in the sort key; this is the lowest possible sparsification curve given the same residual distribution. The \emph{sparsification error} is
\[
\mathrm{SE}(f)\;=\;\mathrm{err}_{\sigma}(f)-\mathrm{err}_{\mathrm{oracle}}(f)\;\ge\;0,
\]
and its area integrates to AUSE,
\[
\mathrm{AUSE} \;=\; \int_{0}^{1}\bigl[\mathrm{err}_{\sigma}(f) - \mathrm{err}_{\mathrm{oracle}}(f)\bigr]\,df,
\]
the standard scalar uncertainty-quality metric \citep{Ilg2018, Poggi2020}. Lower is better; $\mathrm{AUSE}\!=\!0$ iff $\sigma$ ranks pixels identically to $|r|$. The integral is evaluated by the trapezoidal rule over 100 equally spaced fractions. The metric is computed on the full pixel population.

\begin{figure}[h]
  \centering
  \kg{\includegraphics[width=\textwidth]{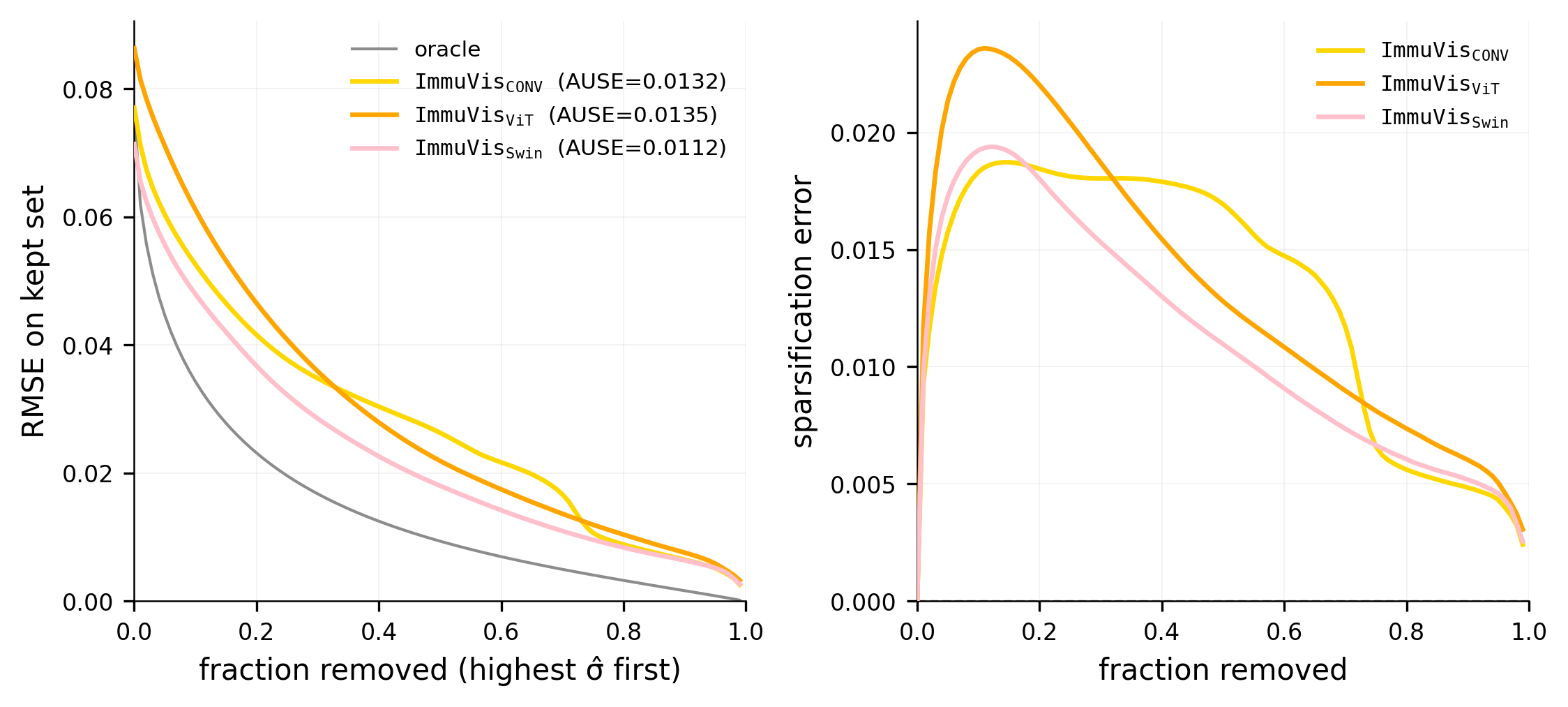}}
  \caption{\kg{\textbf{Sparsification analysis - global, out-of-cohort LOO (RMSE base, matches Table~\ref{tab:uncertainty_calibration}).} \emph{Left:} residual RMSE on the kept $(1-f)$ fraction of pixels after removing the most-uncertain $f$ by predicted $\sigma$ (coloured lines per model); the grey oracle curve is obtained by ranking with $|r|$ instead and represents the lowest achievable curve at this residual distribution. AUSE in the legend is the area between each model's curve and the oracle. \emph{Right:} the sparsification-error integrand $\mathrm{SE}(f)=\mathrm{err}_{\sigma}(f)-\mathrm{err}_{\mathrm{oracle}}(f)$. All three \name variants track the oracle closely across the full range; the residual gap is concentrated in the highest-$\sigma$ tail ($f\!\lesssim\!0.2$), where outlier pixels are hardest to rank exactly. $\nameSwin$ achieves the lowest AUSE, while $\nameViT$ exhibits the largest peak sparsification error.}}
  \label{fig:app_sparsification}
\end{figure}

\begin{figure}[h]
  \centering
  \kg{\includegraphics[width=\textwidth]{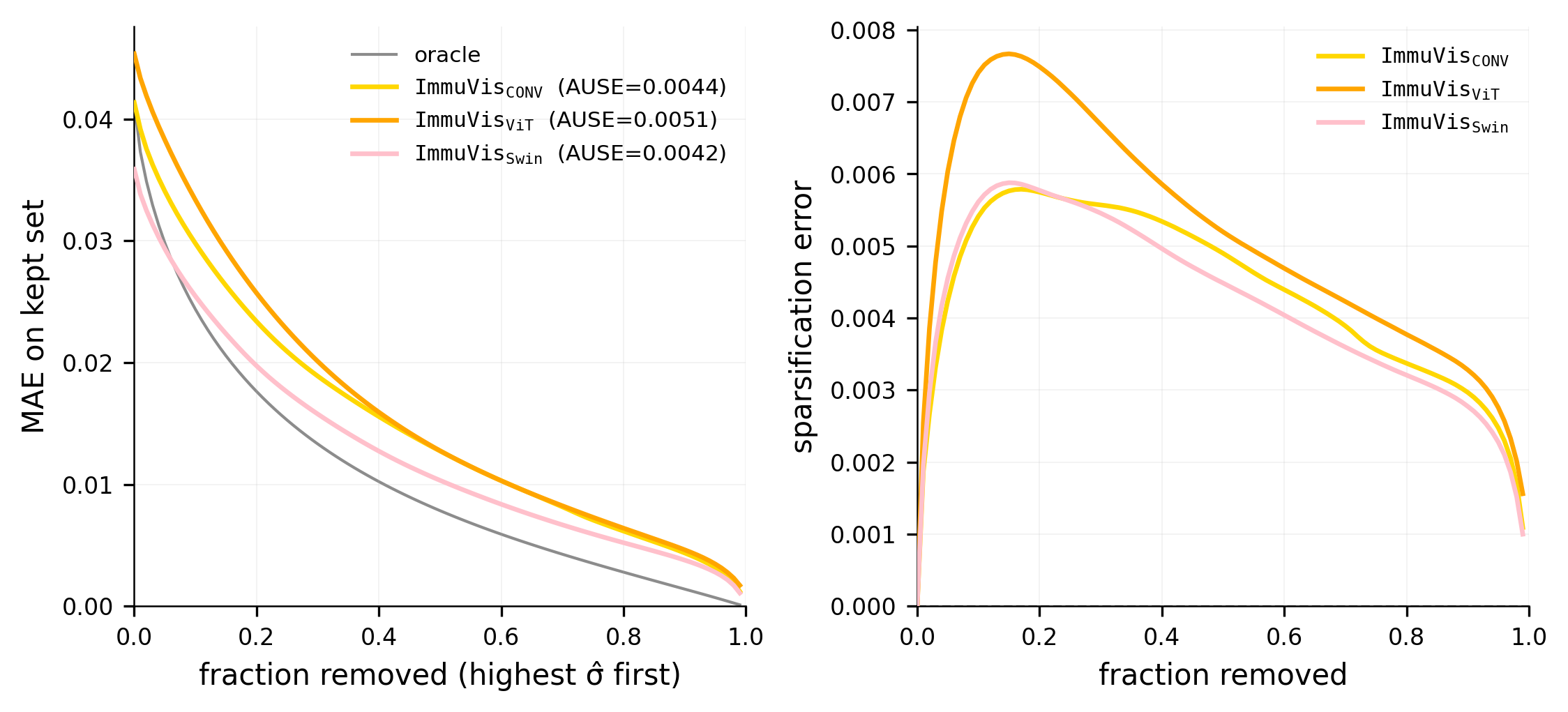}}
  \caption{\textbf{Same construction as Figure~\ref{fig:app_sparsification}} with MAE in place of RMSE. The model ranking is preserved ($\nameSwin$ < $\nameConv$ < $\nameViT$), confirming the $\sigma$-ordering quality differences are robust to the error norm choice.}
  \label{fig:app_sparsification_mae}
\end{figure}

\subsection{\kg{Retrospective PD-1 uncertainty analysis}}
\label{app:pd1}
\kg{We use $\nameConv$ to predict PD-1 leave-one-out on 21,058 segmented CD8+ cells from IMMUcan NSCLC2 panel 1. We extract cell-centred $32\times32$ crops, exclude PD-1 from the inputs, and average predicted expression over the cell mask. Reference PD-1 labels are assignments from a two-component Gaussian mixture fitted to measured, preprocessed cell-mean expression on the arcsinh scale; the component with the larger mean is labelled positive. A prediction threshold is chosen by maximizing Youden's $J$ against these reference labels on the evaluated cells.}

\kg{Cell uncertainty is summarized by the mean predicted pixel standard deviation. The error-ranking ROC-AUC is .78, with error rates of 8.4\% and 77\% in the lowest- and highest-uncertainty deciles. Mean pixel variance and maximum pixel variance give ROC-AUCs of .78 and .74, respectively. These are descriptive uncertainty summaries, rather than estimates of the variance of the cell mean under a validated pixel-covariance model. Since reference-label fitting and threshold selection use the evaluated cells, this is a retrospective error-ranking analysis, not an independent assessment of clinical diagnostic performance.}

\subsection{Single-cell pre-processing}
\label{app:celltyping}
For each segmented cell from the ~\cite{Eling2025} dataset, we extract a $32\!\times\!32$ pixel patch centred at the cell mask centroid.
All pixels outside the target cell mask are zeroed to remove neighborhood information while preserving within-cell marker intensities.
Each crop is encoded by the frozen encoder (no fine-tuning) of all considered models from $\name$ family, and VirTues to obtain a spatial embedding map, which we reduce to a fixed-length vector ($d_{\mathrm{lat}}$) via spatial average pooling.

\subsection{Cell phenotyping results for different cell phenotyping methods}\label{app:phenotyping-methods}

Quantitative results for cell phenotyping across model families and preprocessing configurations are presented in Table~\ref{tab:cell_phenotyping_f1}, extending the main results reported in Table~\ref{tab:cell_phenotyping_agg_f1}. In addition to the \name cell-typing approach used in the main paper (denoted as \texttt{patch}), we also evaluate the cell-typing strategy introduced by the VirTues authors (denoted as \texttt{context}). Subscripts denote training and preprocessing configurations: \texttt{OP} refers to ImmuVis preprocessing, \texttt{VP} to VirTues preprocessing, and \texttt{IMC17M} to VirTues trained on our dataset. 
\kg{The mean-marker baseline requires segmentation masks and
produces feature vectors whose dimensions correspond to the
measured markers. Transferring a classifier across different
panels therefore requires explicit alignment of marker features.
This baseline does not provide virtual staining.}

In the \texttt{patch} method, for each cell identified from the segmentation mask, we extract a $32\!\times\!32$ pixel patch centered at the cell centroid. All pixels outside the target cell mask are zeroed to remove neighborhood information. Each patch is then processed by the frozen encoder of all considered models from the $\name$ family, as well as VirTues, to obtain a spatial embedding map, which is subsequently reduced to a fixed-length vector ($d_{\mathrm{lat}}$) via spatial average pooling.

The \texttt{context} method follows the design proposed by the VirTues authors and incorporates both cellular and local microenvironment information. Each image, together with its corresponding cell mask, is partitioned into overlapping $128 \times 128$ crops using a stride of 42 pixels. Each crop is further decomposed into $8 \times 8$ patches. For VirTues, which is based on a Vision Transformer architecture, embeddings are produced natively at the patch level. For ImmuVis, each $8 \times 8$ patch is independently embedded as an image region. To obtain cell-level representations, patch embeddings are aggregated using a mask-informed weighted average: for each cell, the contribution of a given patch is proportional to the fraction of pixels within that patch belonging to the cell, ensuring that patches overlapping multiple cells contribute proportionally to each cell's embedding.

\kg{The reported $\nameConv$ context scores use the original protocol of independently encoding $8\times8$ patches, although the network was trained on $128\times128$ inputs and downsamples fourfold per stage. Encoding the full crop once and pooling patch-aligned latent regions with the same mask weights improves the Eling context Macro-F1 from .533 to .752 (patch protocol: .790; $\nameViT$ context: .759). Danenberg and Cords context scores (.488 and .666) were not recomputed. Tables~\ref{tab:cell_phenotyping_f1} and \ref{tab:cell_pehnotyping_ap} retain the original context protocol for $\nameConv$; the corrected Eling score is reported separately here.}

\begin{figure}[ht]
  \begin{center}
    \centerline{\kg{\includegraphics[width=\textwidth]{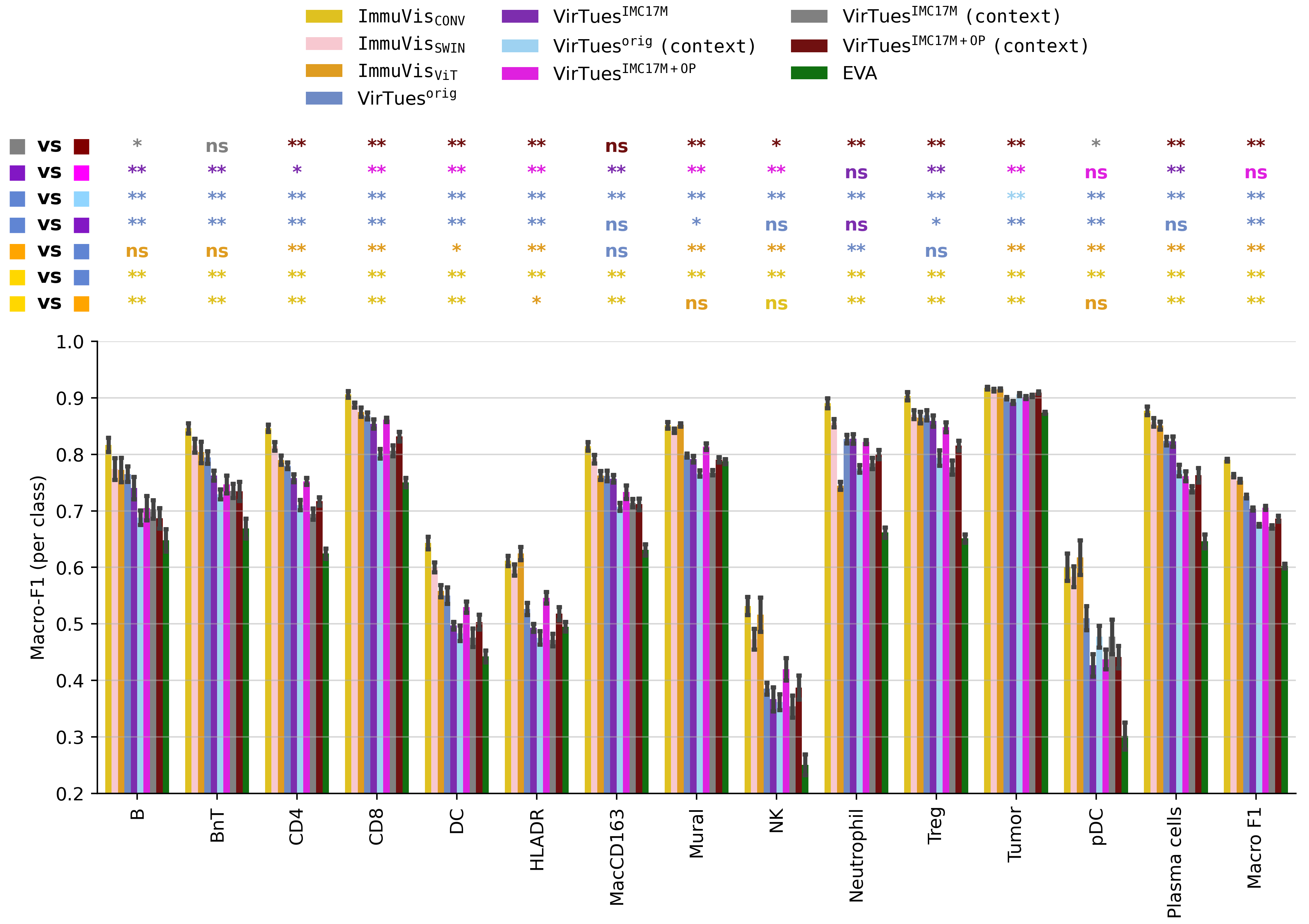}}
    }
    \caption{
    \textbf{Cell-typing performance from learned representations on the \cite{Eling2025} dataset.} Bars show mean marco-F1 across cross-validation folds with confidence intervals (whiskers); higher is better. The seven rows above the plot report paired Wilcoxon signed-rank tests on per-fold scores for the corresponding model pairs (indicated by the colored squares on the left). Significance markers are color-coded to match the model with the higher mean score. Significance is shown after FDR correction (ns - not significant; $(^{*}) <\!0.05$; $(^{**}) <\!0.01$; $(^{***}) <\!0.001$).}
    \label{fig:celltyping_gs}
  \end{center}
\end{figure}

\begin{figure}[ht]
  \begin{center}
    \centerline{\kg{\includegraphics[width=\textwidth]{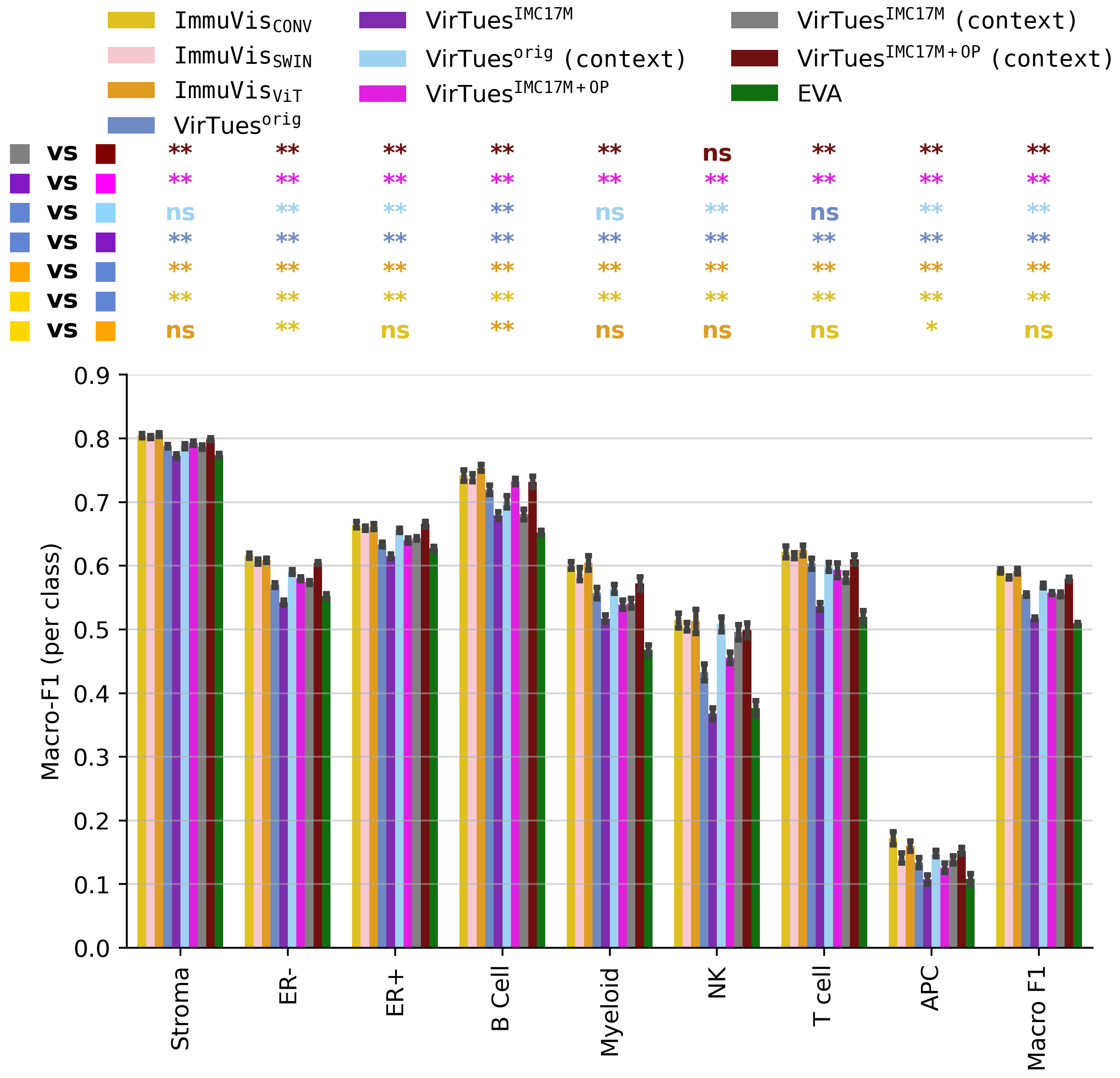}}
    }
    \caption{
    \textbf{Cell-typing performance from learned representations on the \cite{Danenberg2022} dataset.} Bars show mean marco-F1 across cross-validation folds with confidence intervals (whiskers); higher is better. The seven rows above the plot report paired Wilcoxon signed-rank tests on per-fold scores for the corresponding model pairs (indicated by the colored squares on the left). Significance markers are color-coded to match the model with the higher mean score. Significance is shown after FDR correction (ns - not significant; $(^{*}) <\!0.05$; $(^{**}) <\!0.01$; $(^{***}) <\!0.001$).}
    \label{fig:celltyping_danenberg}
  \end{center}
\end{figure}

\begin{figure}[ht]
  \begin{center}
    \centerline{\kg{\includegraphics[width=\textwidth]{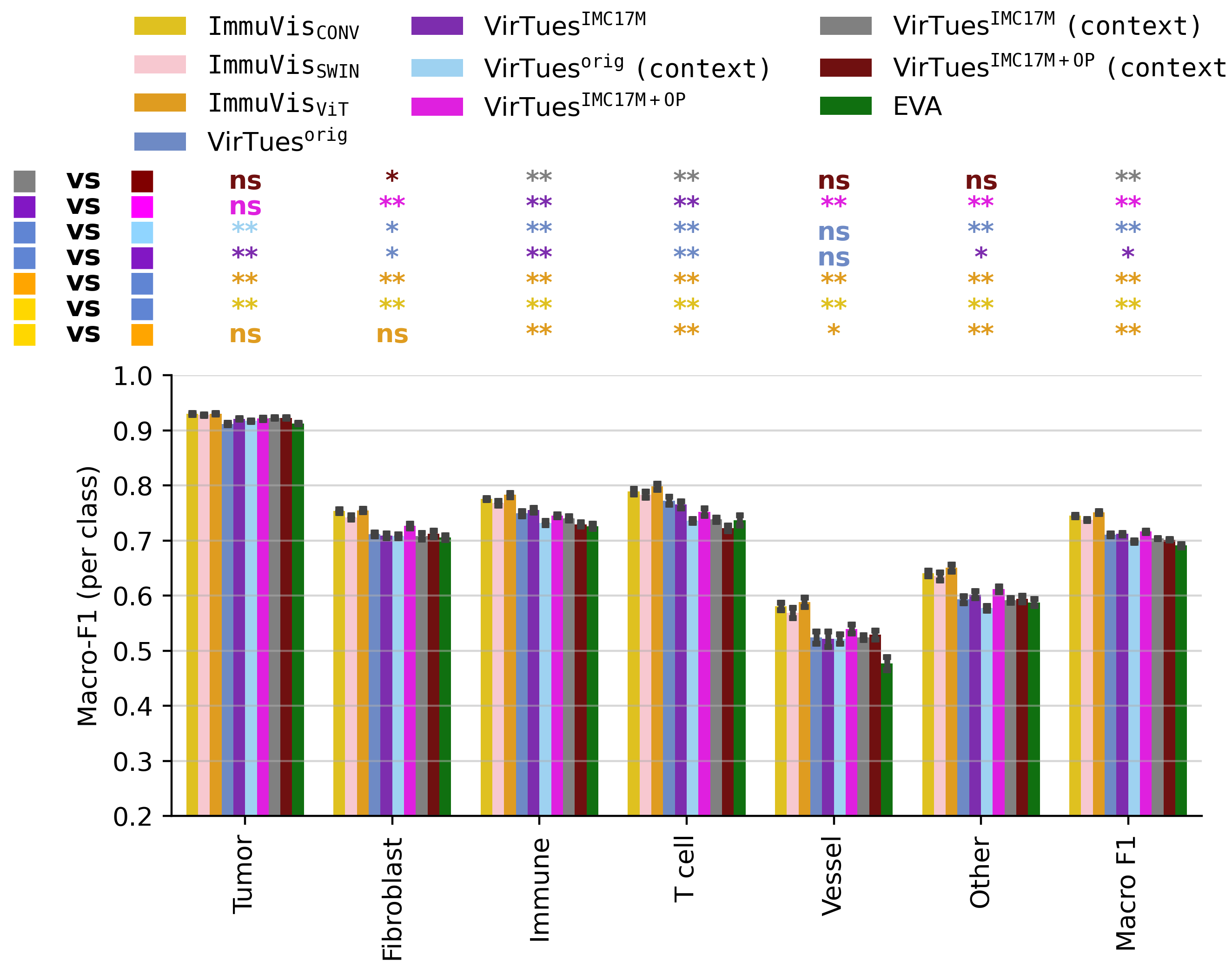}}
    }
    \caption{
    \textbf{Cell-typing performance from learned representations on the \cite{Cords2024} dataset.} Bars show mean marco-F1 across cross-validation folds with confidence intervals (whiskers); higher is better. The seven rows above the plot report paired Wilcoxon signed-rank tests on per-fold scores for the corresponding model pairs (indicated by the colored squares on the left). Significance markers are color-coded to match the model with the higher mean score. Significance is shown after FDR correction (ns - not significant; $(^{*}) <\!0.05$; $(^{**}) <\!0.01$; $(^{***}) <\!0.001$).}
    \label{fig:celltyping_cords}
  \end{center}
\end{figure}

Complementary results in terms of Average Precision are reported in Tables~\ref{tab:cell_phenotyping_agg_ap} and \ref{tab:cell_pehnotyping_ap}, while Figures~\ref{fig:celltyping_gs}, \ref{fig:celltyping_danenberg}, and \ref{fig:celltyping_cords} provide dataset-specific macro-F1 comparisons with statistical significance testing across all evaluated datasets.

\begin{table}[ht]
\caption{\textbf{Representation learning performance for cell typing.}
Macro-F1 scores are reported for 10-fold cross-validation; higher values indicate better performance.
Standard deviations across folds were below $10^{-2}$ for all entries and are therefore omitted.
The best score for each endpoint is shown in \textbf{bold}. The best method per endpoint / method is \underline{underlined}.}

  \label{tab:cell_phenotyping_f1}
  \centering
  \begin{small}
  \begin{threeparttable}
    \begin{tabular}{lcccccccc}
      \toprule
      {Model} &
        \multicolumn{2}{c}{~\cite{Eling2025}} &
        \multicolumn{2}{c}{~\cite{Danenberg2022}}
        &\multicolumn{2}{c}{~\cite{Cords2024}}
        &\multicolumn{2}{c}{Avg.}\\
     \midrule
     
     Method & $\texttt{patch}$ & $\texttt{context}$ & $\texttt{patch}$ & $\texttt{context}$ & $\texttt{patch}$ & $\texttt{context}$ & $\texttt{patch}$ & $\texttt{context}$ \\
     \midrule
      {$\text{EVA}_{\texttt{OP}}$} & \underline{.602} & \underline{.450} & \underline{.510} & \underline{.487} & \underline{.691} & \underline{.571} & \underline{.601} & \underline{.503} \\
      \midrule
      {$\text{VirTues}$} & \underline{.726} & .675 & .555 & .569 & .711 & .699 & \underline{.664} & .648\\

      {$\text{VirTues}_{\texttt{IMC17M}}$} & .703 & .672 & .517 & .555 & .712 & \underline{.704} & .644 & .644\\

      {$\text{VirTues}_{\texttt{IMC17M+OP}}$} & .706 & \underline{.686} & \underline{.557} & \underline{.579} & \underline{.716} & .702 & .660 & \underline{.656}\\
      \midrule
      {$\nameViT$} & .753 & \underline{\textbf{.759}} & \underline{\textbf{.592}} & \underline{\textbf{.618}} & \underline{\textbf{.751}} & \underline{\textbf{.740}} & .699 & \underline{\textbf{.706}}\\

    {$\nameConv$} & \underline{\textbf{.790}} & .533 & \underline{\textbf{.592}} & .488 & .745 & .666 & \underline{\textbf{.709}} & .562\\

     {$\nameSwin$} & .762 & .752 & .582 & .611 & .738 & .732 & .694 & .698\\

    \midrule
      {$\texttt{ImmuVis}_{\texttt{VIT+VP}}$} & \underline{.757} & \underline{.750} & \underline{.581} & \underline{.610} & \underline{.723} & \underline{.722} & \underline{.687} & \underline{.694}\\

    {$\texttt{ImmuVis}_{\texttt{Conv+VP}}$} & .671 & .638 & .551 & .540 & .691 & .652 & .638 & .610\\

      \bottomrule
    \end{tabular}
  \end{threeparttable}
  \end{small}
\end{table}

\begin{table}[ht]
\caption{Representation learning performance for cell typing (Average Precision; higher is better).
Scores are reported from 10-fold cross-validation. Standard deviations across folds were below $10^{-2}$ for all entries and are therefore omitted. For VirTues evaluations, the reported value is the higher score obtained using either the authors' original or the \name cell-typing approach (see~\ref{app:phenotyping-methods}).
(\underline{best score} within models family, \textbf{best score} for each endpoint)
}
  \label{tab:cell_phenotyping_agg_ap}
  \centering
  \begin{small}
  \begin{threeparttable}
    \begin{tabular}{lcccc}
      \toprule
      {Model} &
        ~\cite{Eling2025} &
        ~\cite{Danenberg2022}
        &~\cite{Cords2024} &~Avg. \\
     \midrule
      {$\text{EVA}_{\texttt{OP}}$} & \underline{.664} & \underline{.559} & \underline{.748} & \underline{.657} \\
      \midrule
      {$\text{VirTues}$} & \underline{.812} & .635* & .781 & \underline{.743}\\

      {$\text{VirTues}_{\texttt{IMC17M}}$} & .786 & .624* & \underline{.783} & .731\\

      {$\text{VirTues}_{\texttt{IMC17M+OP}}$} & .796 & \underline{.651*} & \underline{.783} & \underline{.743}\\
      \midrule
      {$\nameViT$} & .845 & .662 & \underline{\textbf{.827}} & .778\\

    {$\nameConv$} & \underline{\textbf{.882}} & \underline{\textbf{.668}} & .822 & \underline{\textbf{.791}}\\

    {$\nameSwin$} & .850 & .654 & .807 & .770\\

      \bottomrule
    \end{tabular}
    \begin{tablenotes}
    \item[*]{\footnotesize VirTues cell-typing method gave better results.}
    \end{tablenotes}
  \end{threeparttable}
  \end{small}
\end{table}

\begin{table}[ht]
\caption{\kg{\textbf{Representation learning performance for cell typing.} $\nameConv$ context entries use the original $8\times8$ protocol (see text for corrected Eling Macro-F1).}
Average Precision scores are reported for 10-fold cross-validation; higher values indicate better performance.
Standard deviations across folds were below $10^{-2}$ for all entries and are therefore omitted.
The best score for each endpoint is shown in \textbf{bold}. The best method per endpoint / method is \underline{underlined}.}
  \label{tab:cell_pehnotyping_ap}
  \centering
  \begin{small}
  \begin{threeparttable}
    \begin{tabular}{lcccccccc}
      \toprule
      {Model} &
        \multicolumn{2}{c}{~\cite{Eling2025}} &
        \multicolumn{2}{c}{~\cite{Danenberg2022}}
        &\multicolumn{2}{c}{~\cite{Cords2024}}
        &\multicolumn{2}{c}{Avg.} \\
     \midrule
     
     Method & $\texttt{patch}$ & $\texttt{context}$ & $\texttt{patch}$ & $\texttt{context}$ & $\texttt{patch}$ & $\texttt{context}$ & $\texttt{patch}$ & $\texttt{context}$ \\
     \midrule
      {$\text{EVA}_{\texttt{OP}}$} & \underline{.664} & \underline{.472} & \underline{.559} & \underline{.534} & \underline{.748} & \underline{.606} & \underline{.657} & \underline{.537}\\
      \midrule
      {$\text{VirTues}$} & \underline{.812} & .760 & \underline{.625} & .635 & .781 & .776 & \underline{.739} & .724 \\

      {$\text{VirTues}_{\texttt{IMC17M}}$} & .786 & .756 & .574 & .624 & \underline{.783} & \underline{.778} & .714 & .719 \\

      {$\text{VirTues}_{\texttt{IMC17M+OP}}$} & .796 & \underline{.771} & .624 & \underline{.651} & \underline{.783} & .771 & .734 & \underline{.731} \\
      \midrule
      {$\nameViT$} & .845 & \underline{\textbf{.848}} & .662 & \underline{\textbf{.695}} & \underline{\textbf{.827}} & \underline{\textbf{.821}} & .778 & \underline{\textbf{.788}} \\

    {$\nameConv$} & \underline{\textbf{.882}} & .585 & \underline{\textbf{.668}} & .532 & .822 & .718 & \underline{\textbf{.791}} & .612 \\

     {$\nameSwin$} & .850 & .841 & .654 & .691 & .807 & .811 & .770 & .781 \\

    \midrule
      {$\texttt{ImmuVis}_{\texttt{VIT+VP}}$} & \underline{.847} & \underline{.840} & \underline{.649} & \underline{.686} & \underline{.793} & \underline{.799} & \underline{.763} & \underline{.775} \\

    {$\texttt{ImmuVis}_{\texttt{Conv+VP}}$} & .752 & .714 & .617 & .599 & .752 & .702 & .707 & .672 \\

      \bottomrule
    \end{tabular}
  \end{threeparttable}
  \end{small}
\end{table}

\subsection{Extended clinical endpoint prediction results across model families and preprocessing protocols}
\label{app:clinical_results_under_vp}

Quantitative results for clinical endpoint prediction across model families and preprocessing configurations are reported in Table~\ref{tab:clinical_representation_extended}, extending the main results of Table~\ref{tab:clinical_representation}. In addition to the linear probe (LogReg) used in the main paper, this extended evaluation also includes the abMIL probe. Subscripts denote training and preprocessing configurations: \texttt{OP} refers to ImmuVis preprocessing, \texttt{VP} to VirTues preprocessing, and \texttt{IMC17M} to VirTues trained on our dataset.

\paragraph{\kg{Mean-marker baseline.}}
\kg{We compute mean marker intensities within non-overlapping
$128\times128$ patches and average the resulting vectors per image.
We evaluate raw intensities, OP, and VP preprocessing.
The VP configuration applies per-marker 99th-percentile clipping
based on non-zero pixels, log1p transformation, Gaussian smoothing
($\sigma=1$), and intensity clipping and rescaling.
It does not use the per-marker z-score normalization described for VP.
Features are standardized within each training fold and classified
using logistic regression with stratified 10-fold cross-validation.}

\paragraph{\kg{Statistical testing.}}
\kg{We assess differences between $\name$ and VirTues using
paired Wilcoxon signed-rank tests on per-fold Macro-F1 scores
with Benjamini--Hochberg FDR correction.}

\paragraph{\kg{Extended results.}}
\kg{Table~\ref{tab:clinical_representation_extended} extends
Table~\ref{tab:clinical_representation} with additional preprocessing
configurations and ABMIL results. For $\name$ variants, logistic
regression achieves higher average Macro-F1 than ABMIL.
Among the models in the main table, $\nameSwin$ achieves the highest
displayed means for relapse and Cords grade, and shares the highest
ERBB2 mean with $\nameConv$.
$\nameConv$ and $\nameViT$ share the highest ER mean.
VirTues achieves the highest Danenberg grade mean and shares
the highest Cords subtype mean with $\nameSwin$.}

\kg{These endpoints are constrained by the information in a small
IMC tissue region: relapse depends on follow-up and systemic disease,
while grade and subtype are assessed from wider tissue material.
Small clinical cohorts also limit the precision of between-model
comparisons.}

\begin{table}[ht]
\caption{Representation learning for clinical endpoints (Macro-F1; higher is better).
Values report mean$\pm$std Macro-F1 across 10-fold cross-validation; best performance per endpoint is shown in bold.}
  \label{tab:clinical_representation_extended}
  \centering
  \begin{small}
  \begin{threeparttable}
    \resizebox{\textwidth}{!}{%
    \begin{tabular}{llcccccccc}
      \toprule
      {Model} & {Probe} &
        \multicolumn{4}{c}{Danenberg et al.~\cite{Danenberg2022}}
        & \multicolumn{3}{c}{Cords et al.~\cite{Cords2024}} & \\
      \cmidrule(lr){3-6}\cmidrule(lr){7-9}
     & & PAM50 & Grade & ER & ERBB2 & Subtype & Relapse & Grade & Avg.\\
      \midrule
      
      \multirow{2}{*}{$\text{EVA}_{\texttt{OP}}$} 
      & LogReg &.38±.05 & .50±.08 & .75±.07 & .61±.08 & .79±.03 & .57±.05 & .50±.05 & .585 \\
      & abMIL  & .40±.03 & .45±.03 & .80±.07 & .81±.15 & .81±.02 & .55±.05 & .48±.04 & .614 \\
      \midrule
      
      \multirow{2}{*}{$\text{VirTues}$} 
      & LogReg & .37±.05 & \textbf{.53±.05} & .78±.08 & .76±.11 & .82±.03 & .57±.03 & .51±.07 & .620 \\
      & abMIL  & .42±.06 & .48±.04 & .77±.09 & .66±.11 & .83±.02 & .58±0.03 & .48±.03 & .603 \\
      \addlinespace
      
      \multirow{2}{*}{$\text{VirTues}_{\texttt{IMC17M}}$}  
      & LogReg & .36±.06 & .49±.05 & .77±.06 & .78±.13 & \textbf{.83±.02} & .58±.04 & .54±.06 & .621 \\
      & abMIL  & \textbf{.43±.04} & .52±.06 & .76±.09 & .74±.10 & \textbf{.83±.02} & .56±.03 & .52±.05 & .623 \\
      \addlinespace
      
      \multirow{2}{*}{$\text{VirTues}_{\texttt{IMC17M+OP}}$}  
      & LogReg & .40±.08 & .49±.08 & .80±.06 & .77±.13 & .83±.03 & .57±.05 & .53±.09 & .627 \\
      & abMIL  & .41±.04 & .46±.05 & .78±.08 & .79±.15 & .80±.03 & .54±.04 & .47±0.06 & .607 \\
      \midrule
      
      \multirow{2}{*}{$\nameViT$} 
      & LogReg & .40±.06 & .48±.05 & \textbf{.81±.04} & .77±.12 & .82±.03 & .58±.04 & .54±.06 & .628 \\
      & abMIL  & \textbf{.43±.04} & .46±.02 & .76±.08 & .76±.15 & .81±.02 & .56±.03 & .47±.04 & .608 \\
      \addlinespace
      
      \multirow{2}{*}{$\nameConv$} 
      & LogReg & .40±.09 & .52±.08 & .81±.07 & .80±.13 & .81±.03 & .59±.03 & .54±.05 & .638 \\
      & abMIL  & .42±.05 & .45±.01 & .78±.08 & .79±.16 & .82±.02 & .58±.04 & .48±.04 & .618 \\
      \addlinespace
      
      \multirow{2}{*}{$\nameSwin$} 
      & LogReg & .42±.09 & .49±.06 & .79±.05 & .80±.13 & .83±.03 & \textbf{.61±.04} & \textbf{.55±.06} & \textbf{.641} \\
      & abMIL  & \textbf{.43±.04} & .47±.04 & .78±.07 & .80±.16 & \textbf{.83±.02} & .57±.04 & .50±.04 & .626 \\

      \midrule

      \multirow{2}{*}{$\name_{\texttt{ViT+VP}}$} 
      & LogReg & .39±.11 & .49±.10 & .78±.06 & .79±.10 & .83±.02 & .59±.02 & .51±.04 & .626 \\
      & abMIL  & .43±.07 & .46±.04 & .74±.06 & .81±.10 & .80±.03 & .56±.04 & .46±.06 & .609 \\
      \addlinespace
      
      \multirow{2}{*}{$\name_{\texttt{Conv+VP}}$}  
      & LogReg & .41±.09 & .49±.06 & .80±.08 & .78±.09 & .81±.03 & .57±.03 & .51±.06 & .624 \\
      & abMIL  & .41±.08 & .44±.06 & .73±.05 & \textbf{.82±.07} & .79±.03 & .55±.06 & .45±.04 & .600 \\
      \bottomrule

    \end{tabular}}
  \end{threeparttable}
  \end{small}
\end{table}

\subsection{Locality analysis}
\label{app:locality}

\begin{figure}[ht]
  \begin{center}
    \centerline{\kg{\includegraphics[scale=1.]{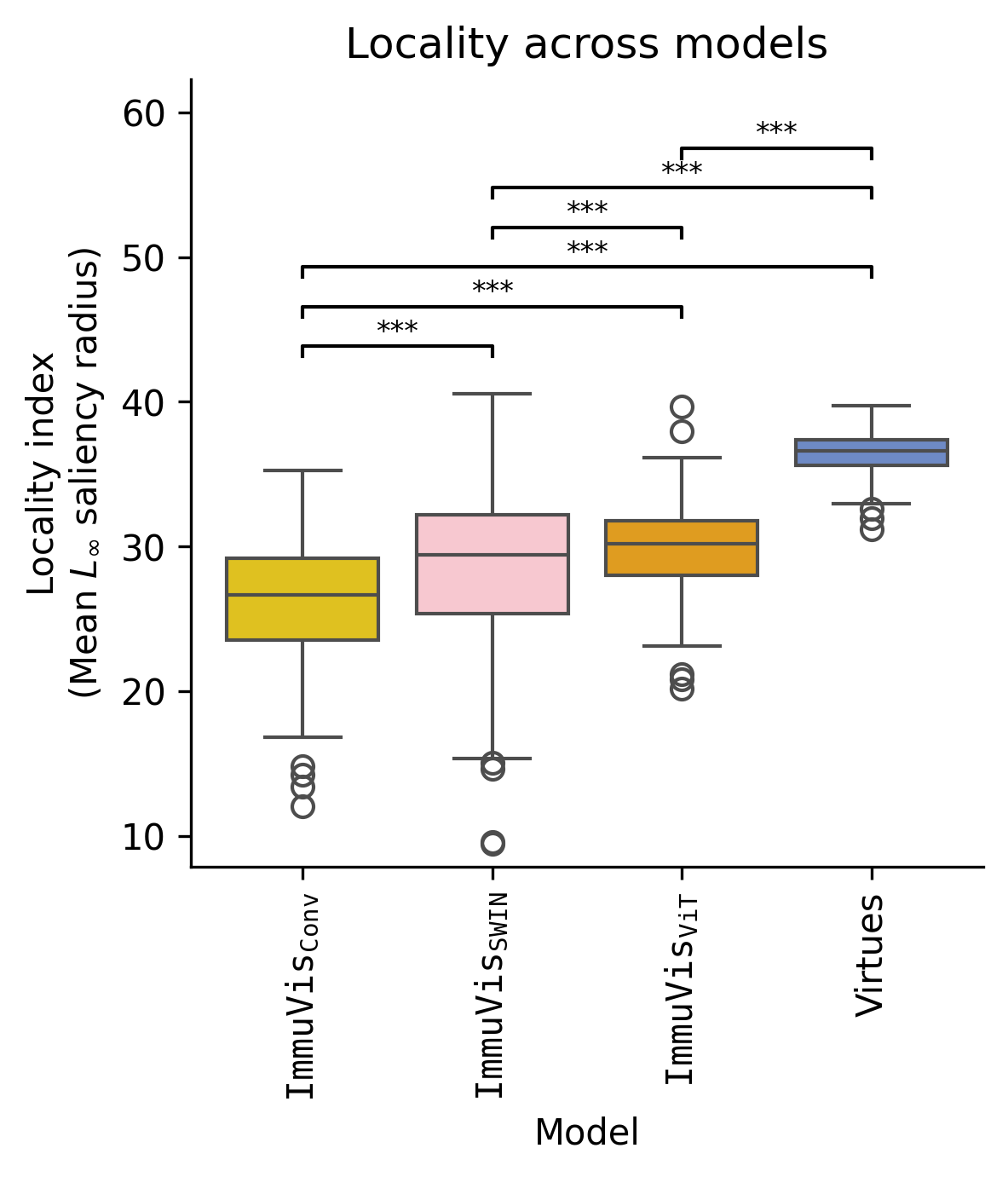}}}
    \caption{
      \textbf{Locality index analysis on HN.}
    Distributions of the mean $L_{\infty}$ saliency radius for
    $\name$ variants and VirTues. Smaller values indicate more
    spatially local influence. Pairwise comparisons use paired
    Wilcoxon signed-rank tests (***: $p<0.001$).
    }
    \label{fig:locality}
  \end{center}
\end{figure}

\kg{In order to assess the locality of different models, we introduced a saliency-based \textit{locality index}. This index was computed for a leave-one-out virtual staining task (performed on the IMMUcan HN, Cords and Danenberg cohorts). For each image and each marker, we computed a saliency map as the element-wise absolute value of the gradient of the predicted marker expression at the central pixel. We then normalized this gradient to sum to 1 across all input marker expression images, yielding a probability distribution. For this distribution, we defined the locality index as the average $L_{\infty}$ distance from the central pixel.}

Figure~\ref{fig:locality} shows the locality-index distributions on HN, while Table~\ref{tab:locality-replication} reports mean locality radii across HN, Cords, and Danenberg.
$\nameConv$ achieved the lowest average locality index across images and markers, outperforming $\nameSwin$ and $\nameViT$. VirTues exhibited a substantially higher locality index.

\kg{$\nameConv$ and $\nameSwin$ fuse markers after the stem, whereas the $\nameViT$ hyperkernel acts as the $8\times8$ tokenizer; token-based alternatives add marker identity after tokenization. The replication supports the separation from VirTues, but differences within the $\name$ family are small and the index does not explain their task-performance ordering.}
\begin{table}[ht]
\centering\small
\caption{\kg{Mean saliency locality radius on three cohorts; smaller values indicate more local influence.}}
\label{tab:locality-replication}
\begin{tabular}{lccc}
\toprule
\kg{Model} & \kg{Immucan H\&N} & \kg{Cords} & \kg{Danenberg}\\
\midrule
\kg{$\nameConv$} & \kg{26.0} & \kg{26.1} & \kg{24.1}\\
\kg{$\nameSwin$} & \kg{28.5} & \kg{26.6} & \kg{25.8}\\
\kg{$\nameViT$} & \kg{29.9} & \kg{29.8} & \kg{29.1}\\
\kg{VirTues} & \kg{36.5} & \kg{37.9} & \kg{37.8}\\
\bottomrule
\end{tabular}
\end{table}

\FloatBarrier
\kg{\section*{NeurIPS Paper Checklist}

\begin{enumerate}

\item {\bf Claims}
    \item[] Question: Do the main claims made in the abstract and introduction accurately reflect the paper's contributions and scope?
    \item[] Answer: \answerYes{} %
    \item[] Justification: The abstract and introduction accurately summarize the paper's contributions, including the marker-adaptive \name model family, the \datasetNameNoSpace dataset, uncertainty-aware virtual staining, and downstream evaluations. The claims are supported by the reported experiments and are framed within the intended IMC/spatial proteomics setting without overstating generalization beyond the evaluated data. %

\item {\bf Limitations}
    \item[] Question: Does the paper discuss the limitations of the work performed by the authors?
    \item[] Answer: \answerYes{} %
    \item[] Justification: Discussion section includes dedicated paragraph that mentions current limitations of the \name family as well as the future work paragraph that addresses what is not yet covered by the proposed work. %

\item {\bf Theory assumptions and proofs}
    \item[] Question: For each theoretical result, does the paper provide the full set of assumptions and a complete (and correct) proof?
    \item[] Answer: \answerYes{}%
    \item[] Justification: Across the entire paper all mathematical/theoretical claims are justified in-place or the reader is redirected to a dedicated Appendix section covering the proofs supporting any presented claims. %

    \item {\bf Experimental result reproducibility}
    \item[] Question: Does the paper fully disclose all the information needed to reproduce the main experimental results of the paper to the extent that it affects the main claims and/or conclusions of the paper (regardless of whether the code and data are provided or not)?
    \item[] Answer: \answerYes{} %
    \item[] Justification: The paper provides the architecture, preprocessing, training setup, dataset provenance, splitting strategy, baseline checkpoints, and evaluation protocols required to reproduce the main results. Public datasets are cited to their original sources, and restricted consortium-controlled data are documented with their access constraints. The codebase is provided in anonymized form. %

\item {\bf Open access to data and code}
    \item[] Question: Does the paper provide open access to the data and code, with sufficient instructions to faithfully reproduce the main experimental results, as described in supplemental material?
    \item[] Answer: \answerYes{} %
    \item[] Justification: The paper provides an anonymized repository with code and instructions for reproducing the main experiments, including environment setup, preprocessing, training, and evaluation scripts. Public datasets are cited with access instructions, while consortium-controlled IMMUcan data cannot be redistributed and are used under the applicable data-governance terms. %

\item {\bf Experimental setting/details}
    \item[] Question: Does the paper specify all the training and test details (e.g., data splits, hyperparameters, how they were chosen, type of optimizer) necessary to understand the results?
    \item[] Answer: \answerYes{} %
    \item[] Justification: The paper specifies the main experimental settings in the core text, including dataset splits, held-out cohorts, preprocessing protocols, evaluation tasks, and baseline comparisons. Detailed training configurations, model architectures, hyperparameters, optimizer settings, downstream evaluation procedures, and additional reproducibility details are provided in the appendices and accompanying repository. %

\item {\bf Experiment statistical significance}
    \item[] Question: Does the paper report error bars suitably and correctly defined or other appropriate information about the statistical significance of the experiments?
    \item[] Answer: \answerYes{} %
    \item[] Justification: Main quantitative results include appropriate uncertainty or significance reporting, including cross-validation mean$\pm$standard deviation and paired statistical tests with FDR correction where applicable. %

\item {\bf Experiments compute resources}
    \item[] Question: For each experiment, does the paper provide sufficient information on the computer resources (type of compute workers, memory, time of execution) needed to reproduce the experiments?
    \item[] Answer: \answerYes{} %
    \item[] Justification: Inference-time resource usage is reported in Table~\ref{tab:efficiency}; training and evaluation compute are detailed in the Training compute resources paragraph of Appendix A.6. %
    
\item {\bf Code of ethics}
    \item[] Question: Does the research conducted in the paper conform, in every respect, with the NeurIPS Code of Ethics \url{https://neurips.cc/public/EthicsGuidelines}?
    \item[] Answer: \answerYes{} %
    \item[] Justification: The research conforms to the NeurIPS Code of Ethics.  %

\item {\bf Broader impacts}
    \item[] Question: Does the paper discuss both potential positive societal impacts and negative societal impacts of the work performed?
    \item[] Answer: \answerYes{} %
    \item[] Justification:  The Discussion section (Future overlook paragraph) acknowledges that if IMC adoption extends from research to clinical settings, \name could contribute to diagnostic pipelines, and notes the role of calibrated uncertainty as a reliability safeguard. No direct negative societal impacts are foreseen for this biomedical imaging method.%
    
\item {\bf Safeguards}
    \item[] Question: Does the paper describe safeguards that have been put in place for responsible release of data or models that have a high risk for misuse (e.g., pre-trained language models, image generators, or scraped datasets)?
    \item[] Answer: \answerNA{} %
    \item[] Justification: The paper does not release data or models with high misuse risk in the sense of the checklist. The model does not generate natural images, text, or other content categories typically associated with dual-use or unsafe release concerns. %

\item {\bf Licenses for existing assets}
    \item[] Question: Are the creators or original owners of assets (e.g., code, data, models), used in the paper, properly credited and are the license and terms of use explicitly mentioned and properly respected?
    \item[] Answer: \answerYes{} %
    \item[] Justification: Existing datasets, model checkpoints, and code assets used in the paper are credited through citations and documented in the appendix. Public datasets are referenced to their source publications, while consortium-controlled IMMUcan data are used under the applicable data-governance terms and are not redistributed. Baseline models such as VirTues and EVA are cited and used from the authors' released resources where available. Released derived assets will document provenance, access conditions, and usage terms. %

\item {\bf New assets}
    \item[] Question: Are new assets introduced in the paper well documented and is the documentation provided alongside the assets?
    \item[] Answer: \answerYes{} %
    \item[] Justification: The paper introduces new model assets and documents the curated IMC17M resource. We provide documentation of the model architecture, preprocessing, training, evaluation protocols, and limitations in the main paper and appendix. IMC17M includes public datasets and consortium-controlled data; therefore, protected data are not redistributed (until data embargo holds), and access is governed by the corresponding original licenses, approvals, and data-use agreements. The code is available at \url{https://anonymous.4open.science/r/ImmuVis-9173}. %

\item {\bf Crowdsourcing and research with human subjects}
    \item[] Question: For crowdsourcing experiments and research with human subjects, does the paper include the full text of instructions given to participants and screenshots, if applicable, as well as details about compensation (if any)? 
    \item[] Answer: \answerNA{}  %
    \item[] Justification: The paper does not involve crowdsourcing, participant-facing experiments, or collection of new data from human subjects. The study uses existing IMC datasets and consortium-controlled data; no instructions were given to participants and no compensation was provided by the authors.  %

\item {\bf Institutional review board (IRB) approvals or equivalent for research with human subjects}
    \item[] Question: Does the paper describe potential risks incurred by study participants, whether such risks were disclosed to the subjects, and whether Institutional Review Board (IRB) approvals (or an equivalent approval/review based on the requirements of your country or institution) were obtained?
    \item[] Answer: \answerYes{} %
    \item[] Justification: The paper does not collect new human-subject data or interact with participants. It analyzes existing IMC datasets with clinical annotations: public datasets are used under their original data-use terms and approvals, and IMMUcan consortium data are used under the consortium's controlled data-governance framework. No additional participant-facing procedures were performed by the authors. %

\item {\bf Declaration of LLM usage}
    \item[] Question: Does the paper describe the usage of LLMs if it is an important, original, or non-standard component of the core methods in this research? Note that if the LLM is used only for writing, editing, or formatting purposes and does \emph{not} impact the core methodology, scientific rigor, or originality of the research, declaration is not required.
    \item[] Answer: \answerNA{} %
    \item[] Justification: \answerNA{} %

\end{enumerate}}

\end{document}